\documentclass[10pt,twocolumn,letterpaper]{article}

\usepackage{cvpr}

\usepackage[utf8]{inputenc}
\usepackage[T1]{fontenc}
\usepackage{url}
\usepackage{booktabs}
\usepackage{amsfonts}
\usepackage{nicefrac}
\usepackage{microtype}
\usepackage{xcolor}
\usepackage{multirow}
\usepackage{graphicx}
\usepackage{subcaption}
\usepackage{caption}
\usepackage{array}
\usepackage{makecell}
\usepackage{amssymb}
\usepackage{xspace}
\usepackage{amsmath}
\usepackage[table]{xcolor}
\usepackage{bbm}
\usepackage{pifont}
\usepackage{makecell}
\usepackage{enumitem}
\usepackage{tikz}
\usepackage{svg}

\definecolor{cvprblue}{rgb}{0.21,0.49,0.74}
\usepackage[pagebackref,breaklinks,colorlinks,allcolors=cvprblue]{hyperref}
\hypersetup{
    colorlinks=true,
    urlcolor=magenta
}

\def\paperID{3948}
\def\confName{CVPR}
\def\confYear{2026}

\definecolor{darkmagenta}{RGB}{186, 0, 186}
\definecolor{slateblue}{RGB}{90, 70, 230}

\definecolor{semantic}{HTML}{b4445c}     %
\definecolor{normal}{HTML}{3b8b5f}       %
\definecolor{depth}{HTML}{4a6fa5}        %
\definecolor{opticalflow}{HTML}{7e57c2}  %
\definecolor{sceneflow}{HTML}{d9822b}    %
\definecolor{shading}{HTML}{003153}      %
\definecolor{albedo}{HTML}{a97439}       %

\definecolor{todo}{HTML}{E15554}
\definecolor{author1}{HTML}{40A050}
\definecolor{author2}{HTML}{C33C54}
\definecolor{author3}{HTML}{E6AA68}
\definecolor{author4}{HTML}{0F4893}

\newcommand{\Enc}{\mathcal{E}}       %
\newcommand{\Dec}{\mathcal{D}}       %
\newcommand{\teaserrow}[2]{%
    \includegraphics[width=\imgsize]{figures/teaser/#1/#2_.jpg}%
    & \includegraphics[width=\imgsize]{figures/teaser/#1/#2_next.jpg}%
    & \includegraphics[width=\imgsize]{figures/teaser/#1/#2_semantic.jpg}%
    & \includegraphics[width=\imgsize]{figures/teaser/#1/#2_normal.jpg}%
    & \includegraphics[width=\imgsize]{figures/teaser/#1/#2_depth.jpg}%
    & \includegraphics[width=\imgsize]{figures/teaser/#1/#2_opticalflow.jpg}%
    & \includegraphics[width=\imgsize]{figures/teaser/#1/#2_sceneflow.jpg}%
    & \includegraphics[width=\imgsize]{figures/teaser/#1/#2_shading.jpg}%
    & \includegraphics[width=\imgsize]{figures/teaser/#1/#2_albedo.jpg}%
}
\newcommand{\teaserrowasterix}[2]{%
    \includegraphics[width=\imgsize]{figures/teaser/#1/#2_.jpg}%
    & \includegraphics[width=\imgsize]{figures/teaser/#1/#2_next.jpg}%
    & \begin{tikzpicture}%
      \node[anchor=south west, inner sep=0] (image) at (0,0) {\includegraphics[width=\imgsize]{figures/teaser/#1/#2_semantic.jpg}};%
      \begin{scope}[shift={(image.north east)}]%
        \node[fill=white, font=\bfseries, inner sep=1pt] at (-.1,-.15) {*};%
      \end{scope}%
    \end{tikzpicture}%
    & \includegraphics[width=\imgsize]{figures/teaser/#1/#2_normal.jpg}%
    & \includegraphics[width=\imgsize]{figures/teaser/#1/#2_depth.jpg}%
    & \includegraphics[width=\imgsize]{figures/teaser/#1/#2_opticalflow.jpg}%
    & \includegraphics[width=\imgsize]{figures/teaser/#1/#2_sceneflow.jpg}%
    & \includegraphics[width=\imgsize]{figures/teaser/#1/#2_shading.jpg}%
    & \includegraphics[width=\imgsize]{figures/teaser/#1/#2_albedo.jpg}%
}
\newcommand{\OURS}{StableMTL}
\newcommand{\OURSSS}{$\text{StableMTL-}\mathcal{S}$}
\newcommand{\alltask}{\mathcal{T}}
\newcommand{\token}{T}
\newcommand{\tokenss}{\tau}

\newcommand{\tasklabel}{y}
\newcommand{\annotationset}{\mathcal{Y}}
\newcommand{\inputimg}{i}

\newcommand{\modelparams}{\phi}
\newcommand{\modelparamsss}{\theta}

\newcommand{\unet}{U_{\modelparams,\token}}
\newcommand{\unetsimple}{U_{\modelparams}}
\newcommand{\unetsimpless}{U_{\modelparamsss}}
\newcommand{\unetss}{U_{\modelparamsss\!,\tokenss}}
\newcommand{\mapset}[1]{\mathbb{M}^{#1}}

\newcommand{\vkitti}{VKITTI 2}

\newcommand{\kitti}{KITTI}
\newcommand{\diode}{DIODE}
\newcommand{\hypersim}{Hypersim}
\newcommand{\flying}{FlyingThings3D}
\newcommand{\cs}[0]{Cityscapes}

\newcommand{\depth}{\textbf{\textcolor{depth}{depth}}}
\newcommand{\normal}{\textbf{\textcolor{normal}{normal}}}
\newcommand{\semantic}{\textbf{\textcolor{semantic}{semantic}}}
\newcommand{\shading}{\textbf{\textcolor{shading}{shading}}}
\newcommand{\albedo}{\textbf{\textcolor{albedo}{albedo}}}
\newcommand{\opticalflow}{\textbf{\textcolor{opticalflow}{optical flow}}}
\newcommand{\sceneflow}{\textbf{\textcolor{sceneflow}{scene flow}}}
\newcommand{\optflow}{\textbf{\textcolor{opticalflow}{opt. flow}}}
\newcommand{\scflow}{\textbf{\textcolor{sceneflow}{sc. flow}}}

\newcommand{\Depth}{\textbf{\textcolor{depth}{Depth}}}
\newcommand{\Normal}{\textbf{\textcolor{normal}{Normal}}}
\newcommand{\Semantic}{\textbf{\textcolor{semantic}{Semantic}}}
\newcommand{\Shading}{\textbf{\textcolor{shading}{Shading}}}
\newcommand{\Albedo}{\textbf{\textcolor{albedo}{Albedo}}}
\newcommand{\Opticalflow}{\textbf{\textcolor{opticalflow}{Optical Flow}}}
\newcommand{\Sceneflow}{\textbf{\textcolor{sceneflow}{Scene Flow}}}

\newcommand{\SSemantic}{\textbf{\textcolor{semantic}{Seg.}}}
\newcommand{\NNormal}{\textbf{\textcolor{normal}{Normal}}}
\newcommand{\DDepth}{\textbf{\textcolor{depth}{Depth}}}
\newcommand{\OOpticalflow}{\textbf{\textcolor{opticalflow}{Opt.Flow}}}
\newcommand{\SSceneflow}{\textbf{\textcolor{sceneflow}{Sc.Flow}}}
\newcommand{\SShading}{\textbf{\textcolor{shading}{Sha.}}}
\newcommand{\AAlbedo}{\textbf{\textcolor{albedo}{Alb.}}}

\newcommand{\condenseparagraph}[1]{{\vspace*{0.1em}\noindent\textbf{#1}\quad}}

\usepackage[capitalize]{cleveref}
\crefname{section}{Sec.}{Secs.}
\Crefname{section}{Section}{Sections}
\Crefname{table}{Table}{Tables}
\crefname{table}{Tab.}{Tabs.}

\makeatletter
\DeclareRobustCommand\onedot{\futurelet\@let@token\@onedot}
\def\@onedot{\ifx\@let@token.\else.\null\fi\xspace}

\def\eg{\emph{e.g}\onedot} 
\def\ie{\emph{i.e}\onedot} 
\def\cf{\emph{cf}\onedot} 
\def\etc{\emph{etc}\onedot} \def\vs{\emph{vs}\onedot}

\newcommand{\stl}[1]{{\textcolor{gray}{#1}}}

\newcommand{\imgsize}{.125\textwidth}
 
\newcolumntype{C}[1]{>{\centering\let\newline\\\arraybackslash\hspace{0pt}}m{#1}}
\newcolumntype{S}[1]{@{\hspace{#1}}}
\newcolumntype{H}{>{\setbox0=\hbox\bgroup}c<{\egroup}@{}}

\newcommand{\verti}[2]{\multirow{1}{*}[#1]{\rotatebox[origin=c]{90}{#2}}}

\definecolor{best}{HTML}{C6EFCE}    %
\definecolor{second}{HTML}{FFF2CC}  %
\newcommand{\dataset}[1]{{\scriptsize\textcolor{gray}{#1}}}

\newcommand{\deltam}{\Delta_{\rm m}}
\newcommand{\mIoU}{mIoU}
\newcommand{\AbsRel}{AbsRel}
\newcommand{\MAE}{mAE}
\newcommand{\EPEtwoD}{EPE-2D}
\newcommand{\EPEthreeD}{EPE-3D}
\newcommand{\RMSE}{RMSE}

\definecolor{mappedRoad}{RGB}{128,64,128} %
\definecolor{mappedSky}{RGB}{70,130,180} %
\definecolor{mappedVegetation}{RGB}{107,142,35} %
\definecolor{mappedBuilding}{RGB}{70,70,70}
\definecolor{mappedPole}{RGB}{153,153,153}
\definecolor{mappedLight}{RGB}{250,170,30}    %
\definecolor{mappedSign}{RGB}{220,220,0}      %
\definecolor{mappedVehicle}{RGB}{0,0,142}
\definecolor{mappedIgnore}{RGB}{0,0,0}        %

\newcommand{\colorsquare}[1]{\colorbox{#1}{\rule{0pt}{1.ex}\rule{1.ex}{0pt}}}

\newcommand{\colorRoad}{\colorsquare{mappedRoad}}
\newcommand{\colorSky}{\colorsquare{mappedSky}}
\newcommand{\colorVegetation}{\colorsquare{mappedVegetation}}
\newcommand{\colorBuilding}{\colorsquare{mappedBuilding}}
\newcommand{\colorPole}{\colorsquare{mappedPole}}
\newcommand{\colorLight}{\colorsquare{mappedLight}}
\newcommand{\colorSign}{\colorsquare{mappedSign}}
\newcommand{\colorVehicle}{\colorsquare{mappedVehicle}}
\newcommand{\colorIgnore}{\colorsquare{mappedIgnore}}

\title{
\OURS{}: Repurposing Latent Diffusion Models\\for Multi-Task Learning from Partially Annotated Synthetic Datasets
}
\author{
  Anh-Quan Cao$^{1,2}$ \quad
	  Ivan Lopes$^2$ \quad
        Raoul de Charette$^2$
       \vspace{0.2cm}\\
	$^1$Valeo.ai \qquad $^2$Inria \vspace{0.2cm}\\[-0.05em]
    \href{https://github.com/astra-vision/StableMTL}{https://github.com/astra-vision/StableMTL}
}

\begin{document}

\twocolumn[{
\renewcommand\twocolumn[1][]{#1}
    \centering
    \vspace{-2em}
    \maketitle
    \vspace{-1.0em}
    \centering
    \footnotesize
    \setlength{\tabcolsep}{0pt}
    \renewcommand{\arraystretch}{0}

    \newcommand{\shift}{7.7ex}
    \newcommand{\prompt}{\cellcolor{gray!40}}
	\resizebox{\linewidth}{!}{
    \begin{tabular}{c@{\hspace{1.5pt}}p{1.8cm} H cS{.2ex}cS{.2ex}cS{.2ex}cS{.2ex}cS{.2ex}cS{.2ex}cS{.2ex}}
        & \makecell{Input} & \makecell{Input 2\\{\scriptsize (flow only)}} & \Semantic & \Normal & \Depth & \Opticalflow & \Sceneflow & \Shading & \Albedo
        \\

        \verti{5ex}{\scalebox{0.9}{\textbf{KITTI}}} & \includegraphics[width=\imgsize]{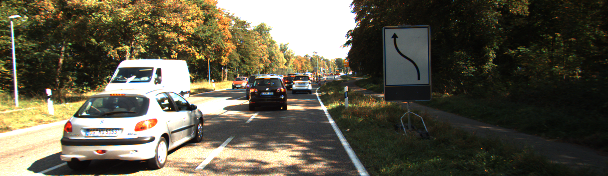}
        & \includegraphics[width=\imgsize]{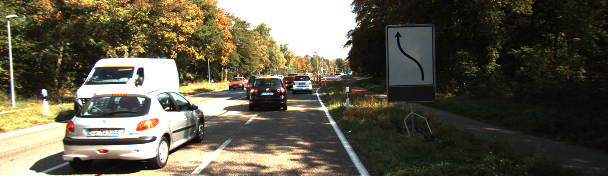}
        & \includegraphics[width=\imgsize]{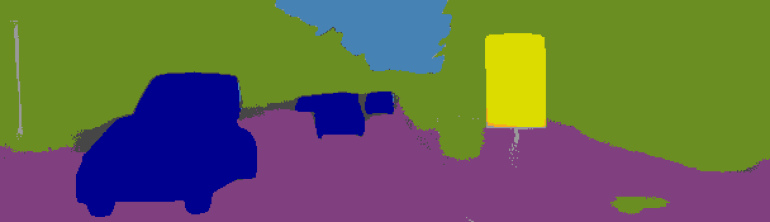}
        & \includegraphics[width=\imgsize]{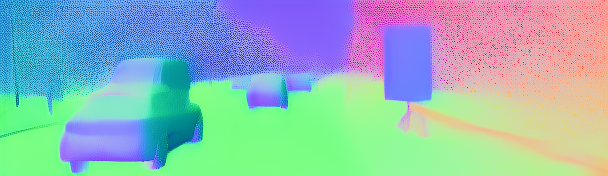}
        & \includegraphics[width=\imgsize]{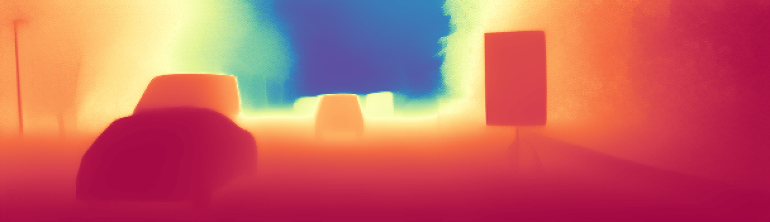}
        & \includegraphics[width=\imgsize]{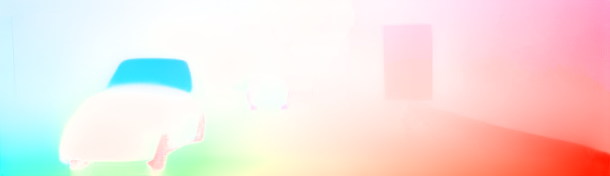}
        & \includegraphics[width=\imgsize]{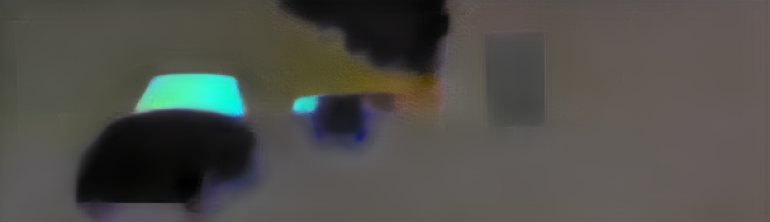}
        & \includegraphics[width=\imgsize]{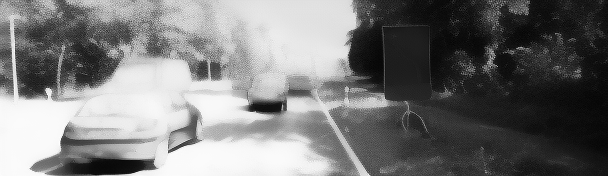}
        & \includegraphics[width=\imgsize]{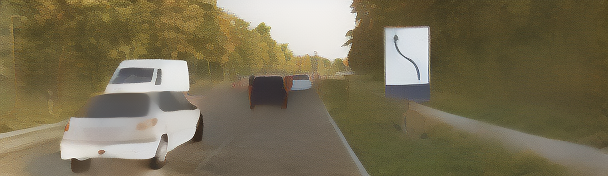}
       	\\ [0.5ex]

        \verti{9ex}{\scalebox{0.9}{\textbf{Waymo}}} & \includegraphics[width=\imgsize]{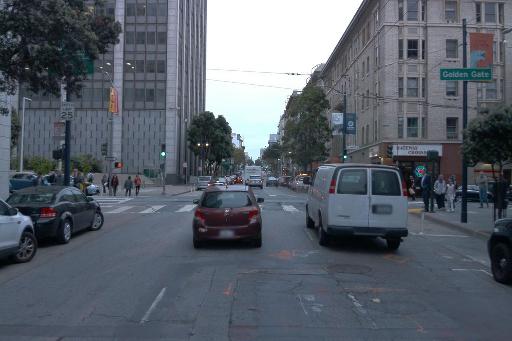}
        & \includegraphics[width=\imgsize]{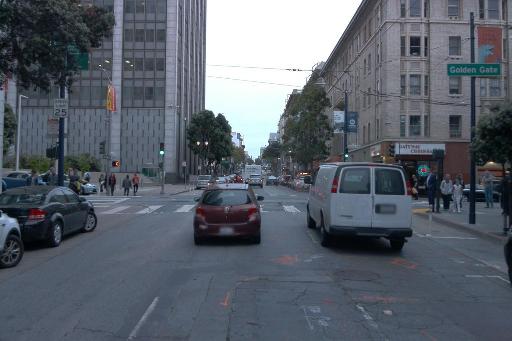}
        & \includegraphics[width=\imgsize]{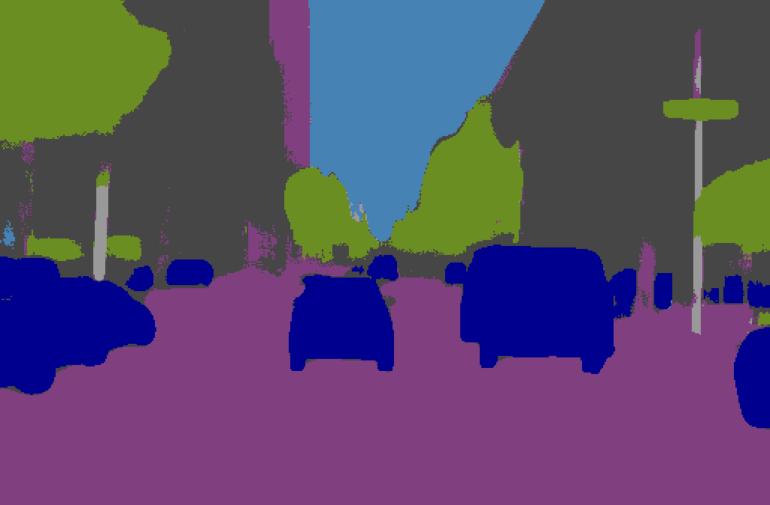}
        & \includegraphics[width=\imgsize]{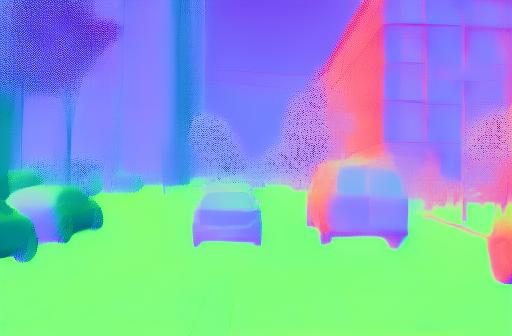}
        & \includegraphics[width=\imgsize]{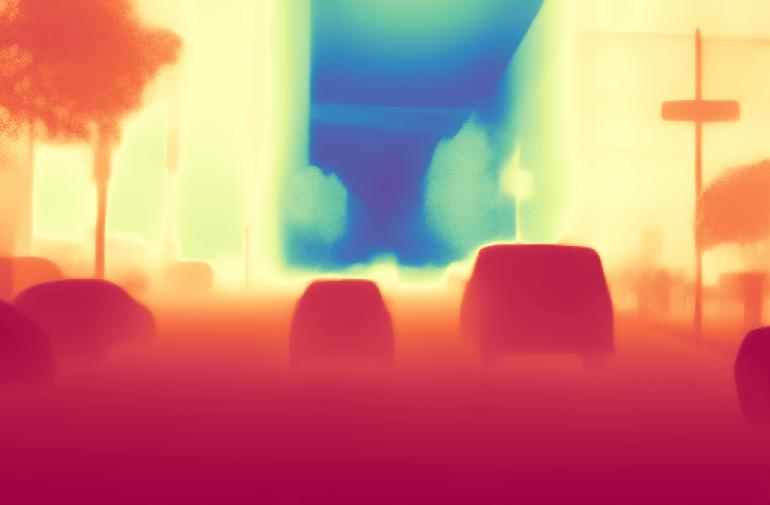}
        & \includegraphics[width=\imgsize]{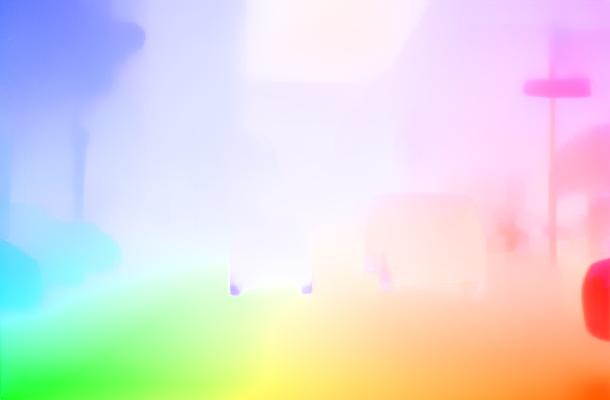}
        & \includegraphics[width=\imgsize]{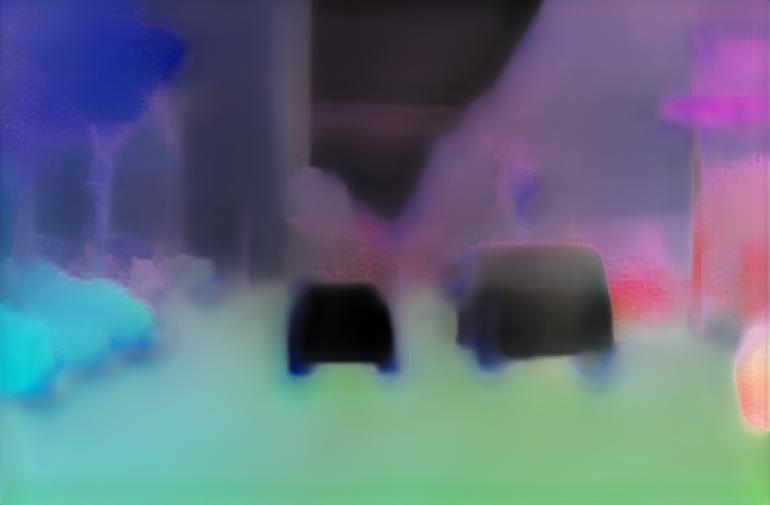}
        & \includegraphics[width=\imgsize]{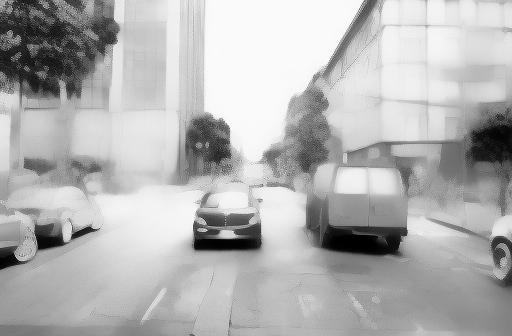}
        & \includegraphics[width=\imgsize]{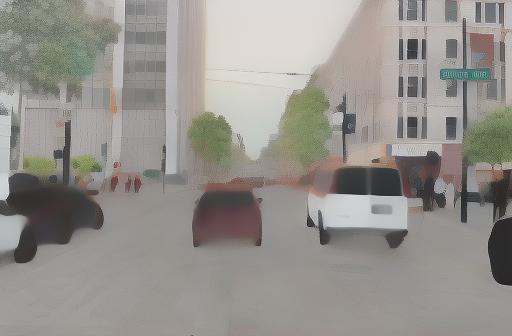}

        \\ [0.5ex]
        \verti{8.5ex}{\scalebox{0.9}{\textbf{Cityscapes}}}
        & \includegraphics[width=\imgsize]{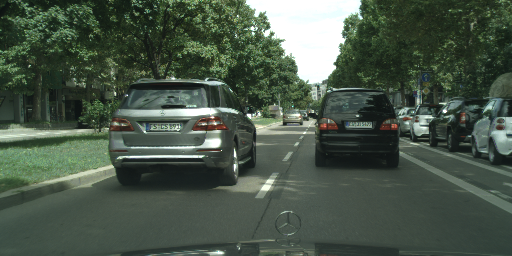}
        & \includegraphics[width=\imgsize]{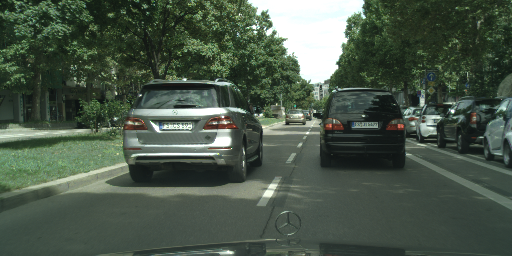}
        & \includegraphics[width=\imgsize]{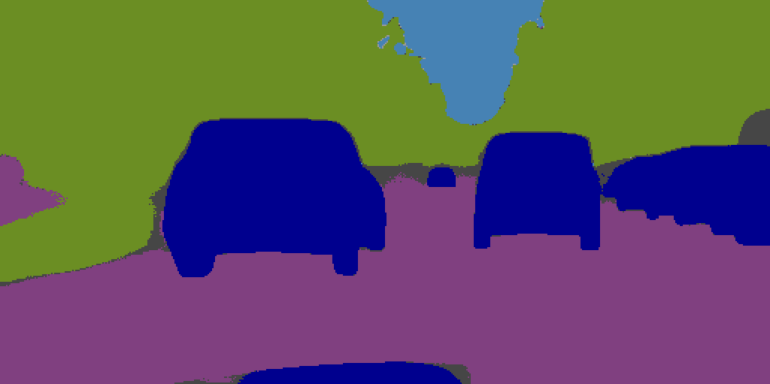}
        & \includegraphics[width=\imgsize]{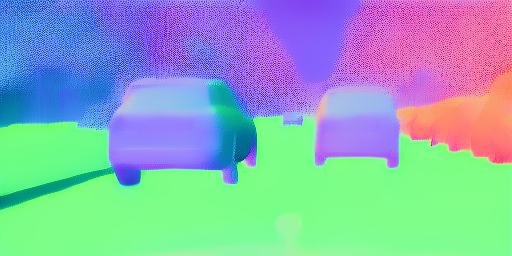}
        & \includegraphics[width=\imgsize]{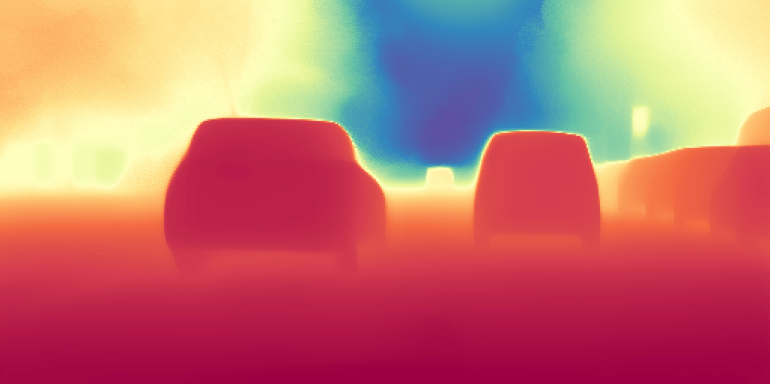}
        & \includegraphics[width=\imgsize]{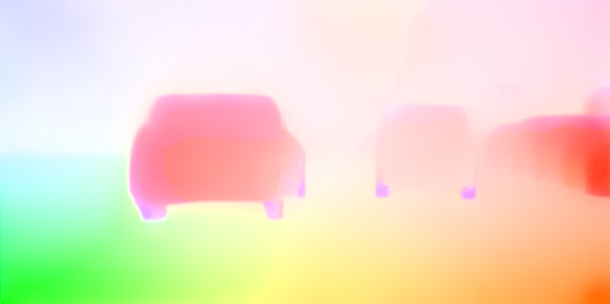}
        & \includegraphics[width=\imgsize]{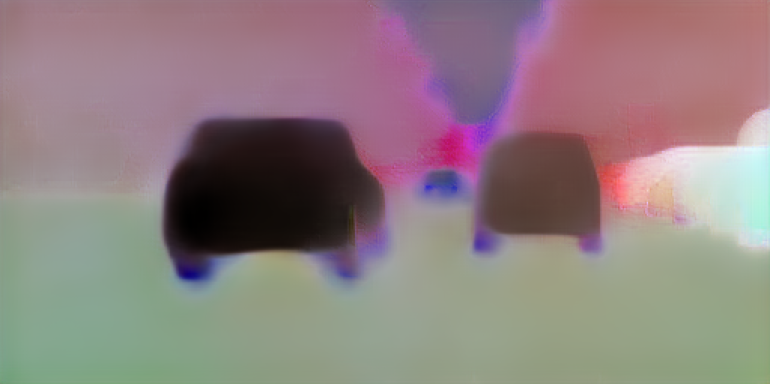}
        & \includegraphics[width=\imgsize]{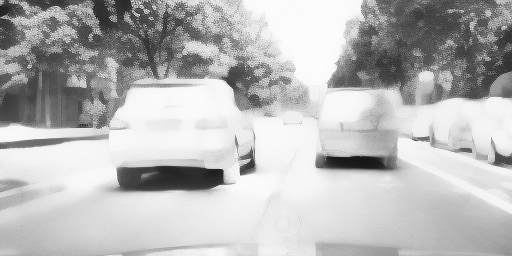}
        & \includegraphics[width=\imgsize]{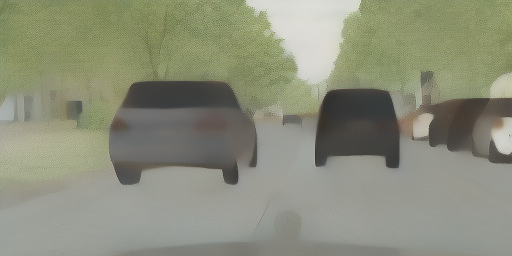}
        \\ [0.5ex]

        \verti{9ex}{\scalebox{0.9}{\textbf{DDAD}}}
        & \teaserrow{ddad}{val_35_000012_0}
        \\
        [0.5ex]

        \verti{7ex}{\scalebox{0.9}{\textbf{Youtube-VOS}}}

        & \teaserrowasterix{youtubevos}{1a894a8f98_00020}\\[0.2ex]

        & \teaserrowasterix{youtubevos}{a9af84b17b_00075}\\

        \verti{8ex}{\scalebox{0.9}{\textbf{DAVIS}}}
        & \teaserrowasterix{davis}{motocross-jump_00028}\\

    \end{tabular}}
    \vspace{-0.7em}
    \captionof{figure}{\textbf{Results on unseen data.} StableMTL demonstrates robust generalization capability despite being trained on partially labeled synthetic datasets.
    *Note that \semantic{} is trained on closed-set driving classes although it generalizes to some extent to other domains.
    \vspace{0.8em}
    }
    \label{fig:teaser}
}]

\begin{abstract}
Multi-task learning for dense prediction is hindered by the need for annotations for every single task, although recent works explore training with partial task labels.
Leveraging diffusion models’ generalization capability, we extend partial learning to domain generalization, training a multi-task model on synthetic datasets, each only labeled for a subset of tasks.
Our method, StableMTL, repurposes an image generator for latent regression, adapting a denoising framework with task encoding, conditioning, and a tailored scheme. Instead of per-task losses requiring careful balancing, our unified latent loss scales to more tasks.
To encourage inter-task synergy, we use a multi-stream model with task-attention that converts N-to-N interactions into efficient N-to-one attention, promoting cross-task sharing. StableMTL outperforms baselines on 7 tasks across 8 benchmarks.
\end{abstract}

\section{Introduction}

Multi-task learning (MTL) is crucial for computer vision systems that interacting with our world, such as robotics or virtual reality, which often require estimating various scene cues (\eg, semantic segmentation, depth, optical flow, \etc) in a time-sensitive manner. Beyond practical aspects, a vast body of literature shows that solving different tasks leads to learning (at least partially) distinct representations that can benefit other tasks~\cite{zamir2018taskonomy,standley2020taskslearnedmultitasklearning}. This makes MTL an ideal setting to enforce information exchange across tasks~\cite{deltam}. However, the benefits of MTL are impeded by the limited number of datasets with overlapping task labels, especially for dense predictions requiring costly pixel-level annotations~\cite{nishi2024jtr}.

An alternative is to train in a partially labeled setup~\cite{MTPSL}, using annotations for only some tasks, which differs from prior semi-supervised settings~\cite{ouali2020semisupervisedsemanticsegmentationcrossconsistency,ghiasi2021multitaskselftraininglearninggeneral} that typically overlook task relationships. Yet, current partially annotated MTL methods~\cite{MTPSL,diffusionmtl,nishi2024jtr} face two main limitations: (i) single-domain training restricting generalization, and (ii) conflicting task objectives needing careful balancing.

For better generalization, one can leverage multiple existing real-world datasets with partial labels for training{. H}owever, dense annotations are often noisy or {unavailable}. For instance, depth data commonly suffer from sensor noise, while tasks such as albedo and shading are {hard to obtain}. Recent works \cite{marigold, he2024lotus, genpercept} instead demonstrate the benefits of training exclusively on synthetic datasets, providing dense, accurate, consistent labels while generalizing to real-world data when leveraging pre-trained diffusion model priors~\cite{rombach2022high}. By preserving the rich latent space of diffusion models and fine-tuning only {its denoiser UNet}, pre-trained latent diffusion models (LDMs) can be repurposed for dense prediction tasks~\cite{marigold, he2024lotus, martingarcia2024diffusione2eft, genpercept} with strong real-world generalization, even when trained on relatively small synthetic datasets.

Inspired by this line of work, we propose a novel setting that leverages multiple synthetic datasets from diverse domains with partial labels, thereby reducing reliance on extensive annotations.
To address this, we introduce \OURS{}, which extends deterministic single-step LDMs \cite{he2024lotus} to a partially labeled multi-task setting with task-token conditioning.
A major benefit of our method is that it trains all tasks with a unified MSE loss, eliminating complex hyperparameter tuning and enabling more tasks.
To achieve {a stable training}, we introduce a task-gradient isolation mechanism that addresses the notable gradient interference problem in MTL~\cite{deltam} and further encourages cross-task information exchange through an efficient N-to-{one} multi-task attention mechanism. As a result, we demonstrate qualitative and quantitative improvements across seven tasks and multiple real-world datasets.
Our methodological contributions are twofold:
\begin{itemize}
  \setlength\itemsep{0.2pt}
  \setlength\topsep{0.5pt}
    \item We introduce a more realistic partially labeled MTL setting, training on multiple synthetic datasets, each with partial task annotations.

    \item {We repurpose a pre-trained image generation LDM for discriminative MTL.} We propose a multi-stream architecture that extends the Stable Diffusion architecture with N-to-{one} attention, enabling effective cross-task knowledge transfer. Through a unified latent loss and tailored training scheme, our approach eliminates task-specific losses and complex inter-task balancing.

\end{itemize}
As shown in~\cref{fig:teaser}, while training on three relatively small-scale synthetic datasets ($\approx{}$80k images), \mbox{\OURS{}} generalizes to many real-world scenarios, including unseen structures and strongly out-of-distribution domains (\eg, DAVIS and Youtube-VOS).
{Furthermore}, it outperforms existing partially labeled MTL methods on {seven} tasks, across {eight} benchmarks, with a +83.54 $\deltam$ overall improvement.

\section{Related works}

\condenseparagraph{Multi-task learning (MTL).}
Given the large body of literature on MTL, including on dense predictions~\cite{deltam}, we focus on the most relevant techniques. Recent adapter-based methods~\cite{liu2022polyhistor, liang2022effective} leverage large pre-trained models but typically rely on discriminative priors. We are the first to utilize a generative prior for MTL. Traditional approaches often require complex mechanisms like adaptive loss weighting~\cite{kendall2018multi}, gradient manipulation~\cite{chen2018gradnorm, chen2020just, yu2020gradient, senushkin2023independent}, or Pareto optimization~\cite{momma2022multi, lin2019pareto} to balance tasks. Our method achieves this balance naturally through our tailored training scheme and an unified latent loss, eliminating such complexities.

{Enhancing task exchange~\cite{zamir2018taskonomy} is another key challenge, often addressed via consistency losses~\cite{lu2021taskology}, or joint task spaces~\cite{nishi2024jtr, MTPSL}.} {Such approaches can benefit out-of-distribution performance~\cite{zamir2020robust, saha2021learning}. However, they heavily rely on pairwise interactions and thus scale poorly.} {Other approaches like DeMT~\cite{DeMT}, InvPT++~\cite{InvPT}, and BridgeNet~\cite{bridgenet} simplify quadratic task interactions by learning task-generic features.} {Our method differs significantly as we adapt the base network by introducing only a minimal number of parameters and use a two-stage process to train our pipeline.
Unlike prior methods which require full annotation sets,} {StableMTL operates with partial annotations across multiple datasets. Finally, while task switching mechanisms~\cite{taskswitching,tca,pgt} and token-based} {conditioning in Stable Diffusion have been explored in past works~\cite{rgbx}, we extend this concept to a more general multi-task learning setting with a broader range of tasks.}

\begin{figure*}
    \centering
    \includegraphics[clip,width=0.95\linewidth]{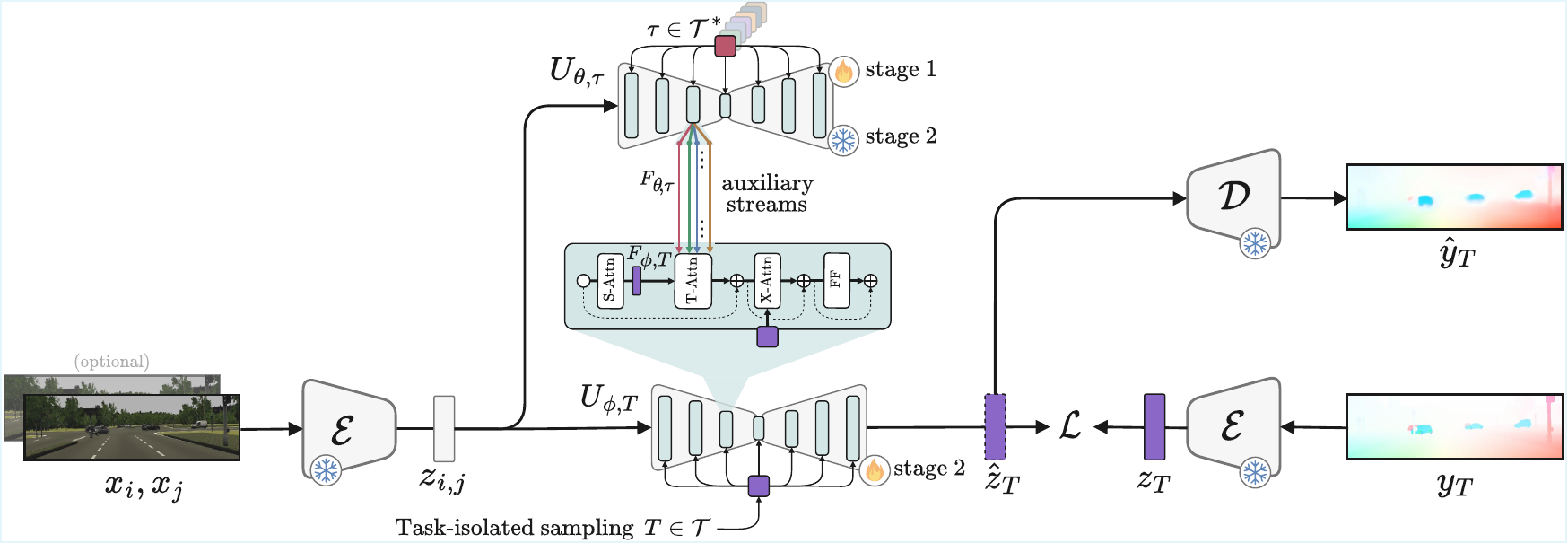}
    \caption{\textbf{\OURS{}.} Our pipeline has two training stages.
    The first stage (\cref{sec:singlestream}) fine-tunes a UNet ($\unetss$) to predict target annotation latents from input image latents, conditioned on multi-task tokens sampled via our training scheme to isolate task gradient. To encourage task exchanges, the second stage (\cref{sec:multistream}) uses a multi-stream architecture combining the frozen single-stream UNet trained during Stage 1 and a trainable \emph{main} UNet ($\unet$).
    It further benefits from a novel task attention mechanism which enriches the main UNet's features with task-specific information from its auxiliary counterparts. Both stages are trained with a single MSE latent loss, instead of task-specific losses.
    }
    \label{fig:pipeline}
\end{figure*}

\condenseparagraph{MTL with partial annotation.} Early efforts addressed partially labeled shallow models~\cite{liu2007semi, zhang2009semi, wang2009semi}, subsequent deep learning approaches like MTPSL~\cite{MTPSL}, DejaVu~\cite{dejavu}, DiffusionMTL~\cite{diffusionmtl}, and JTR~\cite{nishi2024jtr} have predominantly relied on simulating missing labels within single, fully-annotated datasets. Refer to Fontana \textit{et al.}~\cite{fontana2024multitask} for a comprehensive review.
This setup not only fails to fully capture real-world complexities where annotations are naturally sparse across different data sources but also {relies} on the use of real-world datasets which are expensive to annotate. SemiMTL~\cite{semiMTL} {explores multi-dataset training, but is restricted to two tasks and uses real data.} Motivated by these gaps and the prohibitive cost of annotating numerous tasks on real-world data, we propose a more realistic, scalable, and challenging scenario: learning a greater number of tasks from multiple synthetic datasets, each with naturally partially overlapping sets of task annotations.

\condenseparagraph{Repurposing LDMs for dense predictions.}
Latent diffusion models are known for robust representations and were first repurposed by Marigold~\cite{marigold, marigoldjournal} for depth estimation, demonstrating strong generalization from small synthetic datasets.
Subsequent efforts extended this to other individual tasks such as geometry~\cite{geowizard,stablenormal}, motion~\cite{stablemotion} and also improved efficiency with enhanced sampling~\cite{depthfm} or single-step latent regression~\cite{he2024lotus,martingarcia2024diffusione2eft,genpercept}. {In} the LDM context, holistic MTL remains largely unexplored. Recently, large-scale LDM-based models like Diception~\cite{zhao2025diception} and OneDiffusion~\cite{onediffusion} emerged, handling diverse tasks by training on large-scale real-world data.
However, their primary focus is not on holistic MTL capabilities when restricted to synthetic-only training or small-scale, partially labeled data.
Our approach is complementary, offering a method to fine-tune these models for multi-task using only small-scale, partially labeled synthetic datasets.

\section{\OURS{}}
\label{sec:method}

We address the problem of multi-task partially supervised learning~\cite{MTPSL} by extending it to a more realistic setting where \emph{partial} labels are obtained from \emph{multiple synthetic} datasets. In summary, our aim is to learn $K$ tasks from a {set} of $N$ synthetic datasets, where each dataset {only provides} access to labels for fewer than $K$ tasks.

Our method, coined \OURS{}, builds on recent advances in diffusion models repurposed for dense task estimation~\cite{he2024lotus} allowing us to generalize to the real domain. Importantly, we frame multi-task learning as a latent regression problem thus \textit{alleviating the need for task-specific losses and weighting}.
To train our complete architecture, depicted in \cref{fig:pipeline}, we follow a two-stage process.
In the first stage, detailed in \cref{sec:singlestream}, we fine-tune a latent diffusion model repurposed for dense prediction~\cite{he2024lotus}, which we adapt to multi-task learning by incorporating task tokens, a unified latent loss and a custom training scheme.
In the second stage, {we encourage task exchange by training a separate task-conditioning model where interactions are modeled using} a task-attention mechanism.
A byproduct of our unified latent loss and our careful design choices is that \OURS{} can scale to various spatial and temporal tasks without the usual need for cumbersome task balancing~\cite{fontana2024multitask}.

We detail the group of datasets used for training in~\cref{tab:training-datasets}.

\begin{table}[h]
    \setlength{\tabcolsep}{2pt}
    \centering

    \newcommand{\collabel}[1]{\rotatebox[origin=c]{35}{#1}}

    \resizebox{\linewidth}{!}{
    \begin{tabular}{l|HccH| ccccccc}
        \toprule
        Dataset & Domain & type & size & resolution
        & \collabel{\Semantic}
        & \collabel{\Normal}
        & \collabel{\Depth}
        & \collabel{\Opticalflow}
        & \collabel{\Sceneflow}
        & \collabel{\Shading}
        & \collabel{\Albedo}\\
        \midrule
        Hypersim~\cite{hypersim} & Synthetic & Indoor &  39K & 288x384 & & \checkmark & \checkmark & & & \checkmark & \checkmark\\
        Virtual KITTI 2~\cite{virtualkitti2}    & Synthetic & Urban & 20K & 187x621 & \checkmark & \checkmark & \checkmark & \checkmark & \checkmark & \\
        Flying Things 3D~\cite{flyingthings3d}  & Synthetic & Objects & 20K & 268x480 &  & & &  \checkmark & \checkmark & & \\
        \bottomrule
    \end{tabular}}
    \caption{\textbf{Training set.} {Our ensemble of three synthetic datasets covers seven tasks. These include: high-level \semantic{} segmentation; two geometrical tasks, namely \normal{} and \depth{}; two dynamic flow-based estimations, \opticalflow{} (pixel displacement in screen-space) and \sceneflow{} (3D displacement in camera-space); and two low-level intrinsic component estimations, \shading{} (grayscale irradiance for lighting) and \albedo{} (diffuse material reflectance).}
    }
    \label{tab:training-datasets}
\end{table}

\subsection{Single-stream architecture}
\label{sec:singlestream}

{Considering pixel-wise predictions}, we denote $\mapset{} = \mathbb{R}^{H \times W}$ for brevity. Let $x_\inputimg \in \mapset{3}$ be an input image and $\tasklabel_{\tokenss} \in \annotationset_{\tokenss}$ the corresponding annotation for a task $\tokenss$ from a {task set} $\alltask$. Our {problem} revolves around a latent space $\mathcal{Z} \subset \mapset{4}$ in which we perform multi-task learning. We employ the pretrained StableDiffusion~\cite{rombach2022high} VAE consisting of a frozen encoder $\Enc: \mathbb{M}^3 \to \mathcal{Z}$ and a decoder $\Dec: \mathcal{Z} \to \mapset{3}$ for the inverse operation.
An image latent is computed as $z_{\inputimg} = \Enc(x_\inputimg)$. {Since} annotations $\tasklabel_{\tokenss}$ are of different nature (\eg, \semantic{} encodes discrete class indices while \depth{} is continuous), {those} must first be processed by a task-specific function $f_{\tokenss}: \annotationset_{\tokenss} \to \mapset{3}$.
As a result, task latents are obtained as: $z_{\tokenss} = \Enc{}(f_{\tokenss}(\tasklabel_{\tokenss}))$. A UNet model, {noted} $\unetsimpless$, parametrized by $\modelparamsss$, is fine-tuned to predict the task latent $\hat{z}_{\tokenss}$ from the image latent $z_{\inputimg}$ in a single step, as $\hat{z}_{\tokenss} = \unetsimpless(z_{\inputimg})$. For tasks requiring multi-frame inputs -- such as flow estimations -- we extend our pipeline to receive multiple input frames.
{Then, the decoder $\Dec$ is used} to map latent representations back to the RGB space. {Finally, predictions} $\hat{\tasklabel}_{\tokenss}$ require a post-processing step $p_{\tokenss}\,: \mapset{3} \to \annotationset_{\tokenss}$ after being decoded, resulting in $p_{\tokenss}(\Dec{}(\hat{z}_{\tokenss}))$.

{We build upon the single-step deterministic diffusion formulation \cite{he2024lotus}, which offers efficient training and inference while maintaining strong generalization compared to multi-step approaches \cite{marigold}.} {To extend the framework to the multi-task setting, we utilize task tokens. Initially introduced for dense prediction \cite{taskswitching} and later adapted for diffusion models~\cite{rgbx}, it serves as a unifying mechanism to enable a single network to handle multiple tasks jointly.
Each task token conditions the model to learn a different distribution mode, \ie, tasks, and is facilitated by the use of cross-attention with the model's hidden states. This approach ensures a fully shared representation at no additional parameter cost.}

The UNet predicts the task latent as $\hat{z}_{\tokenss} = \unetsimpless(z_{\inputimg}, c_{\tokenss})$, written $\unetss(z_{\inputimg})$ for short.
 Thus, all tasks share the same parameters $\theta$, with {task conditionings provided via tokens $c_{\tokenss}$}. Compared to prior works~\cite{rgbx}, we demonstrate that the use of task tokens can be scaled to a much broader setting multi-task and multi-dataset training.

\condenseparagraph{Multi-frame adaptation.}
{Certain} tasks such as flow estimation can be solved from monocular inputs~\cite{aleotti2021learning,argaw2021optical} but are better conditioned using a pair of temporal frames ($i$, $j$). To accommodate such spatio-temporal tasks, \OURS{} takes as input a channel-wise concatenated pair of latents, $z_{i,j} = \texttt{concat}(z_{i}, z_{j})$. The model then predicts the output as $\hat{z}_{\tokenss} = \unetss(z_{i,j})$. For single-frame tasks (\eg, \semantic{}, \depth{}), we simply set $j=i$, thereby concatenating $z_{i}$ with itself. This simple scheme provides a basis for handling both spatial and spatio-temporal tasks in a unified framework.

\condenseparagraph{Unified MSE latent loss.} For each task, we minimize the Mean Squared Error (MSE) loss {in latent space} between the predicted latent $\hat{z}_{\tokenss}$ and ground-truth latent $z_{\tokenss}$:
\begin{equation} \label{eq:mse_loss}
    \mathcal{L}(\modelparamsss) = \| \hat{z}_{\tokenss} - z_{\tokenss} \|_2^2 = \| \unetss(z_{i,j}) - \Enc(f_{\tokenss}(\tasklabel_{\tokenss})) \|_2^2\,.
\end{equation}
The loss naturally mitigates task imbalance because it is computed in a unified latent space shared across all tasks, and averaging over task tokens provides inherent normalization across heterogeneous tasks and resolutions. {This design enables scaling to a broader set of tasks without task-specific losses or complex balancing heuristics, avoiding costly hyperparameter tuning.}

\condenseparagraph{Task-gradient isolation scheme.}
Although the unified latent loss reduces task imbalance, some tasks still exhibit larger gradient magnitudes and conflicting directions, as {observed in traditional frameworks~\cite{fontana2024multitask}. This causes tasks} with weaker gradients to be dominated by stronger ones. To mitigate this, we adopt a task-isolation training scheme in which each training step uses a mini-batch from a single task. During gradient accumulation, gradients are accumulated only over mini-batches of the same task, after which an optimizer step is taken and gradients are reset before proceeding to the next task. While this scheme enforces a single task per batch, {stochasticity will sample the same image} with different task labels over time, ensuring exposure to all annotations. Combined with the unified latent loss, it alleviates task imbalance and removes the need for complex task balancing.

\subsection{Multi-stream architecture}
\label{sec:multistream}

The above single-stream architecture can handle multiple tasks simultaneously, but does not incorporate any explicit tasks exchange mechanisms. Prior works have addressed this by training N tasks jointly with an N-to-N multi-task attention mechanism, where each task attends to all others to learn task interactions~\cite{bruggemann2021atrc,densemtl}. Such strategy, however, scales poorly with the number of tasks. Therefore, we propose a multi-stream architecture, where the main stream attends tasks, processed via the single-stream (\cref{sec:singlestream}), making use of a more scalable {N-to-one} attention.

\condenseparagraph{Architecture.} Our multi-stream architecture, depicted in \cref{fig:pipeline}, extends the single-stream architecture {from \cref{sec:singlestream} by duplicating the UNet}.
This \emph{main} trainable UNet, denoted $\unet$, is responsible {for the final output, and is also} conditioned on a \emph{main} task token $\token$.
The single-stream UNet trained in \cref{sec:singlestream}, denoted $\unetss$, is kept frozen and referred as \textit{auxiliary} since it generates task-specific {features which are beneficial} for solving the main task.
During inference, $\unetss$ produces auxiliary features for all tasks $\tokenss$, other than the main task $\token$. The main UNet $\unet$ then predicts the output for $\token$, while attending to these features. {Model $\unet$ has its weights initialized from $\theta$ and trained similar to \cref{sec:singlestream}.}

\begin{figure}[h]
    \centering
    \includegraphics[trim={3ex 0 0 0},clip,width=\linewidth]{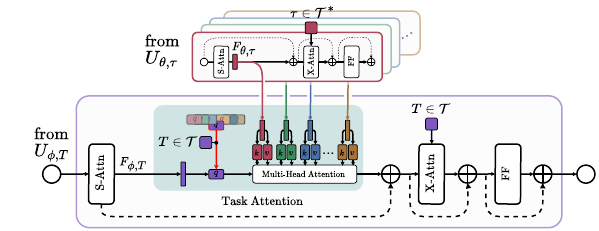}
    \caption{\textbf{Proposed task attention.} In addition to standard spatial and cross-attention mechanisms, our transformer blocks in the main UNet incorporate multi-stream information from auxiliary tasks. This is achieved by connecting the dedicated frozen single-stream UNet ($\unetss$) to the main UNet ($\unet$), providing the latter with auxiliary features. $\unetss$ is kept frozen.
    }
    \label{fig:task-attn}
\end{figure}

\condenseparagraph{Task attention.} The UNet is comprised of stacked convolutional residual blocks and transformer blocks. Each transformer block contains a spatial attention layer (S-Attn), a cross-attention layer (X-Attn), and a feed-forward network (FF). Spatial attention aggregates information from correlated 2D spatial locations, while cross-attention conditions the features on the task token. In the main UNet, $\unet$, we introduce a task-attention layer within each transformer block, placed immediately after the spatial (self-)attention layer. This new layer enriches the main stream features by attending to relevant task-specific features from the auxiliary streams, as depicted in \cref{fig:task-attn}.

We define $F_{\modelparams,\token}$ as the feature map from the spatial attention layer in the main UNet $\unet$, conditioned on task $\token$. {Given $\alltask^*{=}\alltask \setminus \{\token\}$, let} $\{F_{\modelparamsss,\tokenss} \mid \tokenss \in \alltask^*\}$ be the set of feature maps from the frozen auxiliary UNet $\unetss$ (the auxiliary streams), each conditioned on an auxiliary task token $\tokenss$.
The cross-attention {between the main stream and} the auxiliary streams is computed as
$\operatorname{Attention}(Q,K,V) = \operatorname{softmax}(QK^{\!\top}/\sqrt{d})V,$
where
\begin{align}
Q &=\bigl[q_{\token}\bigl(F_{\modelparams,\token}\bigr)\bigr], \\(K,V) &= \bigl[(k_{\tokenss}(F_{\modelparamsss,\tokenss}),
              v_{\tokenss}(F_{\modelparamsss,\tokenss}))
        \mid \tokenss \in \alltask^* \bigr],
\end{align}
and $d$ is the feature dimension. Each task $t$ has its own projection layers $(q_t,k_t,v_t)$, enabling task-specific adaptation. The brackets $[\cdot]$ denote concatenation along a newly introduced \emph{task} dimension.

\condenseparagraph{Attention-guided task masking.} As previously mentioned, both the main UNet $\unet$ and the auxiliary UNet $\unetss$ are initialized with {the} weights from our single-stream {stage}. This shared initialization can cause the task attention mechanism to concentrate heavily
{on certain auxiliary tasks early during training}. To promote broader exploration and prevent attention saturation, we implement attention-guided task masking within each task-attention layer.

For each spatial location (\ie, "image token") in every task-attention layer, $(\mathbf{s}_{\token\rightarrow \tokenss})_{\tokenss \in \alltask^*}$ denotes the vector of attention scores from the main stream (task $\token$) to the auxiliary streams (tasks $\tokenss \in \alltask^*$). These scores are computed as components of $\operatorname{softmax}(QK^\top/\sqrt{d})$
and represent the attention assigned by task $\token$ to auxiliary task $\tokenss$.
Intuitively, tasks with higher scores should have a higher chance of being masked. We model this with the distribution $\pi_{\token}$ over auxiliary tasks $\alltask^*$:
    $\pi_{\token}(\tokenss) = \frac{\mathbf{s}_{\token\rightarrow \tokenss}}{\sum_{{\tokenss}\in \alltask^*}{\mathbf{s}_{\token \rightarrow {\tokenss}}}}$.

To select a task for masking, we proceed by sampling $m_{\token} \sim \text{Sample}(\pi_{\token})$ which results in per-task masks: $M(t) = \mathbbm{1}[t \neq m_{\token}]$, where $\mathbbm{1}$ is the indicator function. This mask is then applied when computing the  attention score $\{M(\tokenss) *  \mathbf{s}_{\token\rightarrow \tokenss} |\,\tokenss \in \alltask^* \}$.
{To further encourage exploration,} we randomly apply the masking procedure with probability $\rho$. We later ablate the impact of $\rho$ and alternative sampling strategies. {This ensure various tasks interact with each other and prevents over-reliance on dominant tasks.}

\section{Experiments}
\label{sec:exps}

\condenseparagraph{Datasets.} {We train on seven tasks using the three synthetic datasets in~\cref{tab:training-datasets}:~\hypersim{}~\cite{hypersim},~\mbox{\vkitti}~\cite{virtualkitti2}, and~\flying{}~\cite{flyingthings3d}.
At each training step, we uniformly sample a task. If the selected task is labeled in multiple datasets, we sample from a specific dataset according to the following: for \depth{} and \normal{}, we use the sampling of Marigold~\cite{marigold}; for \opticalflow{} and \sceneflow{}, we sample~\flying{} and~\vkitti{} equally.

{Real images are used for evaluation.} \Normal{} is evaluated on DIODE~\cite{diode}; \depth{} on KITTI~\cite{kitti} and DIODE~\cite{diode}; \semantic{} on Cityscapes~\cite{cityscapes} with the class mapping from~\cite{densemtl}; \opticalflow{} and \sceneflow{} on KITTI flow~\cite{kittiflow}; lastly we use MIDIntrinsics~\cite{murmann19mid,careaga2023Intrinsic} for \shading{} and \albedo{}.

{For closer comparison, we also conduct experiments on NYU-v2~\cite{Silberman2012nyud} with the setting of MTPSL~\cite{MTPSL} where "\emph{One}" keeps exactly one task label per image, while "\emph{Random}" uses between one and all-but-one labels per image.}

\begin{table*}[t]
    \centering
    \setlength{\tabcolsep}{2pt}
    \resizebox{.75\textwidth}{!}{
    \begin{tabular}{l| c c c c c c c c|c}
        \toprule
        \multirow{3}{*}{\textbf{Method}} &
        \Semantic &
        \Normal  &
        \multicolumn{2}{c}{\Depth} &
        \textbf{\textcolor{opticalflow}{Opt. Flow}} &
        \textbf{\textcolor{sceneflow}{Sc. Flow}} &
        \Shading &
        \Albedo &
        \textbf{MTL Perf.} \\

        \cmidrule(lr){2-2} \cmidrule(lr){3-3} \cmidrule(lr){4-5} \cmidrule(lr){6-6}
        \cmidrule(lr){7-7} \cmidrule(lr){8-8} \cmidrule(lr){9-9} \cmidrule(lr){10-10}

        & {\footnotesize \mIoU{} \%$\uparrow$}
        & {\footnotesize \MAE{} °$\downarrow$}
        & {\footnotesize \AbsRel{} \%$\downarrow$}
        & {\footnotesize \AbsRel{} \%$\downarrow$}
        & {\footnotesize \EPEtwoD{} px$\downarrow$}
        & {\footnotesize \EPEthreeD{} m$\downarrow$}
        & {\footnotesize \RMSE{}$\downarrow$}
        & {\footnotesize \RMSE{}$\downarrow$}
        & {\footnotesize $\deltam{} \%\uparrow$} \\

        & \dataset{Cityscapes} & \dataset{DIODE} & \dataset{KITTI} & \dataset{DIODE} & \dataset{KITTI} & \dataset{KITTI} & \dataset{MID} & \dataset{MID} & \dataset{Average} \\

        \midrule
        Single-task baseline & \stl{48.17} & \stl{22.27} & \stl{14.21} & \stl{32.56} & \stl{10.36} & \stl{0.2735} & \stl{0.2145} & \stl{0.2551} & \stl{0.00}\\

        \addlinespace

        JTR~\cite{nishi2024jtr} & 45.75 & 40.23 & 26.39 & 66.39 & \stl{n/a} & \stl{n/a} & \stl{n/a} & \stl{n/a} & -- \\
        JTR*~\cite{nishi2024jtr} & 20.46 & 50.91 & 39.27 & 73.14 & 34.92 & 0.5176 & 0.3030 & 0.3565 & --106.87 \\

        DiffusionMTL~\cite{diffusionmtl} & 30.09 & 35.64 & 21.10 & 45.19 & \stl{n/a} & \stl{n/a} & \stl{n/a} & \stl{n/a} & -- \\
        DiffusionMTL*~\cite{diffusionmtl}       & 45.92    & 44.56    & 24.83 & 58.17 & 36.60 & 0.3502 & 0.3004 & 0.3660 & --78.76 \\

        \textbf{\OURSSS{}} & \cellcolor{second}52.57 & \cellcolor{second}23.94 & \cellcolor{second}15.64 & \cellcolor{second}33.36 & \cellcolor{second}12.76 & \cellcolor{second}0.2618 & \cellcolor{best}0.2310 & \cellcolor{second}0.2077 & \cellcolor{second}--1.57 \\
        \textbf{\OURS} & \cellcolor{best}55.79 & \cellcolor{best}23.27 & \cellcolor{best}14.98 &  \cellcolor{best}33.03  & \cellcolor{best}10.76 & \cellcolor{best}0.2313  & \cellcolor{second}0.2346 & \cellcolor{best}0.2016  & \cellcolor{best}+4.78 \\

        \bottomrule
    \end{tabular}
    }
    \caption{\textbf{Multi-task results}. \OURS{} significantly expands task coverage, addressing over twice the tasks of original baselines, while consistently outperforming them. \colorbox{best}{best} and \colorbox{second}{second-best} are highlighted.
    *baselines adapted to full setting
    }
    \label{tab:main-results}
\end{table*}

\begin{figure*}[!h]
    \centering
    \footnotesize
    \setlength{\tabcolsep}{0pt}
    \renewcommand{\arraystretch}{0}
    \newcommand{\imgsizer}{.1\textwidth}
    \newcommand{\shift}{7.7ex}
    \newcommand{\prompt}{\cellcolor{gray!40}}

    \newcommand{\roww}[1]{
        \includegraphics[width=\imgsizer]{figures/qual_comparison/input_#1_.png}
        & \includegraphics[width=\imgsizer]{figures/qual_comparison/jtr_#1_semantic.png}
        & \includegraphics[width=\imgsizer]{figures/qual_comparison/diffusionmtl_#1_semantic.png}
        & \includegraphics[width=\imgsizer]{figures/qual_comparison/stablemtl_#1_semantic.png}

        & \includegraphics[width=\imgsizer]{figures/qual_comparison/jtr_#1_normal.png}
        & \includegraphics[width=\imgsizer]{figures/qual_comparison/diffusionmtl_#1_normal.png}
        & \includegraphics[width=\imgsizer]{figures/qual_comparison/stablemtl_#1_normal.png}

        & \includegraphics[width=\imgsizer]{figures/qual_comparison/jtr_#1_depth.png}
        & \includegraphics[width=\imgsizer]{figures/qual_comparison/diffusionmtl_#1_depth.png}
        & \includegraphics[width=\imgsizer]{figures/qual_comparison/stablemtl_#1_depth.png}
    }

    \resizebox{.99\linewidth}{!}{
    \begin{tabular}{cc S{.2ex} cS{.1ex}cS{.1ex}c S{.2ex} cS{.1ex}cS{.1ex}c S{.2ex} cS{.1ex}cS{.1ex}c}
        && \multicolumn{3}{c}{\Semantic} & \multicolumn{3}{c}{\Normal} & \multicolumn{3}{c}{\Depth} \\
        \cmidrule(lr){3-5} \cmidrule(lr){6-8} \cmidrule(lr){9-11}
        &\scalebox{0.8}{Input} & \scalebox{0.8}{JTR} & \scalebox{0.8}{DiffusionMTL} & \scalebox{0.8}{\textbf{\OURS{}}} & \scalebox{0.8}{JTR} & \scalebox{0.8}{DiffusionMTL}  & \scalebox{0.8}{\textbf{\OURS{}}} & \scalebox{0.8}{JTR} & \scalebox{0.8}{DiffusionMTL} & \scalebox{0.8}{\textbf{\OURS{}}} \\

        \verti{4.2ex}{\scalebox{0.9}{\textbf{Cityscapes}}}&\roww{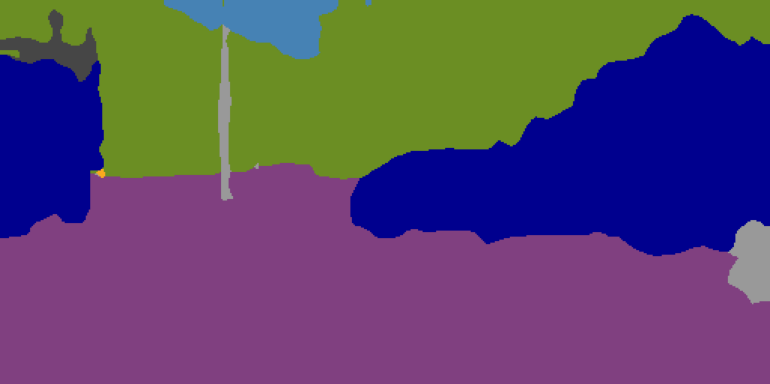} \\
        &\roww{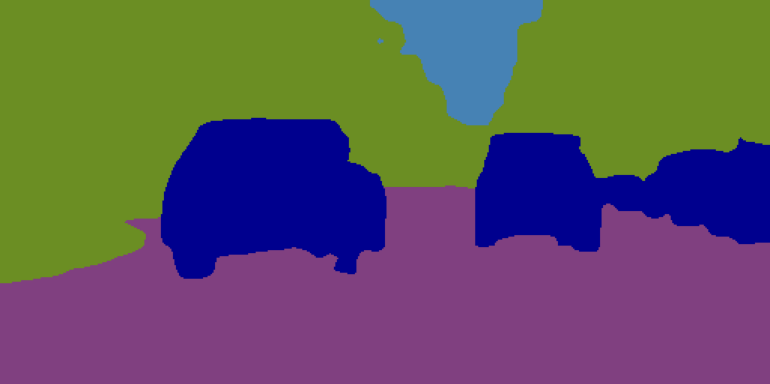} \\[.2ex]
        \verti{2.5ex}{\scalebox{0.9}{\textbf{KITTI}}}&\roww{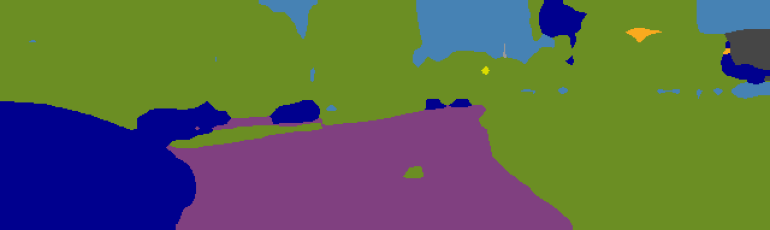} \\
        &\roww{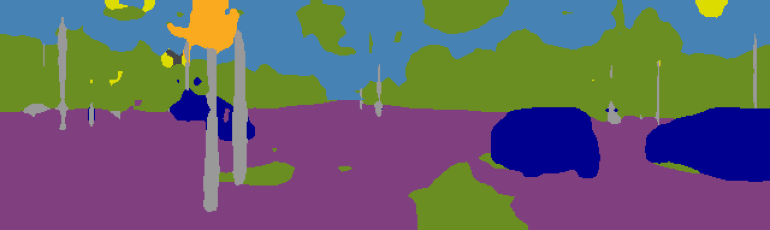} \\
    \end{tabular}}

    \caption{\textbf{Qualitative comparison.} We compare against original baseline versions as these are better performing (\cf~\cref{tab:main-results}) than the adapted full setting variants on the three tasks displayed. \OURS{} demonstrates superior qualitative results.}
    \label{fig:qualitative}
\end{figure*}

\condenseparagraph{Baselines.} We highlight that most MTL methods are designed to be trained on fully annotated datasets, which makes direct comparison impossible. Our main experiments therefore compare against DiffusionMTL~\cite{diffusionmtl} and JTR~\cite{nishi2024jtr}, state-of-the-art methods for MTL with partial annotations. The latter were originally designed for handling at most three tasks (among \semantic{}, \normal{}, and \depth{}) and require task-specific pixel-level losses. We report their original implementation performance on our ensemble of datasets using the baselines losses and loss weights. However, for a more direct comparison with our method, we made significant efforts to adapt the baselines to our full task set $\alltask$ experiment, including support for temporal tasks, which required modifications to the architectures. The adaptations, reported as $\text{DiffusionMTL*}$ and $\text{JTR*}$, use reweighted additional losses for scene flow, optical flow, albedo, and shading to {better balance loss magnitudes}.

In addition to \OURS{}, we report our single-stream architecture, \OURSSS{} (\cref{sec:singlestream}), as it is also capable of performing multitask. Our single-task baseline is Lotus-D~\cite{he2024lotus}, trained separately for each task. All baselines follow their original training protocols, {\ie, sampling all available annotations simultaneously}. DiffusionMTL~\cite{diffusionmtl} and JTR~\cite{nishi2024jtr} use task-specific image-level losses, whereas our single-task baseline and \OURS{} use the MSE loss. Task-gradient isolation is applied only for \OURS{}.

\condenseparagraph{Metrics.}
We measure the mean intersection over union (\mIoU) for \semantic{}; absolute relative error (\AbsRel) for \depth{}; mean angular error in degrees (\MAE) for \normal{}; average endpoint error for \opticalflow{} (\EPEtwoD) and \sceneflow{} (\EPEthreeD); and root mean squared error (\RMSE) for \shading{} and \albedo{}. For unbounded quantities, we perform a per-channel least squares fitting against the ground truth~\cite{marigold,careaga2023Intrinsic}. The delta metric ($\deltam$) is used to quantify multi-task performance~\cite{deltam}. It measures the relative performance over all tasks of a model compared to its single-task counterparts. For \depth{}, the $\deltam$ is averaged over two datasets.

\condenseparagraph{Implementation.}
\OURS{} uses {the Stable Diffusion (SD) v2 architecture} {but could adapted to any other SD variants (\eg, SD-XL, SD-3) due to our minimal architectural modifications.}
We provide details on {the different task encodings in~\cref{sec:appendix:task-encoding}.}
Our architecture follows Lotus-D~\cite{he2024lotus} (discriminative variant) without its detail preserver.
{To support two-frame inputs, we duplicate the input filters in the first convolution and divide these weights by three~\cite{marigold}}. We insert our task-attention layers at each of the {sixteen} transformer blocks of $\unet$.
We employ multi-head attention~\cite{vaswani2017attention} to foster learning diversity. Our optimal model uses four attention heads, discussed in~\cref{sec:discussion}.

\subsection{Main results}

\condenseparagraph{Domain generalization performance} \cref{tab:main-results} compares our method with four baselines on eight benchmarks across seven tasks. When comparing \OURS{} with the best baseline in each task, we observe a significant boost {on all tasks, such as} +9.87 \mIoU{}~(\semantic{}), -12.37 \MAE{}~(\normal{}), and -0.1189~\EPEthreeD{} (\sceneflow{}).

{While the single-stream version, \OURSSS{}, underperforms single-task baseline with --1.57 $\deltam{}$ in MTL performance. Our full model improves the overall performance by +4.78 $\deltam{}$, showing the benefits of a multi-stream architecture and cross-task attention}. We highlight that {since} DiffusionMTL and JTR use task-specific losses, {they have difficulty scaling as their training is unstable with more tasks}. Note how \normal{} and \depth{} degrade in the full setting (\eg, comparing JTR* and JTR).

We further assess qualitative results in~\cref{fig:qualitative}, demonstrating that we consistently outperforms the baselines. Additional results on real-world datasets~\cite{cityscapes,kittiflow,waymo,argoverse2,ddad,pandaset, youtubevos, perazzi2016davis} are shown in~\cref{fig:teaser} and in~\cref{sec:supp:qualitative}, proving it generalizes to diverse out-of-distribution domains. Further qualitative results are available in the supplementary video: \href{https://youtu.be/exsVFj34lrI}{https://youtu.be/exsVFj34lrI}.

\condenseparagraph{Single-dataset performance} {For closer comparison, we report} {in~\cref{tab:nyuv2-results} results in the \emph{single-dataset training with two label regimes} from MTPSL~\cite{MTPSL}: “One” uses a single task label per instance while “Random” uses a random number of task labels per instance. \OURS{} outperforms other SOTA baselines with} {significant margins on all metrics.}

\begin{table}[h]
    \centering
    \resizebox{\linewidth}{!}{
    \begin{tabular}{lcccccc}
        \toprule
        \multirow{3}{*}{Method} & \multicolumn{3}{c}{One} & \multicolumn{3}{c}{Random} \\
        \cmidrule(lr){2-4}\cmidrule(lr){5-7}
        & \Semantic & \Normal & \Depth & \Semantic & \Normal & \Depth \\
        & {\footnotesize \mIoU{} \%$\uparrow$}
        & {\footnotesize \MAE{} °$\downarrow$}
        & {\footnotesize \AbsRel{} \%$\downarrow$}
        & {\footnotesize \mIoU{} \%$\uparrow$}
        & {\footnotesize \MAE{} °$\downarrow$}
        & {\footnotesize \AbsRel{} \%$\downarrow$} \\
        \midrule
        MTPSL~\cite{MTPSL}       & 30.36 &  32.08 & 0.6088 & 34.91  & 31.20 & 0.5738 \\
        JTR~\cite{nishi2024jtr} & 31.96  & 30.80 & 0.5919 & 37.08  & 29.44 & 0.5541 \\
        DiffusionMTL~\cite{diffusionmtl} & 44.47 & 25.84 & 0.5059 & 46.82 & 24.75 & 0.4743 \\
        \OURS{}    & \textbf{47.81} & \textbf{22.45} & \textbf{0.4136} & \textbf{50.43} & \textbf{21.91} & \textbf{0.4034} \\
        \bottomrule
    \end{tabular}
    }
    \caption{\textbf{NYUv2 partial labels.} Train/val setting following~\cite{MTPSL}}
    \label{tab:nyuv2-results}
\end{table}

\subsection{Discussion}
\label{sec:discussion}

\begin{table*}
	\centering
	\setlength{\tabcolsep}{2pt}
	\resizebox{.85\textwidth}{!}{
		\begin{tabular}{c c c c c c | c c c c c c c c|c}
			\toprule
			&\SSemantic{} \& \SShading{} \& \AAlbedo{} & \multicolumn{2}{c}{\OOpticalflow{} \& \SSceneflow{}} &
			\multicolumn{2}{c|}{\NNormal{} \& \DDepth{}} &

			\textbf{\textcolor{semantic}{Semantic}} &
			\textbf{\textcolor{normal}{Normal}}  &
			\multicolumn{2}{c}{\textbf{\textcolor{depth}{Depth}}} &
			\textbf{\textcolor{opticalflow}{Opt. Flow}} &
			\textbf{\textcolor{sceneflow}{Sc. Flow}} &
			\textbf{\textcolor{shading}{Shading}} &
			\textbf{\textcolor{albedo}{Albedo}} &
			\textbf{MTL Perf.} \\

			\cmidrule(lr){2-2} \cmidrule(lr){3-4} \cmidrule(lr){5-6} \cmidrule(lr){7-7} \cmidrule(lr){8-8} \cmidrule(lr){9-10}
			\cmidrule(lr){11-11} \cmidrule(lr){12-12} \cmidrule(lr){13-13} \cmidrule(lr){14-14} \cmidrule(lr){15-15}

			&\multirow{2}{*}{\makecell{Hypersim \&\\Virtual KITTI 2}} & Virtual &  Flying  &  Virtual  & \multirow{2}{*}{Hypersim}
			& {\footnotesize \mIoU{} \%$\uparrow$}
			& {\footnotesize \MAE{} °$\downarrow$}
			& {\footnotesize \AbsRel{} \%$\downarrow$}
			& {\footnotesize \AbsRel{} \%$\downarrow$}
			& {\footnotesize \EPEtwoD{} px$\downarrow$}
			& {\footnotesize \EPEthreeD{} m$\downarrow$}
			& {\footnotesize \RMSE{}$\downarrow$}
			& {\footnotesize \RMSE{}$\downarrow$}
			& {\footnotesize $\deltam{} \%\uparrow$} \\

			& &  KITTI 2 & Things 3D & KITTI 2 & & \dataset{Cityscapes} & \dataset{DIODE} & \dataset{KITTI} & \dataset{DIODE} & \dataset{KITTI} & \dataset{KITTI}  & \dataset{MID} & \dataset{MID} & \dataset{Avg} \\

			\midrule
			\#1&\checkmark & \checkmark  & \checkmark  & \checkmark  & \checkmark  & \cellcolor{second}55.08 & \cellcolor{second}23.36 & \cellcolor{second}14.03 & \cellcolor{best}32.97 & \cellcolor{best}11.22 & \cellcolor{best}0.2297 & 0.2350 & \cellcolor{second}0.2061 & \cellcolor{best}+4.14  \\
			\#2&\checkmark & - & \checkmark   & \checkmark &  \checkmark  & 53.47 & 23.69 & \cellcolor{best}13.74 & 33.54 & 24.15 & 0.4126  & 0.2367 & 0.2103 & --24.30 \\
			\#3&\checkmark & \checkmark &  - & \checkmark & \checkmark & \cellcolor{best}55.28 & \cellcolor{best}22.29 & 14.82 & \cellcolor{second}33.21  & 13.60 & 0.2541 &  0.2383  & \cellcolor{best}0.2012  & \cellcolor{second}+0.01\\
			\#4&\checkmark & \checkmark & \checkmark & - & \checkmark & 44.55 & 23.79 & 17.98 & 34.13 & \cellcolor{second}11.99 & \cellcolor{second}0.2456 &  \cellcolor{best}0.2236 & 0.2110 & --3.10\\
			\#5&\checkmark & \checkmark & \checkmark & \checkmark & - & 50.12 & 52.65 & 18.26 & 45.27 & 12.40 & 0.2484  & \cellcolor{second}0.2279 & 0.2107 & --23.46 \\

			\bottomrule
	\end{tabular}}

	\caption{\textbf{Ablation study of training datasets.} Removing task supervision from any dataset degrades performance on the corresponding tasks~(\cf. text for details). We highlight \colorbox{best}{best} and \colorbox{second}{second-best}.}
	\label{tab:ablate-training-datasets}
\end{table*}

\begin{table*}
    \centering
    \begin{subfigure}[b]{0.8\textwidth}
        \centering
        \resizebox{\linewidth}{!}{
        \begin{tabular}{l H| c c c c c c c c|c}
            \toprule
            \multirow{3}{*}{\textbf{Method}} &  \multirow{3}{*}{\makecell{\textbf{Multi-frame}\\ \textbf{adaptation}}} &
            \textbf{\textcolor{semantic}{Semantic}} &
            \textbf{\textcolor{normal}{Normal}} &
            \multicolumn{2}{c}{\textbf{\textcolor{depth}{Depth}}} &
            \textbf{\textcolor{opticalflow}{Opt. Flow}} &
            \textbf{\textcolor{sceneflow}{Scene Flow}} &
            \textbf{\textcolor{shading}{Shading}} &
            \textbf{\textcolor{albedo}{Albedo}} &
            \textbf{MTL Perf.} \\

            \cmidrule(lr){3-3} \cmidrule(lr){4-4} \cmidrule(lr){5-6}
            \cmidrule(lr){7-7} \cmidrule(lr){8-8} \cmidrule(lr){9-9}
            \cmidrule(lr){10-10} \cmidrule(lr){11-11}

            & & {\footnotesize \mIoU{} \%$\uparrow$}
            & {\footnotesize \MAE{} °$\downarrow$}
            & {\footnotesize \AbsRel{} \%$\downarrow$}
            & {\footnotesize \AbsRel{} \%$\downarrow$}
            & {\footnotesize \EPEtwoD{} px$\downarrow$}
            & {\footnotesize \EPEthreeD{} m$\downarrow$}
            & {\footnotesize \RMSE{}$\downarrow$}
            & {\footnotesize \RMSE{}$\downarrow$}
            & {\footnotesize $\deltam{} \%\uparrow$} \\

            & & \dataset{Cityscapes} & \dataset{DIODE} & \dataset{KITTI} & \dataset{DIODE} & \dataset{KITTI} & \dataset{KITTI}  & \dataset{MID} & \dataset{MID} & \dataset{Avg} \\

            \midrule
            \textbf{\OURS{}($\rho{=}0$)} & \multirow{4}{*}{Duplicate}
              & \cellcolor{best}54.90
              & \cellcolor{best}22.88
              & \cellcolor{best}14.90
              & \cellcolor{best}32.57
              & \cellcolor{second}11.45
              & \cellcolor{second}0.2400
              & 0.2351
              & \cellcolor{best}0.2023
              & \cellcolor{best}+3.41 \\

            w/o separate $(q_t,k_t,v_t)$ &
              & 48.38
              & 24.15
              & 16.17
              & 33.24
              & \cellcolor{best}11.06
              & \cellcolor{best}0.2327
              & \cellcolor{second}0.2295
              & 0.2066
              & \cellcolor{second}+0.85 \\

            w/o single-stream init. &
              & 44.69
              & 23.35
              & 15.60
              & \cellcolor{second}33.21
              & 12.17
              & 0.2647
              & 0.2325
              & 0.2110
              & --3.11 \\

            w/o multi-stream &
              & \cellcolor{second}52.57
              & 23.94
              & 15.64
              & 33.36
              & 12.76
              & 0.2618
              & 0.2310
              & 0.2077
              & --1.57 \\
            \bottomrule
        \end{tabular}
        }
    \end{subfigure}
    \caption{\textbf{Multi-stream ablation.} Each design choice contributes to the best performance.}
    \label{tab:multistream-ablation}
\end{table*}

\begin{figure}
	\centering
	\centering
    \begin{subfigure}[m]{0.62\linewidth}
        \small
        \centering
        \setlength{\tabcolsep}{1pt}
        \begin{tabular}{cc}
            \resizebox{0.33\linewidth}{!}{\Semantic{} (mIoU $\uparrow$)} & \resizebox{0.33\linewidth}{!}{\Normal{}~(MAE $\downarrow$)} \\
            \includegraphics[width=0.45\linewidth]{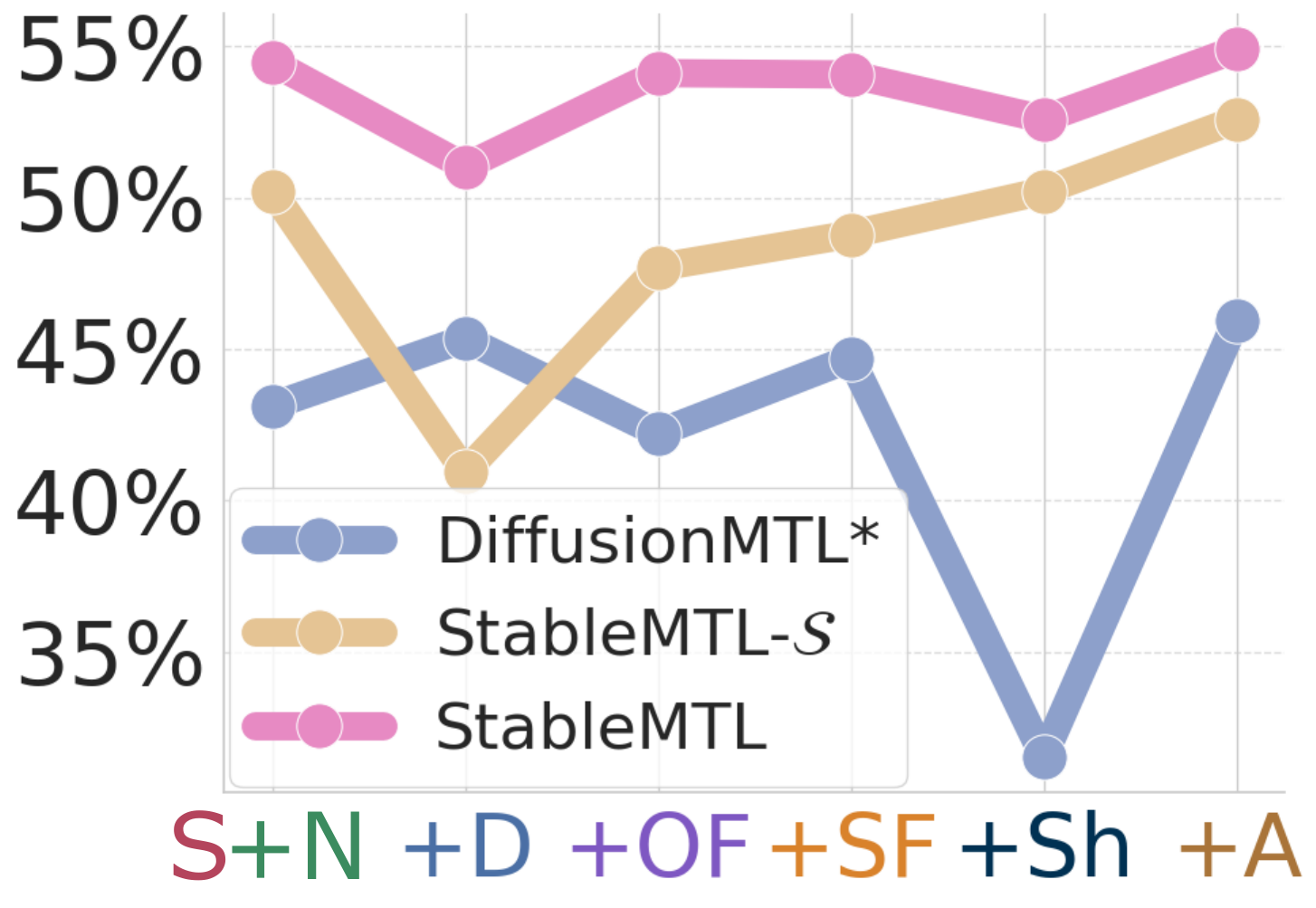} & \includegraphics[width=0.45\linewidth]{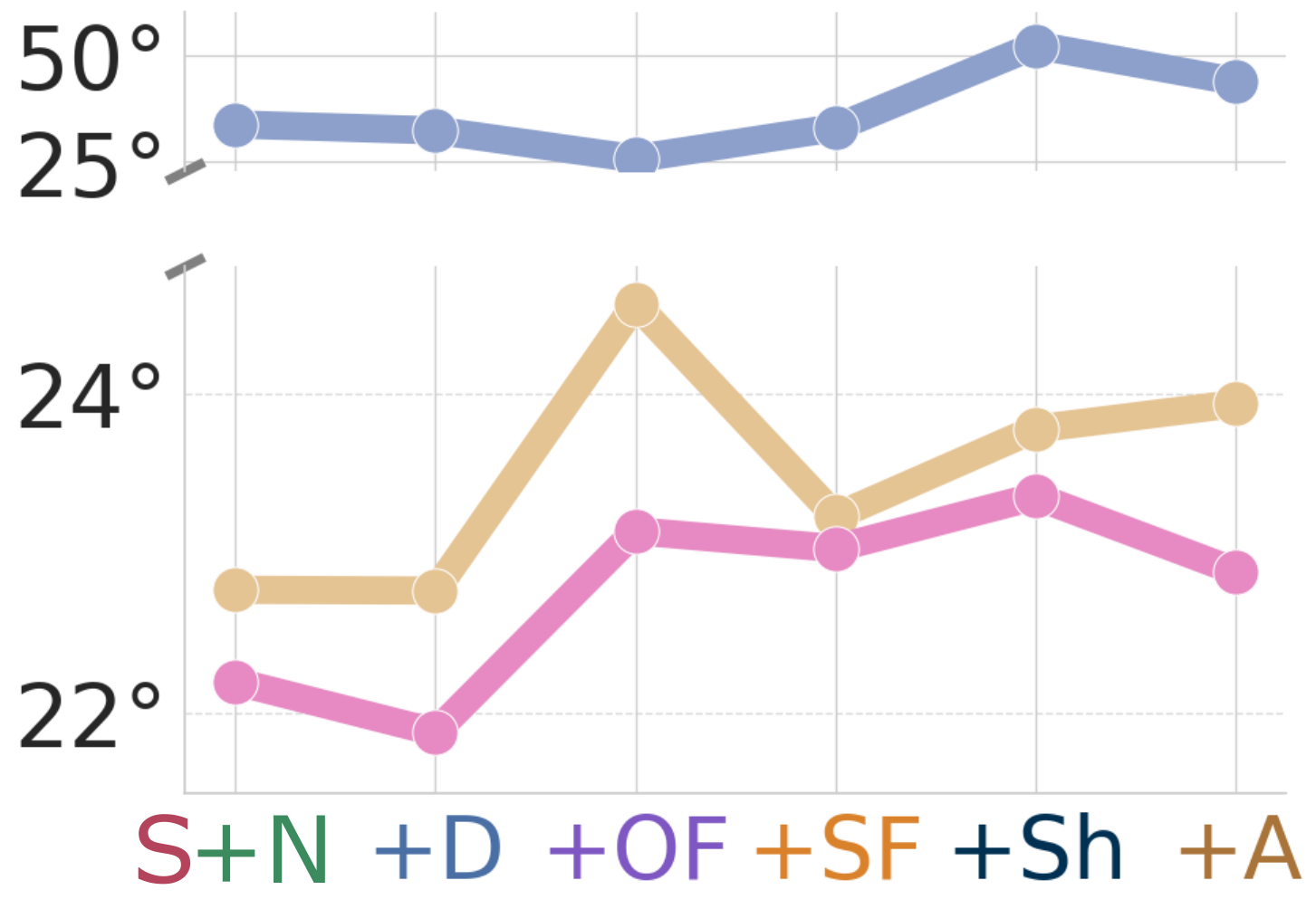}\\[-0.5em]
        \end{tabular}
        \caption{Tasks ablation.}
        \label{fig:ablate-tasks-combination}
    \end{subfigure}\hfill
    \begin{subfigure}[m]{0.37\linewidth}
        \resizebox{1.0\textwidth}{!}{
        \begin{tabular}{ccc |c}
            \toprule
            $\mathcal{L}_{\rm low}$ & $\mathcal{L}_{\rm high}$ &
            $\mathcal{L}_{\rm task}$ & $\deltam{} \%\uparrow$ \\
            \midrule
            \checkmark &&& +4.78 \\
            && \checkmark & +2.77 \\
            \checkmark & \checkmark & & +5.05 \\
            \checkmark & \checkmark & \checkmark & \cellcolor{second}+6.88 \\
            \checkmark & &\checkmark & \cellcolor{best}+6.94 \\

            \bottomrule
        \end{tabular}}
        \caption{Loss ablation.}
        \label{tab:loss-ablation}
    \end{subfigure}

    \caption{\textbf{Tasks and loss ablations.}
        {In \cref{fig:ablate-tasks-combination}, we separately train with an increasing number of tasks and show the \semantic{} and \normal{} metrics. In \cref{tab:loss-ablation}, we compare different loss objectives; our latent loss yields strong results, further enhanced by task-specific losses at the cost of higher memory usage.}\vspace{-0.5em}
    }
\end{figure}

\condenseparagraph{Impact of training datasets.}\cref{tab:ablate-training-datasets} shows that removing individual datasets degrades performance, depending on domain similarity and annotation quality.
For example, removing \vkitti{} for flow tasks (row \#2) leads to bigger {larger} degradation on \scflow{} and \optflow{}, compared to removing \flying{} (row \#3). {This shows} \vkitti{} better aligns with our evaluation domain.
Removing \hypersim{} for geometrical tasks (row \#5) drastically impacts \depth{} and \normal{}, especially on the \diode{} dataset, which may result from its higher annotation quality and domain match.
Interestingly,~\kitti{}~\depth{} exhibits similar drops regardless of the dataset removed, suggesting the domain proximity of ~\vkitti{} {outweighs its lower annotation quality, while~\hypersim{} alone} cannot overcome the larger domain gap.

\condenseparagraph{Multi-stream ablation.} In~\cref{tab:multistream-ablation}, {we ablate various design choices of the method}.
Removing separate projection layer per task \ie, "w/o separate $(q_t,k_t,v_t)$" improves performance on certain tasks like \opticalflow{}, \sceneflow{}, and \shading{}, but degrades others and leads to an overall drop of 2.56 in $\deltam$. {We attribute this to the highly similar attention patterns across tasks which limits task interactions.}
Initializing $\unetsimple$ with Stable Diffusion weights {instead of $\theta$} \ie, "w/o single-stream init.", significantly degrades performance by 6.52 in $\deltam$.
We also report using only single-stream \ie, "w/o multi-stream" also degrading performance on all tasks (drop of 4.98 in $\deltam$), since the single streams doe not benefit from task exchanges. Additionally, we ablate our multi-frame adaptation on the single-stream only for single-input tasks, with ours being $z_{i,j} {=} \texttt{concat}(z_{i}, z_{j})$, compared to $z_{i,j} {=} \texttt{avg}(z_{i}, z_{j})$, and $z_{i,j} {=} \texttt{concat}(z_{i}, \emptyset)$; they {yield} $-1.57$, $-4.11$, and $-2.72$ in $\deltam$, respectively.

\condenseparagraph{Benefit of task-gradient isolation.} \Cref{tab:ablate-grad-accum-scheme} highlights the efficacy of our isolation scheme, {improving performance} by 2.54 $\deltam$. In particular, it drastically boosts \semantic{} and \albedo{}, which have lower gradient norms as seen in~\cref{fig:gradnorm_ablate_training}. The latter further reveals an inverse relationship between gradient norms and the performance drop when removing our isolation scheme. Besides, evaluation at different training steps in~\cref{fig:deltamtl_vs_iteration_gradient_isolation} proves our scheme is sample efficient.

\condenseparagraph{Impact of training tasks.} \Cref{fig:ablate-tasks-combination} shows performance on \semantic{} and \normal{} when training on different tasks sets. Adding new tasks to the pipeline affects single-task performance, which is a clear issue for \OURSSS{} but is mitigated by the multi-stream architecture of \OURS{} thanks to its task-attention mechanism.

\begin{figure*}
	\centering
	\setlength{\tabcolsep}{1pt}
	\begin{subfigure}[c]{0.59\textwidth}
		\resizebox{1.0\linewidth}{!}{
			\begin{tabular}{l| c c c c c c c c|c}
				\toprule
				\multirow{3}{*}{\textbf{Method}} &
				\textbf{\textcolor{semantic}{Semantic}} &
				\textbf{\textcolor{normal}{Normal}}  &
				\multicolumn{2}{c}{\textbf{\textcolor{depth}{Depth}}} &
				\textbf{\textcolor{opticalflow}{Opt. Flow}} &
				\textbf{\textcolor{sceneflow}{Scene Flow}} &
				\textbf{\textcolor{shading}{Shading}} &
				\textbf{\textcolor{albedo}{Albedo}} &
				\textbf{MTL Perf.} \\

				\cmidrule(lr){2-2} \cmidrule(lr){3-3} \cmidrule(lr){4-5}
				\cmidrule(lr){6-6} \cmidrule(lr){7-7} \cmidrule(lr){8-8} \cmidrule(lr){9-9} \cmidrule(lr){10-10}

				& {\footnotesize \mIoU{} \%$\uparrow$}
				& {\footnotesize \MAE{} °$\downarrow$}
				& {\footnotesize \AbsRel{} \%$\downarrow$}
				& {\footnotesize \AbsRel{} \%$\downarrow$}
				& {\footnotesize \EPEtwoD{} px$\downarrow$}
				& {\footnotesize \EPEthreeD{} m$\downarrow$}
				& {\footnotesize \RMSE{}$\downarrow$}
				& {\footnotesize \RMSE{}$\downarrow$}
				& {\footnotesize $\deltam{} \%\uparrow$} \\

				& \dataset{Cityscapes} & \dataset{DIODE} & \dataset{KITTI} & \dataset{DIODE} & \dataset{KITTI} & \dataset{KITTI}  & \dataset{MID} & \dataset{MID} & \dataset{Avg} \\

				\midrule
				\OURSSS{} & 52.57 & 23.94 & 15.64 &  33.36 & 12.76 & 0.2618 &  0.2310 & 0.2077 & --1.57 \\
				w/o task-grad. isolation & 48.68 & 23.93 & 15.52 &  34.82 &  12.96 & 0.2583 & 0.2379  & 0.2178 & --4.11 \\
				\bottomrule
		\end{tabular}}
		\caption{Ablation of task-gradient isolation training scheme.}
		\label{tab:ablate-grad-accum-scheme}
	\end{subfigure}\hfill
	\begin{subfigure}[c]{0.19\textwidth}
		\centering
		\includegraphics[width=\linewidth]{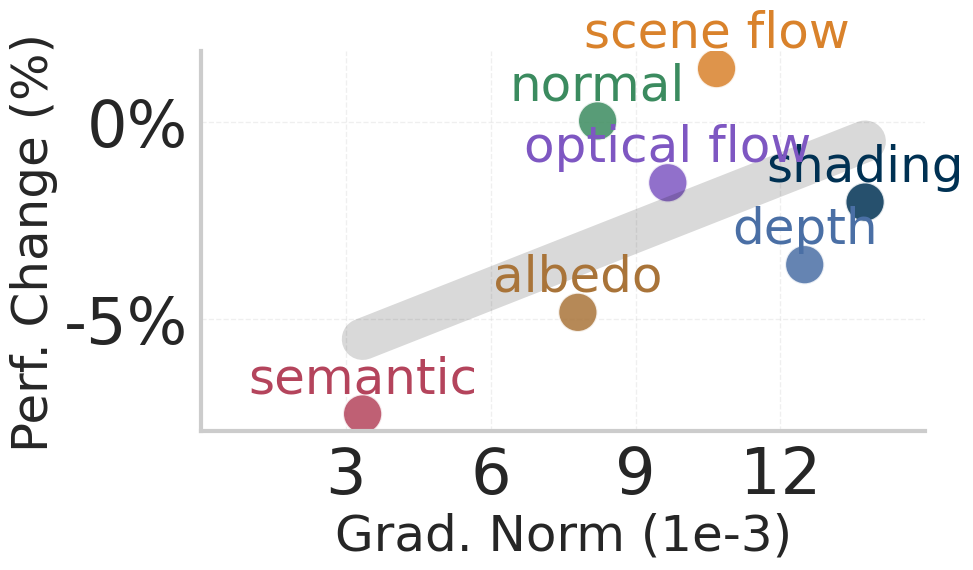}
		\caption{Observed correlation}
		\label{fig:gradnorm_ablate_training}
	\end{subfigure}\hfill
	\begin{subfigure}[c]{0.20\textwidth}
		\centering
		\includegraphics[width=\linewidth]{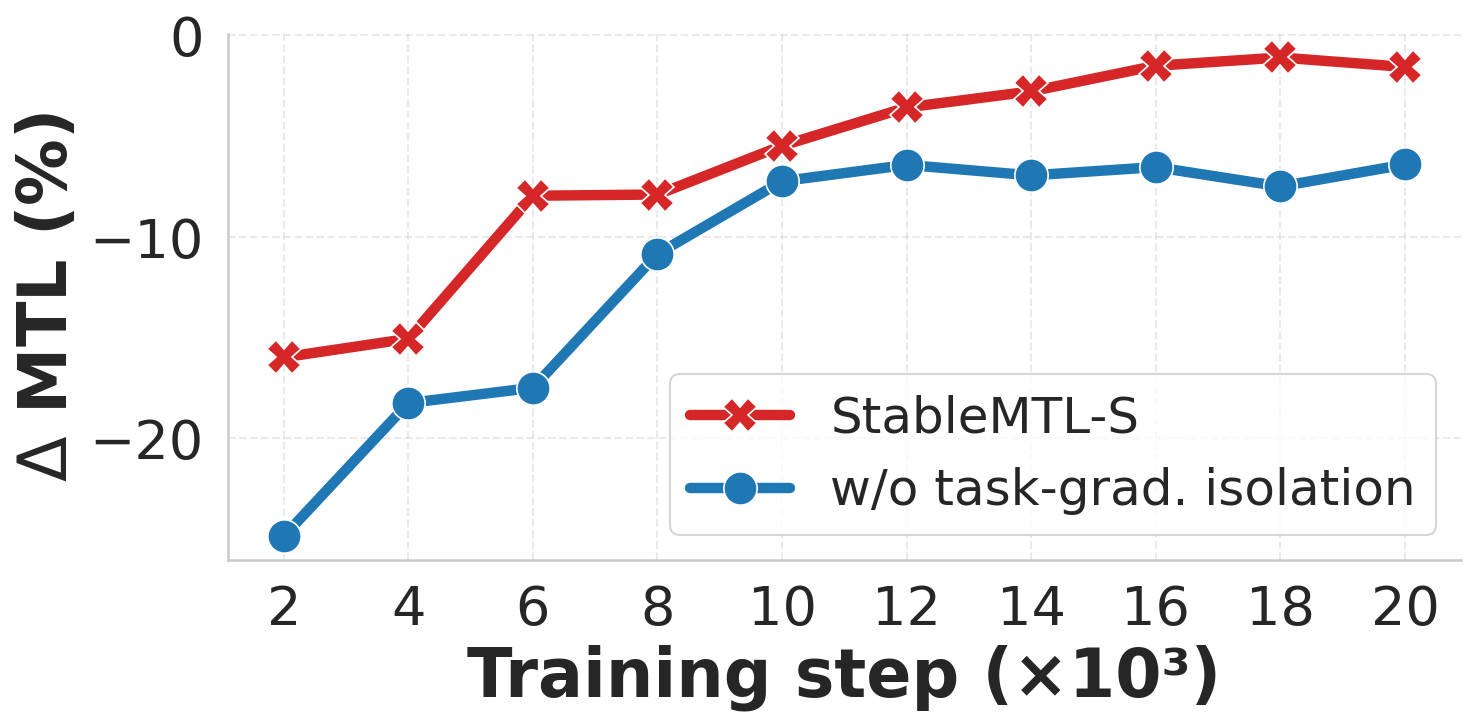}
		\caption{$\deltam$ vs training steps}
		\label{fig:deltamtl_vs_iteration_gradient_isolation}
	\end{subfigure}
	\vspace{-0.8em}
	\caption{\textbf{Task-gradient isolation strategy.} In
		\subref{tab:ablate-grad-accum-scheme}, we report the performance w/ and w/o our isolation strategy, showing that it drastically benefits some tasks (\eg,~\semantic{}) and improves the overall $\deltam{}$ metric.
		\subref{fig:gradnorm_ablate_training} shows that when removing gradient isolation, tasks with smaller gradient magnitudes are overwhelmed by those with larger ones, leading to a significant performance drop.
        Further, by evaluating performance throughout the training, \subref{fig:deltamtl_vs_iteration_gradient_isolation} demonstrates that our scheme is also continuously beneficial and therefore sample efficient.
	}
	\vspace{-0.7em}
\end{figure*}

\condenseparagraph{Loss ablation.} {In \cref{tab:loss-ablation} we evaluate three} {losses}{: $\mathcal{L}_{\rm low}$~(our MSE latent loss), $\mathcal{L}_{\rm high}$ (MSE loss at the VAE output), and $\mathcal{L}_{\rm task}$ (task-specific losses with balanced magnitude). Both $\mathcal{L}_{\rm high}$ and $\mathcal{L}_{\rm task}$ roughly double memory usage due to VAE decoder backpropagation. Using $\mathcal{L}_{\rm task}$ alone performs worse than $\mathcal{L}_{\rm low}$ since the former lacks explicit performance-based weighting (impractical for seven tasks).
Combining $\mathcal{L}_{\rm high}$ with $\mathcal{L}_{\rm low}$ yields a minor gain (+0.27~$\deltam{}$) and even a slight drop with $\mathcal{L}_{\rm low} + \mathcal{L}_{\rm task}$ (-0.06~$\deltam{}$). The best configuration is $\mathcal{L}_{\rm low} + \mathcal{L}_{\rm task}$, highlighting the value of adding task-specific supervision to our unified latent loss.}

{Importantly, performance of task-specific losses vary drastically with the balancing weights used. Therefore, in~\cref{fig:weight-ablation} we compare $\mathcal{L}_{\rm low}$ to $\mathcal{L}_{\rm task}$ in the manageable three tasks setup (\Semantic, \Normal, and \Depth), when varying the task weights used. \OURS{} (using only $\mathcal{L}_{\rm low}$) obtains competitive performance (+3.92\% $\deltam{}$) compared to the best weighted-setup (+4.07\% $\deltam{}$) which, in contradiction to our method, demanded an exhaustive grid search -- unbearable with more tasks.
Even with only three tasks only, certain weight combinations collapse (\eg, -68.94\% $\deltam{}$), highlighting the instability of such task balancing schemes.}

\begin{figure}[h]
	\centering
    \small
    \includegraphics[width=\linewidth]{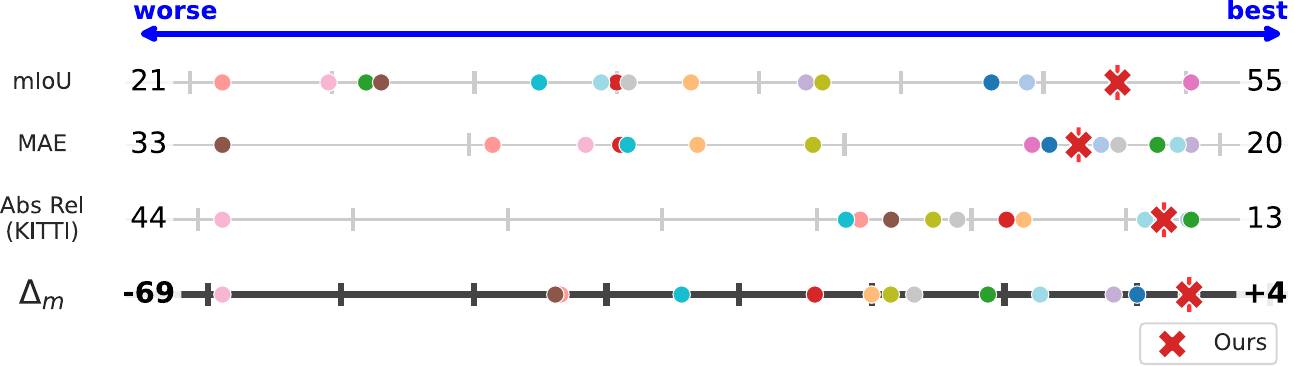}\vspace{-0.3em}
    \caption{{\textbf{Loss weight ablation.} We compare performance when using our unified latent loss ($\textcolor{red}{\pmb{\times}}$) or task-specific loss ($\bullet$) using different task weights combinations encoded as color.
    For each metric axis improvements points to the right. Our method achieves the best MTL metric (bottom axis) \textit{without any explicit balancing}.
	\vspace{-0.3em}}
    }
    \label{fig:weight-ablation}
\end{figure}

\condenseparagraph{Visualizing tasks interactions.}
\Cref{fig:attention} shows the task attention scores from main stream tasks (columns) to auxiliary stream tasks (represented by colored bars) at selected layers: here, layer {\texttt{12} and \texttt{15}.}
The observed patterns reveal strong {mutual} interactions like \normal-\depth, \opticalflow-\sceneflow, and \shading-\albedo, but also that some tasks are one-way beneficiaries{: \opticalflow{} relies on \semantic{} and \semantic{} on \shading{}.}
Further to assess causal cross-task transfer, in~\cref{fig:task_transfer} we mask one auxiliary stream at a time (y-axis) and report the performance change for other tasks (x-axis).
This reveals consistent \emph{mutual} and \textit{directional} interactions (\cf figure's caption). The above interactions support the findings of prior {works~\cite{zamir2018taskonomy,standley2020taskslearnedmultitasklearning}}.

\condenseparagraph{Effects of task attention heads.}Varying task attention to use 1/2/4/8 heads leads to an overall performance of $+2.45$/$+4.14$/$\mathbf{+4.78}$/$+4.41$ in $\deltam$, respectively.
{Although increased head counts promote greater flexibility}, a plateau is quickly reached, likely due to the reduced capacity of each head. {All variants} outperform the single-stream \OURSSS{}, which achieved $-1.57$ $\deltam$.

\DeclareRobustCommand{\vizmarker}[1]{
  \protect\tikz[baseline=-0.6ex]{\node[draw,circle,inner sep=0.7pt,minimum size=1.4ex] {\scriptsize #1};}
}
\begin{figure}[h]
	\centering
    \begin{subfigure}[t]{1.0\columnwidth}
        \setlength{\tabcolsep}{0pt}
        \renewcommand{\arraystretch}{0}
        \newcommand{\shift}{7.7ex}
        \newcommand{\prompt}{\cellcolor{gray!40}}
        \newcommand{\row}[1]{
        	\includegraphics[width=\imgsize]{figures/attention/semantic_#1_0_mean_barplot.png}
        	& \includegraphics[width=\imgsize]{figures/attention/normal_#1_0_mean_barplot.png}
        	& \includegraphics[width=\imgsize]{figures/attention/depth_#1_0_mean_barplot.png}
        	& \includegraphics[width=\imgsize]{figures/attention/optical_flow_#1_0_mean_barplot.png}
        	& \includegraphics[width=\imgsize]{figures/attention/scene_flow_#1_0_mean_barplot.png}
        	& \includegraphics[width=\imgsize]{figures/attention/shading_#1_0_mean_barplot.png}
        	& \includegraphics[width=\imgsize]{figures/attention/albedo_#1_0_mean_barplot.png}
        }

        \centering
        \resizebox{1.0\linewidth}{!}{
            \scriptsize
            \begin{tabular}{H cS{.2ex}cS{.2ex}cS{.2ex}cS{.2ex}cS{.2ex}cS{.2ex}c}
                & \Semantic & \Normal & \Depth & \textbf{\textcolor{opticalflow}{Opt.Flow}} & \textbf{\textcolor{sceneflow}{Sc.Flow}} & \Shading & \Albedo \\
                & \row{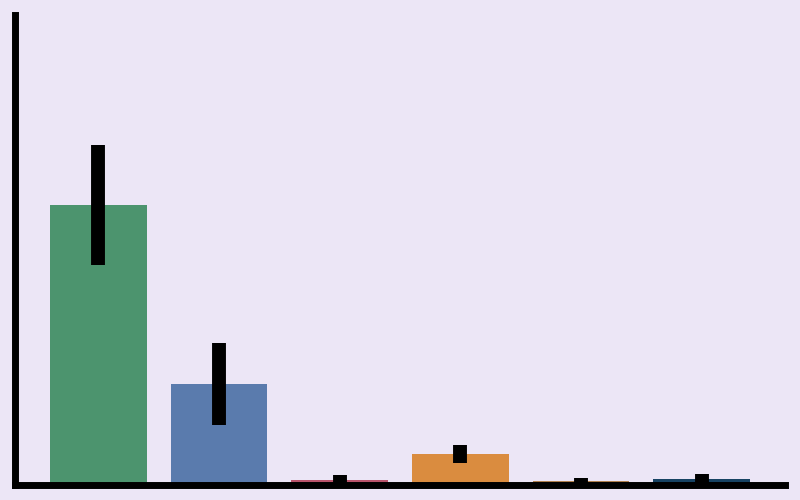} \\[.5ex]
                & \row{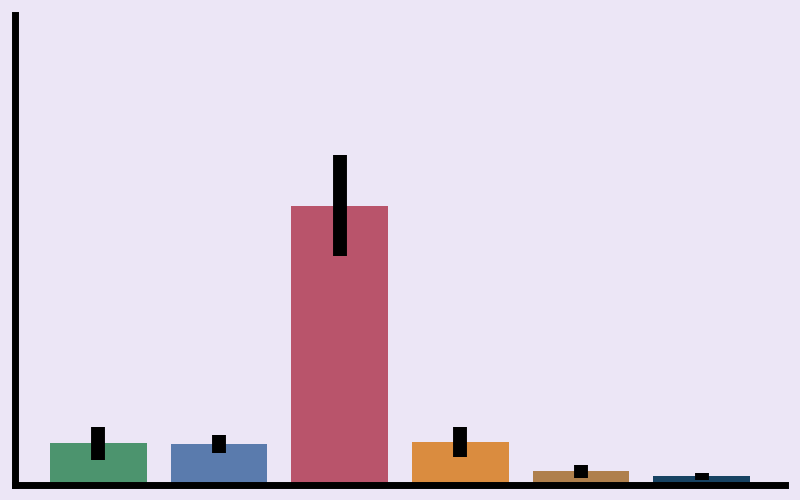} \\[.5ex]
            \end{tabular}
        }
        \caption{Task attention scores}
        \label{fig:attention}
    \end{subfigure}\\[0.2em]
    \begin{subfigure}[t]{0.43\columnwidth}
        \includegraphics[width=\columnwidth]{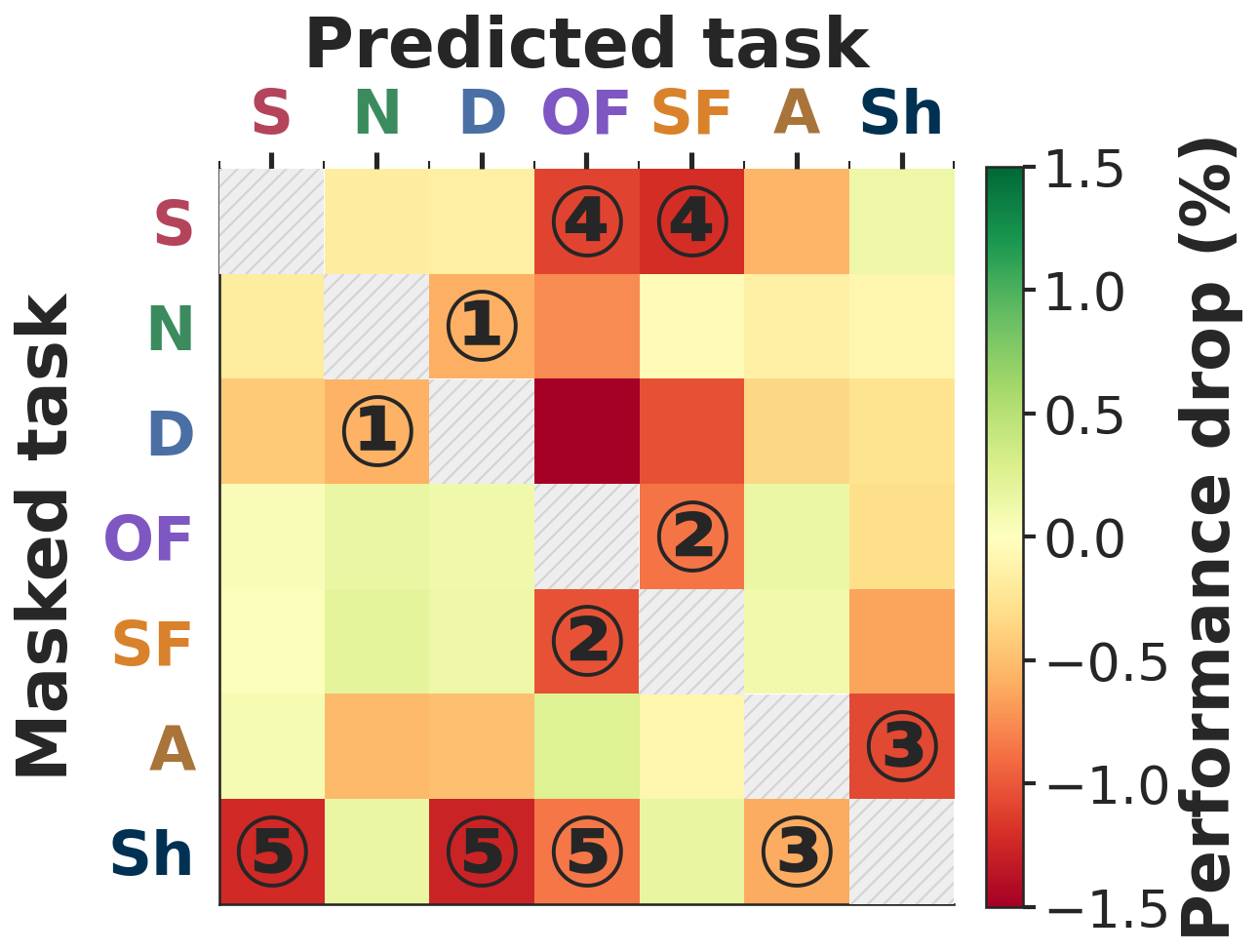}
        \caption{Cross-task transfer}
        \label{fig:task_transfer}
    \end{subfigure}\\[-0.5em]

    \caption{\textbf{Tasks interactions.} \subref{fig:attention} displays attention between tasks, showing strong interactions are taking place only between tasks as observed in the attention scores. In \subref{fig:task_transfer}: Masking out each task reveals two types of interaction: \textit{mutual} (\Normal--\Depth{}~\vizmarker{1}, \Opticalflow--\Sceneflow{}~\vizmarker{2}, and \Shading--\Albedo{}~\vizmarker{3}) and \textit{directional} (masking \Semantic{} degrades \Opticalflow{}/ \Sceneflow{}~\vizmarker{4}; masking \Shading{} impacts \Semantic{}/\Depth{}/\Opticalflow{}~\vizmarker{5}).
	\vspace{-1.0em}
    }
\end{figure}

\condenseparagraph{Limitations.}
{Despite} enhancing multi-task capabilities, \OURS{} performance on some individual tasks lags behind single-task models.
Furthermore, adding some new tasks (\eg, shading) affects the individual performances on other tasks as evidenced in~\cref{fig:ablate-tasks-combination}, showing that cross-task interaction is far from optimal yet.
In terms of inference,  multi-task prediction is currently sequential. We believe this is a constraint that future work could address with models enabling simultaneous multi-output prediction via improved multi-token conditioning. Additionally, the uniform task sampling strategy could be refined through adaptive approaches.
{Affine-invariant metrics effectively capture relative performance differences between methods~\cite{marigold,he2024lotus,midas}, yet may not fully reflect downstream performance.}

\section{Conclusion}
This work extends partially supervised MTL to a more practical setting where the model is trained on multiple synthetic datasets with partial labels. To bridge the synthetic-to-real gap, we adapt a single-step latent diffusion model with multi-task encoding and task-gradient isolation. While capable of multi-tasking, it suffers from degraded overall performance. To address this, we introduce a multi-stream architecture with task attention, allowing the model to leverage inter-task relationships and improve performance. Our approach scales to more tasks and achieves strong results on unseen data.

\condenseparagraph{Acknowledgments.} This work was funded by the French Agence Nationale de la Recherche (ANR) with project SIGHT (ANR-20-CE23-0016) and performed with HPC resources from GENCI-IDRIS (Grants AD011014102R2, AD011014102R1, AD011014389R1, and AD011012808R3). The authors are also grateful for the resources provided by the CLEPS infrastructure from Inria Paris.

\appendix

\section{Additional ablations}
\label{sec:supp:add-ablation}

\condenseparagraph{Effect of task attention masking.} We ablate the task attention masking in~\cref{tab:ablate-attention-masking}.
Masking a single task proportional to its attention weight ($\text{Sample}(\pi_{\token})$) improves performance as the masking probability $\rho$ increases to an optimum (\ie, $\rho{=}4$), as this fosters exploration of diverse task combinations.
However, higher $\rho$ values produces excessive exploration, lowering performance.
Consequently, masking a random number of $k$ tasks at once ($\text{Sample}_k(\pi_{\token})$) is less effective.
Other approaches are also suboptimal: dropping the highest-attention task (argmax) prevents exploitation of the strongest signal, while randomly dropping tasks ($\mathcal{U}(\alltask)$) offers a weaker exploration incentive.

\begin{table}[h]
    \centering
    \setlength{\tabcolsep}{2pt}
    \resizebox{\linewidth}{!}{
    \begin{tabular}{l c| c c c c c c c c|c}
        \toprule
        \multirow{3}{*}{\textbf{Strategy}} &
        \multirow{3}{*}{\textbf{Prob.~($\rho$)}} &
        \textbf{\textcolor{semantic}{Semantic}} &
        \textbf{\textcolor{normal}{Normal}}  &
        \multicolumn{2}{c}{\textbf{\textcolor{depth}{Depth}}} &
        \textbf{\textcolor{opticalflow}{Opt. Flow}} &
        \textbf{\textcolor{sceneflow}{Scene Flow}} &
        \textbf{\textcolor{shading}{Shading}} &
        \textbf{\textcolor{albedo}{Albedo}} &
        \textbf{MTL Perf.} \\

        \cmidrule(lr){3-3} \cmidrule(lr){4-4} \cmidrule(lr){5-6}
        \cmidrule(lr){7-7} \cmidrule(lr){8-8} \cmidrule(lr){9-9} \cmidrule(lr){10-10} \cmidrule(lr){11-11}

        &
        & {\footnotesize \mIoU{} \%$\uparrow$}
        & {\footnotesize \MAE{} °$\downarrow$}
        & {\footnotesize \AbsRel{} \%$\downarrow$}
        & {\footnotesize \AbsRel{} \%$\downarrow$}
        & {\footnotesize \EPEtwoD{} px$\downarrow$}
        & {\footnotesize \EPEthreeD{} m$\downarrow$}
        & {\footnotesize \RMSE{}$\downarrow$}
        & {\footnotesize \RMSE{}$\downarrow$}
        & {\footnotesize $\deltam{} \%\uparrow$} \\

        & & \dataset{Cityscapes} & \dataset{DIODE} & \dataset{KITTI} & \dataset{DIODE} & \dataset{KITTI} & \dataset{KITTI}  & \dataset{MID} & \dataset{MID} & \dataset{Avg} \\

        \midrule
        \multirow{5}{*}{$\text{Sample}(\pi_{\token})$}
        & 0.0
        & \cellcolor{second}54.90
        & \cellcolor{second}22.88
        & 14.90
        & \cellcolor{best}32.57
        & 11.45
        & 0.2400
        & 0.2351
        & \cellcolor{best}0.2023
        & +3.41 \\
        & 0.2 & 54.86 & 23.49 & \cellcolor{second}14.71 & \cellcolor{second}32.61 &  11.00 & 0.2378  & 0.2349 & \cellcolor{second}0.2047 & +3.70 \\
        & \textbf{0.4} & \cellcolor{best}55.08 & 23.36  & \cellcolor{best}14.03 & 32.97 & 11.22 & \cellcolor{second}0.2297 & 0.2350 & 0.2061 & \cellcolor{best}+4.14 \\

        & 0.6 & 53.97 & 23.26 & 15.22  & 32.94 & \cellcolor{best}10.90 & \cellcolor{best}0.2284  & 0.2348 & 0.2029  & \cellcolor{second}+4.00 \\

        & 0.8 & 52.95 & 23.02 & 14.73  & 32.89 & 11.21 & 0.2386  & 0.2349 & 0.2093 & +2.72 \\
        \addlinespace

        $\text{Sample}_k(\pi_{\token})$
        & 0.4 & 54.65 & 23.82 & 15.02 & 33.15  & \cellcolor{second}10.96 & 0.2311  & \cellcolor{second}0.2338  & 0.2064 & +3.50 \\

        $\text{argmax}$
        & {\footnotesize\textit{idem}} & 54.45 & 23.09 & 14.98 & 32.87 & 11.22 & 0.2299  & 0.2339 & 0.2071 & +3.65 \\

        $\mathcal{U}(\alltask)$
        & {\footnotesize\textit{idem}} & 54.77 & \cellcolor{best}22.69 & 15.02 & 32.70 & 11.32 & 0.2398  & \cellcolor{best}0.2313 & 0.2057 & +3.60 \\
        \bottomrule

    \end{tabular}}

    \caption{\textbf{Ablation of attention masking.} Masking heads depending on their attention score using our distribution $\text{Sample}(\pi_{\token})$ boosts performance by encouraging exploration. On the contrary, other methods, detailed in the text, perform worse.
    \colorbox{best}{Best} and \colorbox{second}{Second-best} are highlighted.
    }
    \label{tab:ablate-attention-masking}
\end{table}

\section{Additional metrics}
\label{sec:supp:additional-metrics}

We present IoU scores for each mapped class on the Cityscapes dataset~\cite{cityscapes} in~\cref{tab:supp:semantic}, with class mapping details provided in~\cref{tab:supp:vktocs_mapping_with_colors}. Overall, \OURS{} ranks either first or second on all classes.

\begin{table}[h]
    \centering
    \resizebox{1.0\linewidth}{!}{
    \begin{tabular}{l|c c c c c c c c | c}
        \toprule
        Method
        & \rotatebox{45}{\textcolor{mappedRoad}{$\blacksquare$} road}
        & \rotatebox{45}{\textcolor{mappedBuilding}{$\blacksquare$} building}
        & \rotatebox{45}{\textcolor{mappedPole}{$\blacksquare$} pole}
        & \rotatebox{45}{\textcolor{mappedLight}{$\blacksquare$} traffic light}
        & \rotatebox{45}{\textcolor{mappedSign}{$\blacksquare$} traffic sign}
        & \rotatebox{45}{\textcolor{mappedVegetation}{$\blacksquare$} vegetation}
        & \rotatebox{45}{\textcolor{mappedSky}{$\blacksquare$} sky}
        & \rotatebox{45}{\textcolor{mappedVehicle}{$\blacksquare$} vehicle}
        & \rotatebox{45}{mIoU}\\
        \midrule
        Single-task baseline & \stl{89.36} & \stl{61.82} & \stl{17.13} & \stl{0.42} & \stl{3.21} & \stl{72.75} & \stl{67.86} & \stl{72.82} & \stl{48.17} \\
        \addlinespace

        JTR~\cite{nishi2024jtr} & \cellcolor{best}96.33 & 51.48 & 15.33 & 0.87 & 3.22 & 61.28 & 77.36 & 60.16 & 45.75 \\
        JTR*~\cite{nishi2024jtr} & 71.50 & 0.00 & 0.00 & 0.00 & 0.00 & 38.69 & 53.47 & 0.00 & 20.46 \\

        DiffusionMTL~\cite{diffusionmtl} & 85.17 & 29.60 & 6.64 & 1.14 & 5.19 & 41.05 & 27.34 & 44.57 & 30.09 \\
        DiffusionMTL*~\cite{diffusionmtl} & 95.36 & \cellcolor{second}68.26 & 11.01 & \cellcolor{best}4.27 & \cellcolor{second}6.40 & 65.61 & 48.35 & 68.15 & 45.92 \\

        \textbf{\OURSSS{}} & 92.06 & 67.05 & \cellcolor{second}16.76 & 2.03 & 6.31 & \cellcolor{second}72.53 & \cellcolor{best}83.84 & \cellcolor{second}80.06 & \cellcolor{second}52.57 \\
        \textbf{\OURS} & \cellcolor{second}96.08 & \cellcolor{best}74.24 & \cellcolor{best}20.72 & \cellcolor{second}2.80 & \cellcolor{best}12.59 &\cellcolor{best}76.58 & \cellcolor{second}82.09 & \cellcolor{best}81.23 & \cellcolor{best}55.79 \\

        \bottomrule
    \end{tabular}}\\
    \caption{\textbf{Per-class \Semantic{} performance}. We report the IoU per mapped semantic class as well as the average IoU (mIoU), detailed in~\cref{tab:supp:vktocs_mapping_with_colors}, on Cityscapes.}
    \label{tab:supp:semantic}
\end{table}

For \depth{} evaluation on KITTI~\cite{kitti} and~\diode~\cite{diode}, additional depth metrics are reported in~\cref{tab:supp:depth}; these include the following commonly used metrics~\cite{eigen}: absolute relative error (Abs Rel), squared relative error (Sq Rel), root mean squared error (RMSE), mean $\log_{10}$ error (RMSE log), and threshold accuracies ($\delta$1, $\delta$2, $\delta$3).
On most of these metrics, \OURS{} surpasses all MTL baselines, only performing below the single-stream model \OURSSS{} on one metric in one dataset.

\begin{table}[h!]
	\scriptsize
	\centering
	\setlength{\tabcolsep}{0.005\linewidth}
	\resizebox{\linewidth}{!}{
		\begin{tabular}{l|ccccccc|ccccccc}
			\toprule
			\multirow{2}[2]{*}{\textbf{Method}} & \multicolumn{7}{c|}{\small\textbf{\kitti}} & \multicolumn{7}{c}{\small\textbf{\diode}} \\

			& Abs Rel$\downarrow$ & Sq Rel$\downarrow$ & RMSE$\downarrow$ & RMSE log$\downarrow$ & $\delta$1$\uparrow$ & $\delta$2$\uparrow$  & $\delta$3$\uparrow$  & Abs Rel$\downarrow$ & Sq Rel$\downarrow$ & RMSE$\downarrow$ & RMSE log$\downarrow$ & $\delta$1$\uparrow$  & $\delta$2$\uparrow$  & $\delta$3$\uparrow$  \\
			\midrule
			Single-task baseline & \stl{14.21} & \stl{0.7319} & \stl{4.1734} & \stl{0.2014} & \stl{81.19} & \stl{96.52} & \stl{99.16} & \stl{32.56} & \stl{3.9233} & \stl{3.9632} & \stl{0.3163} & \stl{73.72} & \stl{87.72} & \stl{93.11} \\
                \addlinespace

                JTR~\cite{nishi2024jtr} & 26.39 & 1.7532 & 6.1357 & 0.2852 & 64.60 & 88.30 & 95.86 & 66.39 & 8.1448 & 8.4119 & 0.5482 & 42.26 & 67.66 & 81.59 \\
                JTR*~\cite{nishi2024jtr} & 39.27 & 4.0532 & 9.1290 & 0.4301 & 42.99 & 73.19 & 88.00 & 73.14 & 9.2842 & 8.9962 & 0.5852 & 40.31 & 64.24 & 78.10 \\

                DiffusionMTL~\cite{diffusionmtl} & 21.10 & 1.4998 & 5.8491 & 0.2715 & 68.11 & 89.87 & 96.80 & 45.19 & 4.6659 & 4.9657 & 0.4106 & 56.93 & 78.91 & 88.56 \\
                DiffusionMTL*~\cite{diffusionmtl} & 24.83 & 1.8492 & 6.4705 & 0.3518 & 58.77 & 86.54 & 95.59 & 58.17 & 7.2017 & 7.6032 & 0.5166 & 49.57 & 73.62 & 84.55 \\
                \textbf{\OURSSS{}} & \cellcolor{second}15.64 & \cellcolor{second}0.8268 & \cellcolor{second}4.3713 & \cellcolor{second}0.2499 & \cellcolor{second}77.62 & \cellcolor{second}95.52 & \cellcolor{second}98.72 & \cellcolor{second}33.36 & \cellcolor{best}3.9504 & \cellcolor{second}4.0281 & \cellcolor{second}0.3227 & \cellcolor{second}73.42 & \cellcolor{second}87.17 & \cellcolor{second}92.62 \\
                \textbf{\OURS} & \cellcolor{best}14.98 & \cellcolor{best}0.8224 & \cellcolor{best}4.3707 & \cellcolor{best}0.2170 & \cellcolor{best}79.43 & \cellcolor{best}95.94 & \cellcolor{best}98.87 & \cellcolor{best}33.03 & \cellcolor{second}3.9516 & \cellcolor{best}4.0073 & \cellcolor{best}0.3192 & \cellcolor{best}73.72 & \cellcolor{best}87.45 & \cellcolor{best}92.89 \\
			\bottomrule
		\end{tabular}
	}
	\caption{\textbf{Additional \depth{} metrics.} In all but one metric, \OURS{} ranks first on both KITTI~\cite{kitti} and~\diode~\cite{diode}.}
	\label{tab:supp:depth}
\end{table}

We also report, in~\cref{tab:supp:shading-albedo}, metrics for \shading{} and \albedo{} on the MID dataset~\cite{murmann19mid} following~\cite{careaga2023Intrinsic}: structural similarity index (SSIM)~\cite{ssim}, local mean squared error (LMSE), and root mean squared error (RMSE). These results show our method outperforms all other MTL baselines.

\begin{table}[h]
	\centering
        \resizebox{.8\linewidth}{!}{
		\begin{tabular}{l|ccc|ccc}
			\toprule
			\multirow{2}[2]{*}{\textbf{Method}} & \multicolumn{3}{c|}{\Shading{}} & \multicolumn{3}{c}{\Albedo} \\

			& RMSE$\downarrow$ & SSIM$\uparrow$ & LMSE$\downarrow$ & RMSE$\downarrow$ & SSIM$\uparrow$ & LMSE$\downarrow$    \\
			\midrule
			Single-task baseline & \stl{0.2551} & \stl{0.4277} & \stl{0.0426} & \stl{0.2145} & \stl{0.6067} & \stl{0.0505} \\
                \addlinespace

                JTR*~\cite{nishi2024jtr} & 0.3030 & 0.1843 & 0.0530 & 0.3567 & 0.3102 & 0.0994 \\

                DiffusionMTL*~\cite{diffusionmtl} & 0.3064 & 0.3142 & 0.0487 & 0.3660 & 0.3606 & 0.1620 \\
                \textbf{\OURSSS{}} & \cellcolor{best}0.2311 & \cellcolor{best}0.4449 & \cellcolor{best}0.0363 & \cellcolor{second}0.2077 & \cellcolor{second}0.6151 & \cellcolor{second}0.0496 \\
                \textbf{\OURS} & \cellcolor{second}0.2346 & \cellcolor{second}0.4424 & \cellcolor{second}0.0366 & \cellcolor{best}0.2016 & \cellcolor{best}0.6199 & \cellcolor{best}0.0477 \\
			\bottomrule
		\end{tabular}
        }
	\caption{\textbf{Additional \shading{} and \albedo{} metrics.} On MIDIntrinsics~\cite{murmann19mid}, \OURS{} ranks either first or second; it is only surpassed by our single stream variant~\OURSSS{}.}
	\label{tab:supp:shading-albedo}
    \vspace{-.5em}
\end{table}

\section{In-domain evaluation}
\label{sec:supp:in-domain}

Finally, we report the performances on test splits of in-domain data in~\cref{tab:in-domain}. Our method beats all baselines by a large margin on all metrics.

\begin{table}[h]
    \centering
    \setlength{\tabcolsep}{2pt}
    \resizebox{\linewidth}{!}{
    \begin{tabular}{l| c c c c c c c c c c c c}
        \toprule
        \textbf{Method} &
        \textbf{\textcolor{semantic}{Semantic}} &
        \multicolumn{2}{c}{\textbf{\textcolor{normal}{Normal}}} &
        \multicolumn{2}{c}{\textbf{\textcolor{depth}{Depth}}} &
        \multicolumn{2}{c}{\textbf{\textcolor{opticalflow}{Opt. Flow}}} &
        \multicolumn{2}{c}{\textbf{\textcolor{sceneflow}{Scene Flow}}} &
        \textbf{\textcolor{shading}{Shading}} &
        \textbf{\textcolor{albedo}{Albedo}} &
        \textbf{MTL Perf.} \\

        \cmidrule(lr){2-2} \cmidrule(lr){3-4} \cmidrule(lr){5-6} \cmidrule(lr){7-8} \cmidrule(lr){9-10} \cmidrule(lr){11-11} \cmidrule(lr){12-12} \cmidrule(lr){13-13}

        & {\footnotesize \mIoU{} \%$\uparrow$}
        & {\footnotesize \MAE{} °$\downarrow$}
        & {\footnotesize \MAE{} °$\downarrow$}
        & {\footnotesize \AbsRel{} \%$\downarrow$}
        & {\footnotesize \AbsRel{} \%$\downarrow$}
        & {\footnotesize \EPEtwoD{} px$\downarrow$}
        & {\footnotesize \EPEtwoD{} px$\downarrow$}
        & {\footnotesize \EPEthreeD{} m$\downarrow$}
        & {\footnotesize \EPEthreeD{} m$\downarrow$}
        & {\footnotesize \RMSE{}$\downarrow$}
        & {\footnotesize \RMSE{}$\downarrow$}
        & {\footnotesize $\deltam{} \%\uparrow$} \\

        & \dataset{VKITT2} & \dataset{VKITT2} & \dataset{Hypersim} & \dataset{VKITT2} & \dataset{Hypersim} & \dataset{VKITT2} & \dataset{FlyingThings3D} & \dataset{VKITT2} & \dataset{Hypersim} & \dataset{Hypersim} & \dataset{Hypersim} & \dataset{Avg} \\

        \midrule
        Single-task baseline &
        \stl{64.43} & \stl{24.96} & \stl{17.97} & \stl{19.62} & \stl{17.76} & \stl{9.30} & \stl{12.87} & \stl{0.1508} & \stl{0.4201} & \stl{0.2205} & \stl{0.2478} & \stl{0.00} \\

        \addlinespace
        DiffusionMTL*~\cite{diffusionmtl} &
        38.83 & 56.27 & 68.98 & 34.94 & 34.20 & 32.45 & 32.52 & 0.2630 & 0.7094 & 0.2866 & 0.5724 & -109.02 \\

        JTR*~\cite{nishi2024jtr} &
        36.36 & 51.68 & 54.73 & 66.23 & 59.83 & 32.29 & 35.04 & 0.3516 & 0.7454 & 0.2457 & 0.3865 & -117.00 \\

        DiffusionMTL~\cite{diffusionmtl} &
        51.63 & 27.13 & 23.65 & 29.94 & 26.99 & n/a & n/a & n/a & n/a & n/a & n/a & n/a \\

        JTR~\cite{nishi2024jtr} &
        42.21 & 29.56 & 30.47 & 35.38 & 37.05 & n/a & n/a & n/a & n/a & n/a & n/a & n/a \\

        \textbf{\OURSSS{}} &
        \cellcolor{second}62.53 & \cellcolor{second}27.12 & \cellcolor{second}21.52 & \cellcolor{second}20.69 & \cellcolor{second}20.25 & \cellcolor{second}13.88 & \cellcolor{second}21.10 & \cellcolor{second}0.1611 & \cellcolor{second}0.5571 & \cellcolor{second}0.2021 & \cellcolor{second}0.2423 & \cellcolor{second}-13.24 \\

        \textbf{\OURS{}} &
        \cellcolor{best}67.98 & \cellcolor{best}25.39 & \cellcolor{best}19.30 & \cellcolor{best}19.11 & \cellcolor{best}18.93 & \cellcolor{best}9.44 & \cellcolor{best}14.53 & \cellcolor{best}0.1263 & \cellcolor{best}0.4382 & \cellcolor{best}0.2012 & \cellcolor{best}0.2366 & \cellcolor{best}+1.56 \\
        \bottomrule
    \end{tabular}}

    \caption{\textbf{In-domain model performance.} We compare the different approaches by evaluating on in-domain test sets.}
    \label{tab:in-domain}
    \vspace{-.5em}
\end{table}

\section{Task encoding}
\label{sec:appendix:task-encoding}
For unbounded quantities -- \depth{}, \opticalflow{}, and \sceneflow{} -- we adopt affine-invariant representations as in~\cite{marigold}.
Specifically, we apply independent linear scaling by mapping values from $[a,b]$ to $[-1, +1]$. Where $(a,b)$ are the 2nd/98th percentiles for \depth{} and min/max values for \opticalflow{} and \sceneflow{}.
For categorical tasks such as \semantic{} segmentation, the annotations are discrete, $\tasklabel_{\rm \text{seg}} \in [\![1,C]\!]^{H \times W}$, for $C$ classes in total. We define a mapping $f_{\text{seg}}: [\![1,C]\!] \to [-1, +1]^3$ assigning a unique RGB vector to each class. Following~\cite{careagaColorful}, we tone-map \albedo{} and \shading{} with a scalar scale, then clamp and remap the values to $[-1, +1]$.
After appropriate scaling, we standardize all shapes to match $\mapset{3}$ by repeating channels. For \depth{} and \shading{} (both in $\mapset{}$), the grayscale maps are repeated three times.
For \opticalflow{} ($\tasklabel_{\text{o-flow}} \in \mapset{2}$), we choose to repeat the horizontal flow once.
Finally \sceneflow{}, \normal{}, and \albedo{} already belong to $\mapset{3}$.

As a post-processing step for \semantic{} segmentation, we obtain final prediction $\hat{\tasklabel}_{\rm seg}$ by first decoding the predicted latent with  $\tasklabel'_{\rm seg} = \Dec(\hat{z}_{\rm seg})$, then applying nearest neighbors search in the RGB space for each pixel $(i, j)$: $\hat{\tasklabel}_{\rm seg}(i, j) = \operatornamewithlimits{\arg\min}_{c \in [\![1,C]\!]} \| \hat{\tasklabel}'_{\rm seg}(i, j) - f_{\text{seg}}(c) \|_2$. For \opticalflow{}, the prediction consists of the first two channels of the decoded latent. For \depth{} and \shading{}, we utilize the average of the three decoded channels. Other tasks do not require any post-processing.

\section{Task Visualization}
\label{sec:supp:visualization}
We detail here the visualization for each task. \Semantic{} segmentation maps employ the Cityscapes color scheme (see~\cref{tab:supp:vktocs_mapping_with_colors}). For \opticalflow{} $\mathbf{v}(u,v) = (v_x, v_y)$, we adopt the HSV color encoding from~\cite{flowcolor}, where the lateral flow vector's $(v_x, v_y)$ angle $\text{atan2}(-v_y, -v_x)$ determines hue and its magnitude $\|(v_x, v_y)\|_2$ defines saturation, as illustrated in~\cref{fig:flow-vis}. Similarly, \Sceneflow{} $\mathbf{v}(u,v) = (v_x, v_y, v_z)$ utilizes an HSV representation (cf.~\cref{fig:hsv}): its lateral component $(v_x, v_y)$ determines hue (angle: $\text{atan2}(-v_y, -v_x)$) and saturation (magnitude: $\|(v_x, v_y)\|_2$), while the depth flow component $v_z$ is inversely mapped to value (brightness), with image-specific scaling applied to both the magnitude and $v_z$. For \normal{} visualization, surface normal XYZ coordinates are directly mapped to RGB space. We predict outward-facing surface normals in OpenCV coordinate system following~\cite{bae2024dsine}. \Depth{} visualization involves mapping scale-invariant depth values to the spectral color map before display. Finally, \shading{} and \albedo{} are directly visualized as grayscale and RGB images, respectively.

\begin{figure}[h]
    \centering
    \begin{subfigure}[c]{0.48\linewidth}
        \centering
        \includegraphics[width=\linewidth]{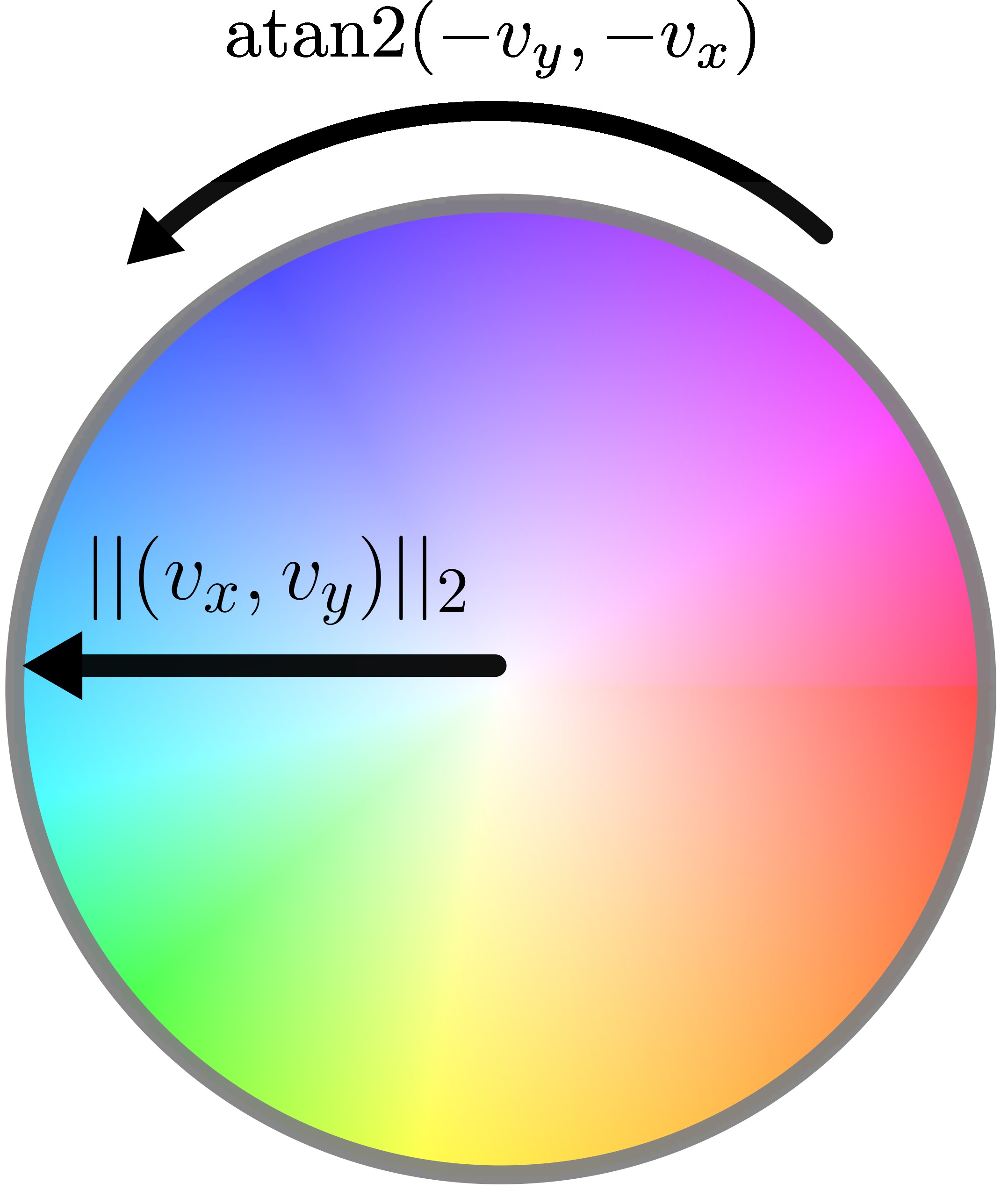}
        \caption{\Opticalflow{} color mapping.}
        \label{fig:flow-vis}
    \end{subfigure}\hfill
    \begin{subfigure}[c]{0.48\linewidth}
        \centering
        \includegraphics[width=\linewidth]{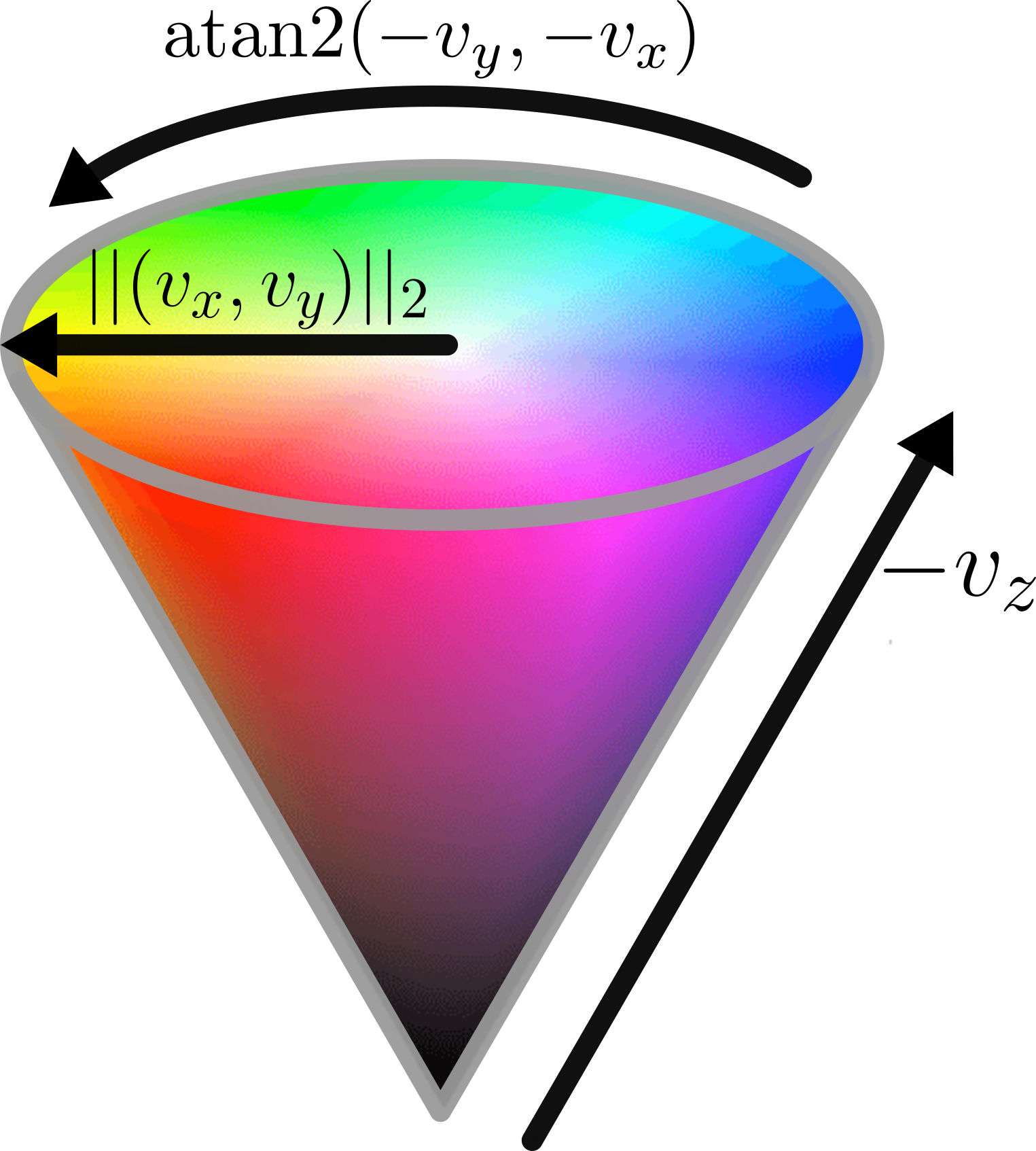}
        \caption{\Sceneflow{} color mapping.}
        \label{fig:hsv}
    \end{subfigure}
    \caption{\textbf{Flow color mappings.} We visualize the mapping used to visualize \subref{fig:flow-vis} \opticalflow{} and \subref{fig:hsv} \sceneflow{}.}
    \label{fig:flow-color-mapping}
\end{figure}

\section{Semantic class mapping}
\label{sec:supp:semantic-mapping}
To allow evaluation of \semantic{} on the Cityscapes dataset, our model is trained using only a subset of 8 classes, following the  \vkitti{}$\mapsto$\allowbreak\cs{} class mapping from~\cite{densemtl}, detailed in~\cref{tab:supp:vktocs_mapping_with_colors}.
\begin{table}[h]
	\centering
	\begin{tabular}{l l H | l l | c}
		\vkitti         & ours            & Color         & \cs           & ours            & Color \\
		\toprule
		terrain     & \textit{ignore}   & \colorIgnore  & road          & road              & \textcolor{mappedRoad}{$\blacksquare$} \\
		sky         & sky               & \colorSky     & sidewalk      & \textit{ignore}   & \textcolor{mappedIgnore}{$\blacksquare$} \\
		tree        & vegetation        & \colorVegetation & building      & building          & \textcolor{mappedBuilding}{$\blacksquare$} \\
		vegetation  & vegetation        & \colorVegetation & wall          & vegetation        & \textcolor{mappedVegetation}{$\blacksquare$} \\
		building    & building          & \colorBuilding & fence         & \textit{ignore}   & \textcolor{mappedIgnore}{$\blacksquare$} \\
		road        & road              & \colorRoad    & pole          & pole              & \textcolor{mappedPole}{$\blacksquare$} \\
		guardrail   & \textit{ignore}   & \colorIgnore  & light         & light             & \textcolor{mappedLight}{$\blacksquare$} \\
		sign        & sign              & \colorSign    & sign          & sign              & \textcolor{mappedSign}{$\blacksquare$} \\
		light       & light             & \colorLight   & vegetation    & vegetation        & \textcolor{mappedVegetation}{$\blacksquare$} \\
		pole        & pole              & \colorPole    & sky           & sky               & \textcolor{mappedSky}{$\blacksquare$} \\
		misc        & \textit{ignore}   & \colorIgnore  & person        & \textit{ignore}   & \textcolor{mappedIgnore}{$\blacksquare$} \\
		truck       & vehicle           & \colorVehicle & rider         & \textit{ignore}   & \textcolor{mappedIgnore}{$\blacksquare$} \\
		car         & vehicle           & \colorVehicle & car           & vehicle           & \textcolor{mappedVehicle}{$\blacksquare$} \\
		van         & vehicle           & \colorVehicle & bus           & vehicle           & \textcolor{mappedVehicle}{$\blacksquare$}\\
		            &                   &               & motorbike     & \textit{ignore}   & \textcolor{mappedIgnore}{$\blacksquare$} \\
		            &                   &               & bike          & \textit{ignore}   & \textcolor{mappedIgnore}{$\blacksquare$} \\
	\end{tabular}
	\caption{\textbf{Classes used for training.} We use a common set of 8 classes (ours), mapping together semantically similar classes of \vkitti{} for proper evaluation of our model on \cs.}
	\label{tab:supp:vktocs_mapping_with_colors}
\end{table}

\section{Image resolutions}
\label{sec:supp:resolution}
We train the model on datasets using the following pixel resolutions, $\texttt{height} {\times} \texttt{width}$: Hypersim ($288{\times}384$), FlyingThings3D ($268{\times}480$), VKITTI2 ($187{\times}621$).

\noindent For evaluations, we use the following resolutions: Cityscapes ($256{\times}512$), \diode~($384{\times}512$), KITTI ($176{\times}608$), MIDIntrinsics ($256{\times}384$).

\section{Additional qualitative results}
\label{sec:supp:qualitative}

In~\cref{fig:supp:qualitative} we report additional qualitative results of StableMTL, and the supplementary video further demonstrate the better of StableMTL compared to the existing baselines.

\section{Scalability of \OURS{}}
\label{sec:supp:sample-efficiency}
 On a H100 GPU, Stage~1 (single-stream) trains in 10hr and Stage~2 (multi-stream) in 20hr. During inference, all tasks are batch-predicted in parallel. As tasks increase 2\,$\rightarrow$\,7, compute (\,$\sim$2.5\,$\rightarrow$\,7.7 TFLOPs) and wall-clock (\,$\sim$0.095\,$\rightarrow$\,0.18s) grow linearly though the latter has a much smaller slope,as shown in~\cref{fig:flops_inference_vs_tasks}.

\begin{figure}[h]
    \centering
    \vspace{-1em}
    \centering
    \includegraphics[width=0.6\linewidth]{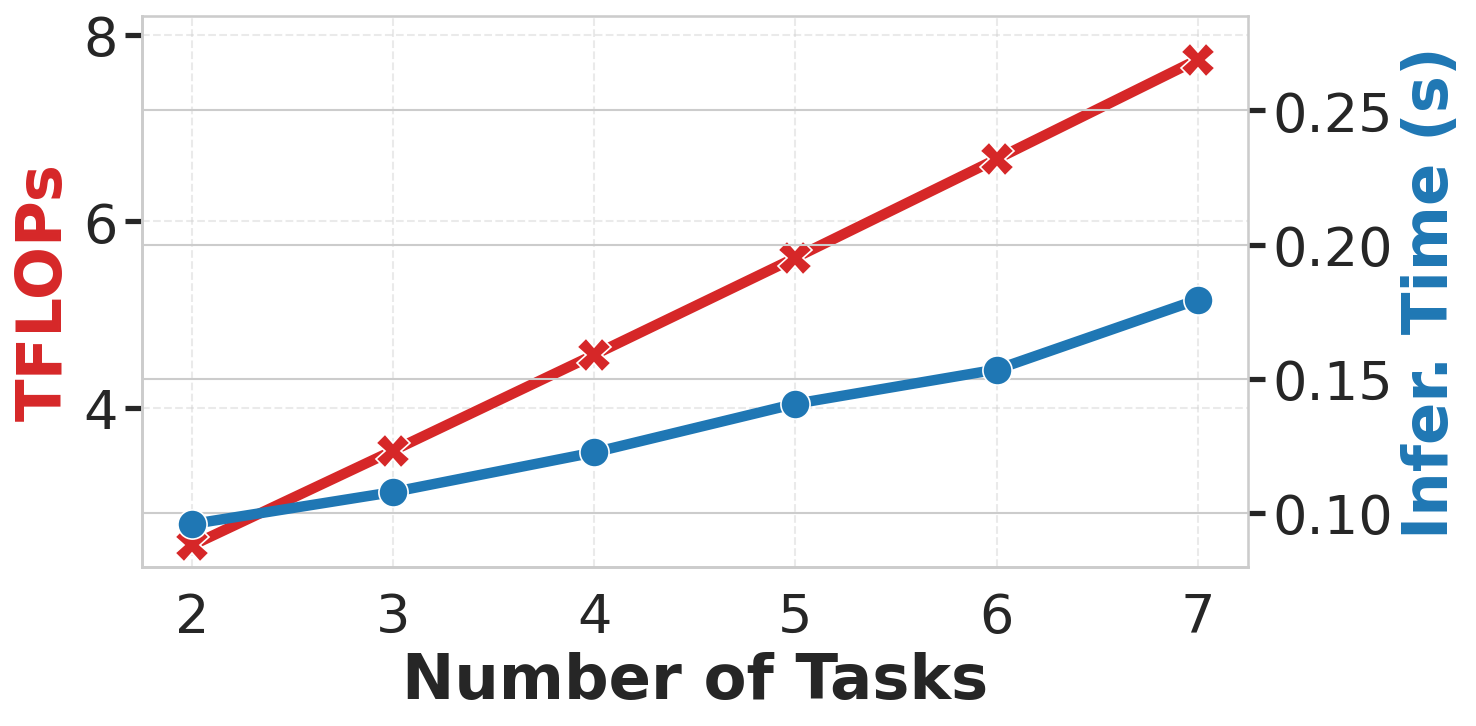}
    \caption{\textbf{Scalability of \OURS{}.} Observing complexity (TFLOPs) and inference time (s) demonstrates that the latter increases slower than complexity when increasing the number of tasks, therefore showcasing scalability of our method.}
    \label{fig:flops_inference_vs_tasks}
\end{figure}

\section{N-to-1 \vs N-to-N task-attention}
\label{sec:supp:n2n}
We conduct a comparison on (\Semantic, \Normal, \Depth), where independently encoded tasks are concatenated and jointly processed to enable full N-to-N  interactions, with supervision applied on all available labels. N-to-N incurs~50\% higher memory and degrades performance (KITTI AbsRel 13.92$\xrightarrow{}$15.72, Normal MAE 21.88$\xrightarrow{}$22.95), especially for semantics (mIoU 51.03$\xrightarrow{}$32.71) which is further dominated, likely due to its smaller gradients (Fig. {\textcolor{red}{6b}}).

\begin{figure}
    \centering

        \resizebox{0.7\columnwidth}{!}{
            \setlength{\tabcolsep}{3pt}
            \begin{tabular}{lcccc}
            \toprule
            \textbf{Method} &
            \rotatebox{0}{\textbf{\shortstack{Time\\(ms)}}} &
            \rotatebox{0}{\textbf{\shortstack{TFLOPs\\{\textcolor{white}{x}}}}} &
            \rotatebox{0}{\textbf{\shortstack{Memory\\(GB)}}} &
            \rotatebox{0}{\textbf{\shortstack{$\deltam{}$\\(\%)}}} \\
            \midrule
            JTR*               & 0.006   & 0.07      & 0.92 & -106.87      \\
            DiffusionMTL*      & 0.110 & 2.89 & 25.69 & -78.76 \\
            \textbf{\OURSSS{}} & 0.086  & 6.08 & 6.93 & -1.57  \\
            \textbf{\OURS{}}   & 0.179 & 7.73 & 11.75 & +4.78 \\
            \bottomrule
            \end{tabular}

        }
        \captionof{table}{\textbf{Inference cost.}}
        \label{tab:inference_vs_baselines}
\end{figure}

\section{Compute cost}
\label{sec:supp:compute}
\cref{tab:inference_vs_baselines} reports H100 inference latency, TFLOPs, and GPU memory, for a $256\!\times\!256$ image and 7 tasks. We adapt DiffusionMTL to scale to 7 tasks and upgrade the backbone (RN18$\xrightarrow{}$RN101) to boost performance.
JTR is kept unchanged because of its unconventional network.
\OURSSS{} runs in 0.086s and \OURS{} in 0.179s for 7 tasks.
Latent decoding drives 56\% of costs; the U-Net takes only 21\% as it operates in 4-channel latent space ($\approx\!64$ times smaller than image-space).
\OURSSS{} is 22\% faster than DiffusionMTL
, and \OURS{} uses about half the memory (11.8 \vs 25.7GB) thanks to parallelized computation, despite higher TFLOPs.

\begin{figure*}
	\centering
	\footnotesize
	\setlength{\tabcolsep}{0pt}
	\renewcommand{\arraystretch}{0}
	\newcommand{\shift}{7.7ex}
	\newcommand{\prompt}{\cellcolor{gray!40}}
	\resizebox{\linewidth}{!}{
		\begin{tabular}{c@{\hspace{1.5pt}} p{1.8cm} cS{.2ex} cS{.2ex}cS{.2ex}cS{.2ex}cS{.2ex}cS{.2ex}cS{.2ex}cS{.2ex}}
            & \ & {\scriptsize (for flow only)} & & & & & &  &
            \\[0.2ex]
            & \makecell{Input} & Input 2 & \Semantic & \Normal & \Depth & \Opticalflow & \Sceneflow & \Shading & \Albedo
            \\[0.2ex]

            \verti{4.5ex}{\textbf{Cityscapes}} & \includegraphics[width=\imgsize]{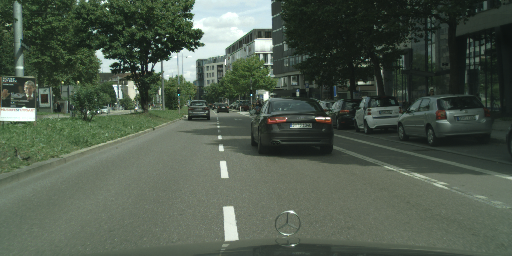}
            & \includegraphics[width=\imgsize]{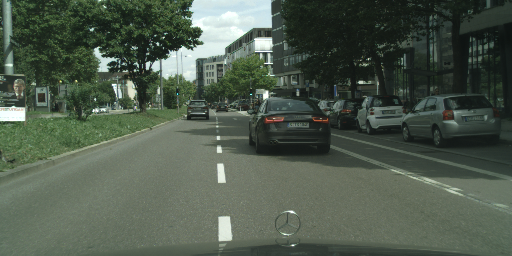}
            & \includegraphics[width=\imgsize]{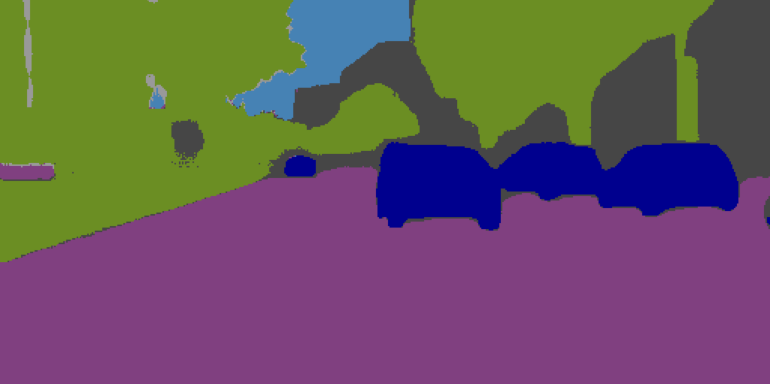}
            & \includegraphics[width=\imgsize]{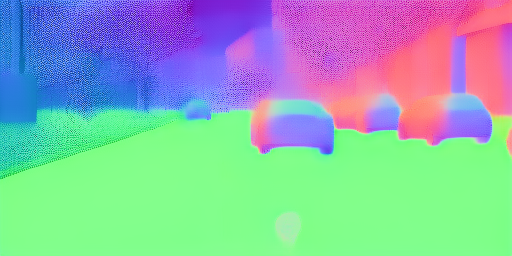}
            & \includegraphics[width=\imgsize]{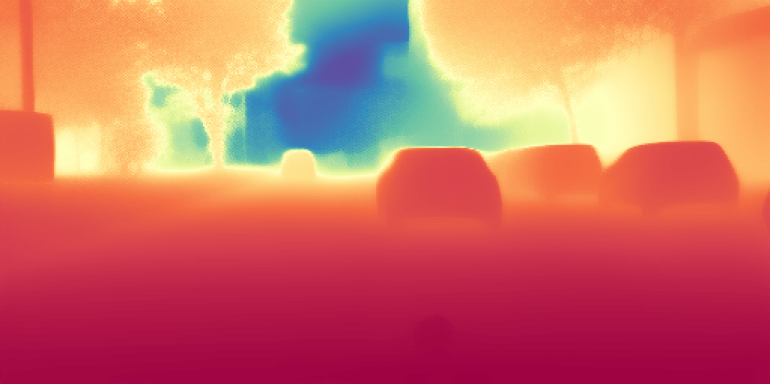}
            & \includegraphics[width=\imgsize]{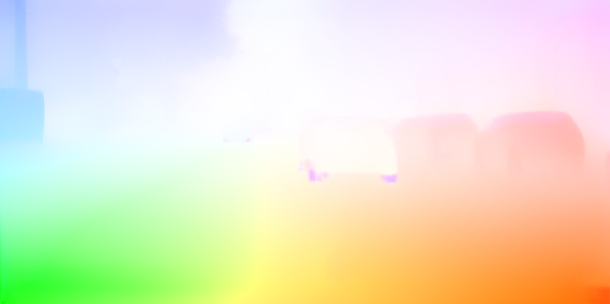}
            & \includegraphics[width=\imgsize]{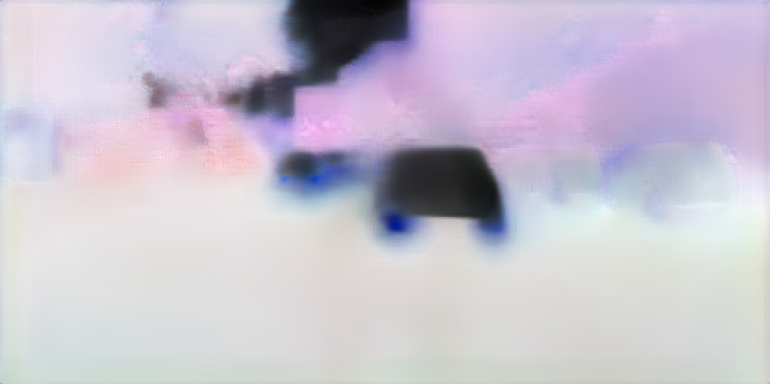}
            & \includegraphics[width=\imgsize]{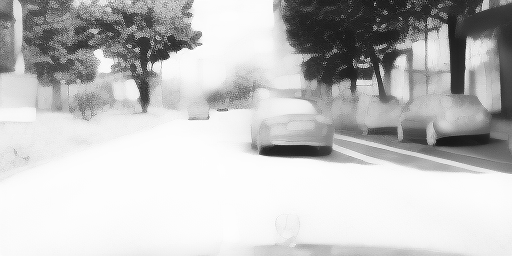}
            & \includegraphics[width=\imgsize]{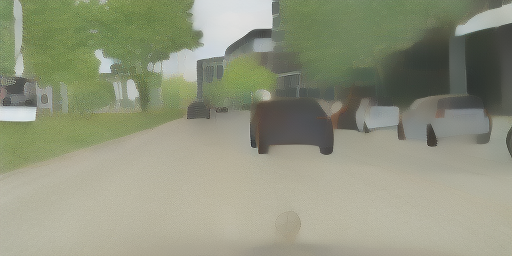}
            \\[0.2ex]
            & \includegraphics[width=\imgsize]{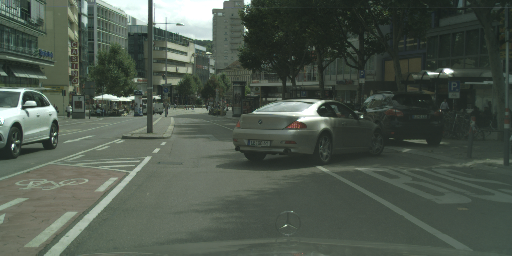}
            & \includegraphics[width=\imgsize]{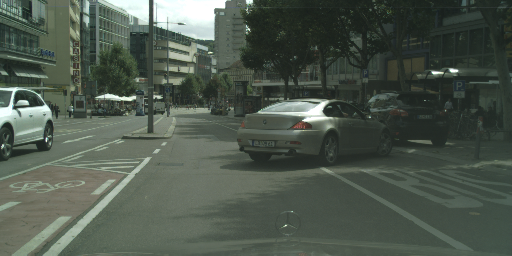}
            & \includegraphics[width=\imgsize]{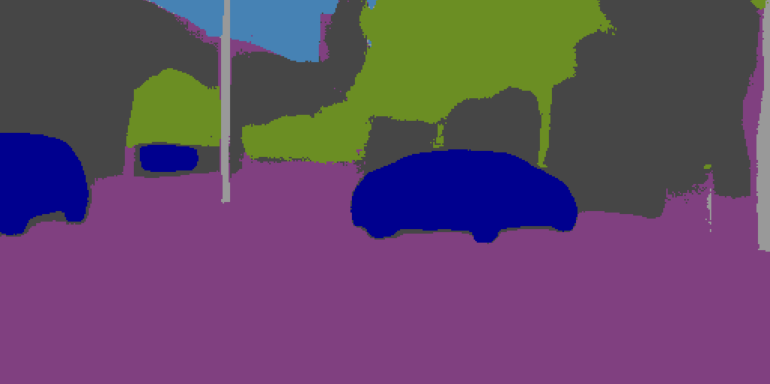}
            & \includegraphics[width=\imgsize]{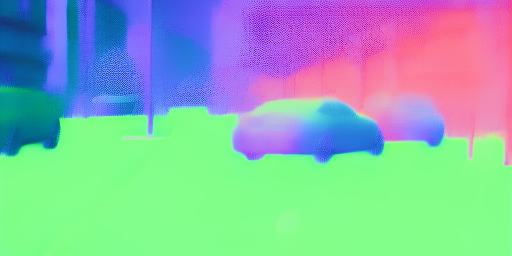}
            & \includegraphics[width=\imgsize]{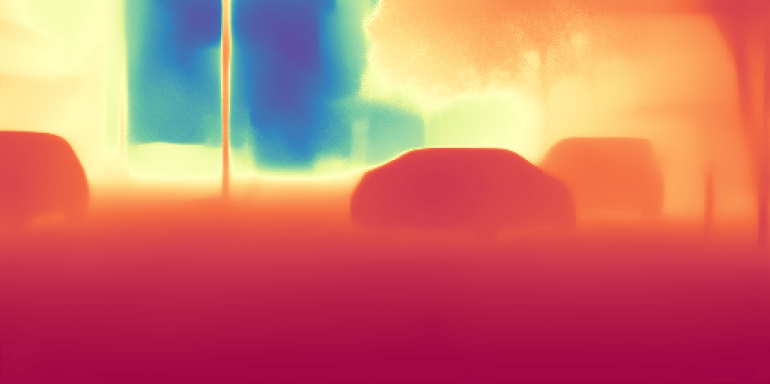}
            & \includegraphics[width=\imgsize]{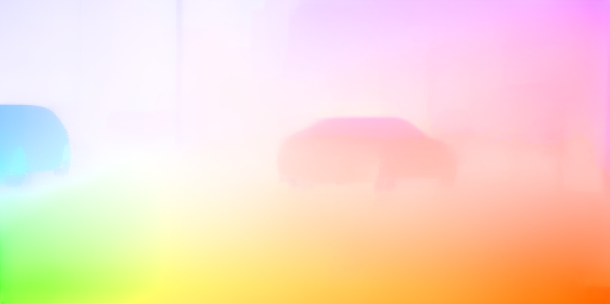}
            & \includegraphics[width=\imgsize]{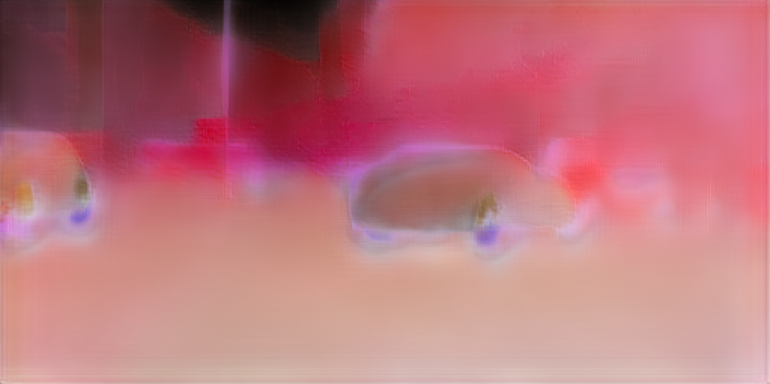}
            & \includegraphics[width=\imgsize]{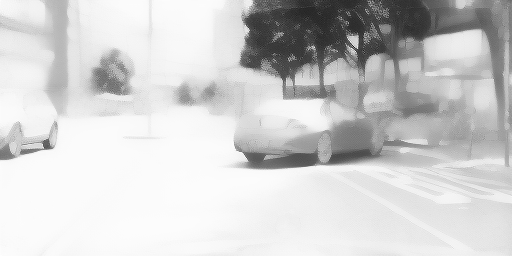}
            & \includegraphics[width=\imgsize]{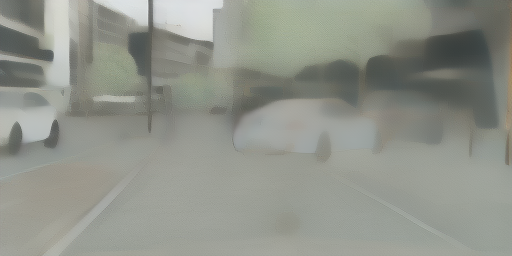}
            \\[0.6ex]

            \verti{3ex}{\textbf{KITTI}} & \includegraphics[width=\imgsize]{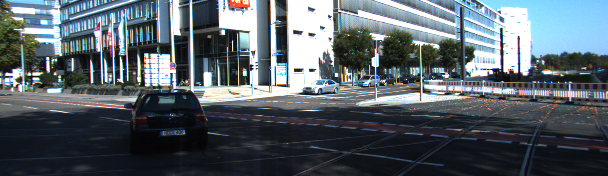}
            & \includegraphics[width=\imgsize]{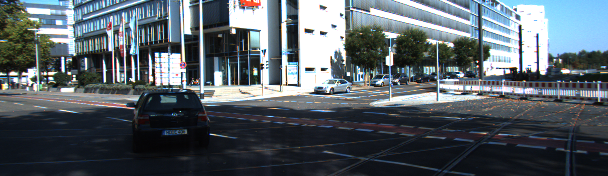}
            & \includegraphics[width=\imgsize]{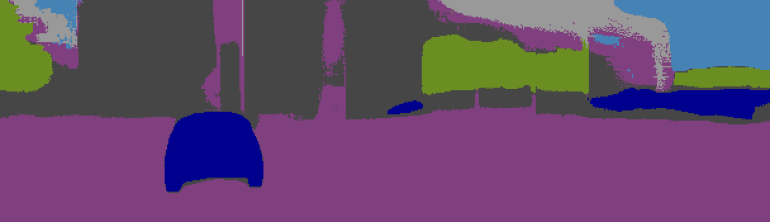}
            & \includegraphics[width=\imgsize]{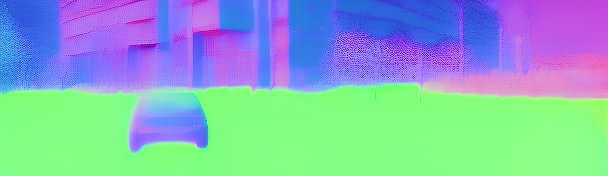}
            & \includegraphics[width=\imgsize]{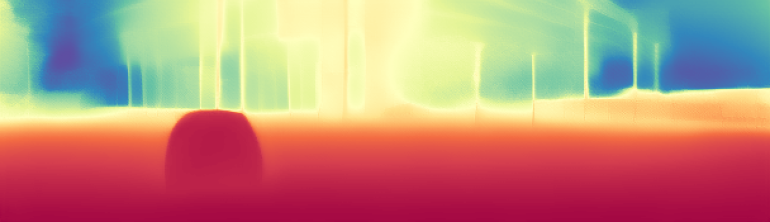}
            & \includegraphics[width=\imgsize]{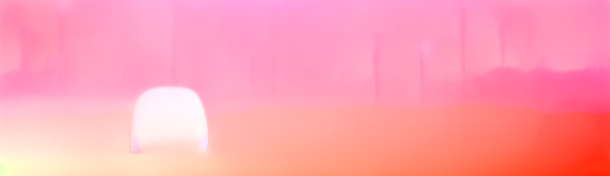}
            & \includegraphics[width=\imgsize]{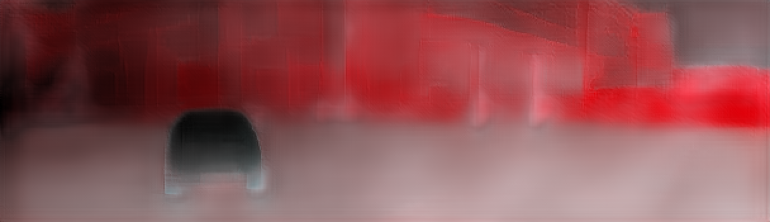}
            & \includegraphics[width=\imgsize]{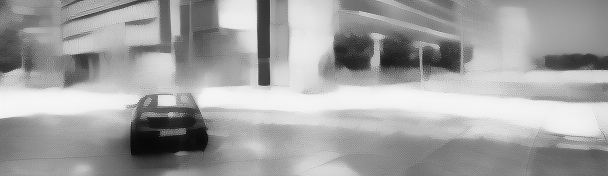}
            & \includegraphics[width=\imgsize]{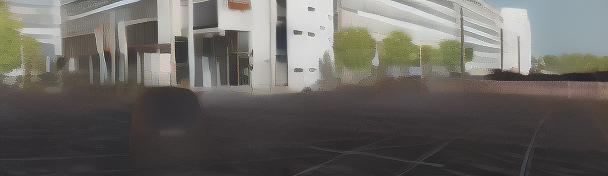}
            \\[0.2ex]

            & \includegraphics[width=\imgsize]{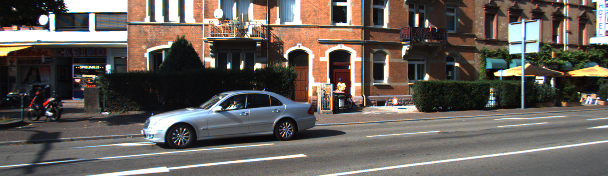}
            & \includegraphics[width=\imgsize]{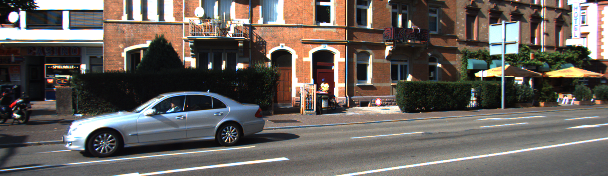}
            & \includegraphics[width=\imgsize]{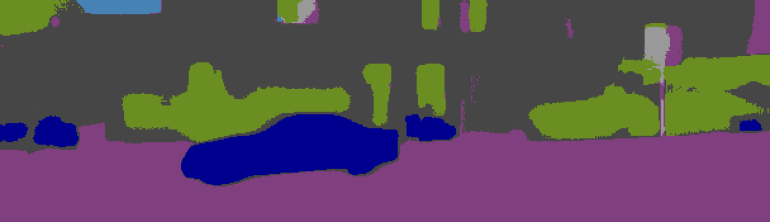}
            & \includegraphics[width=\imgsize]{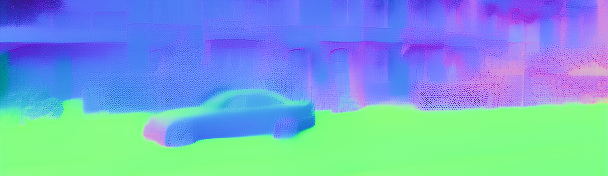}
            & \includegraphics[width=\imgsize]{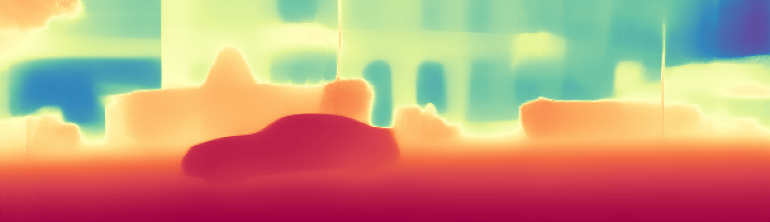}
            & \includegraphics[width=\imgsize]{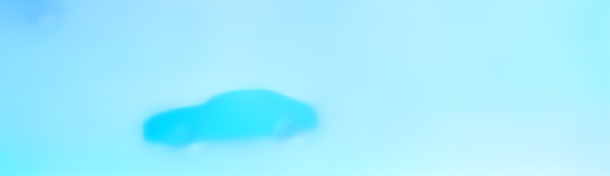}
            & \includegraphics[width=\imgsize]{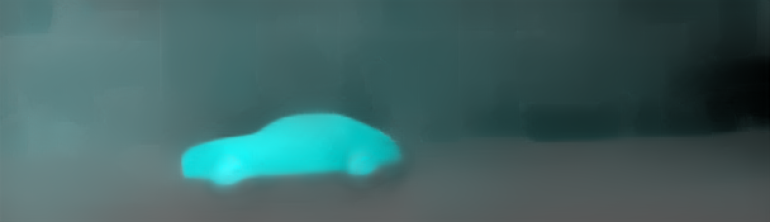}
            & \includegraphics[width=\imgsize]{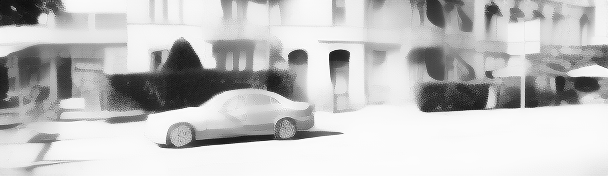}
            & \includegraphics[width=\imgsize]{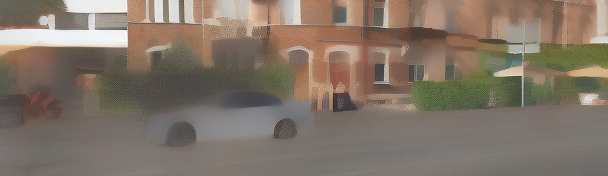}
            \\[0.6ex]

            \verti{3.5ex}{\textbf{Waymo}} & \includegraphics[width=\imgsize]{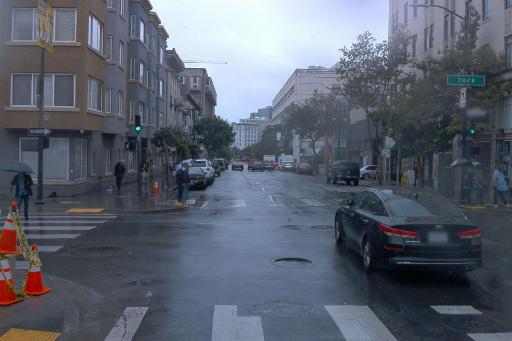}
            & \includegraphics[width=\imgsize]{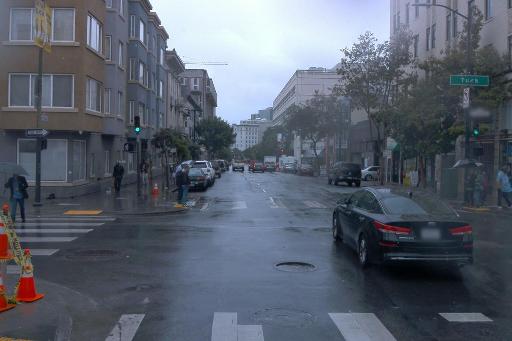}
            & \includegraphics[width=\imgsize]{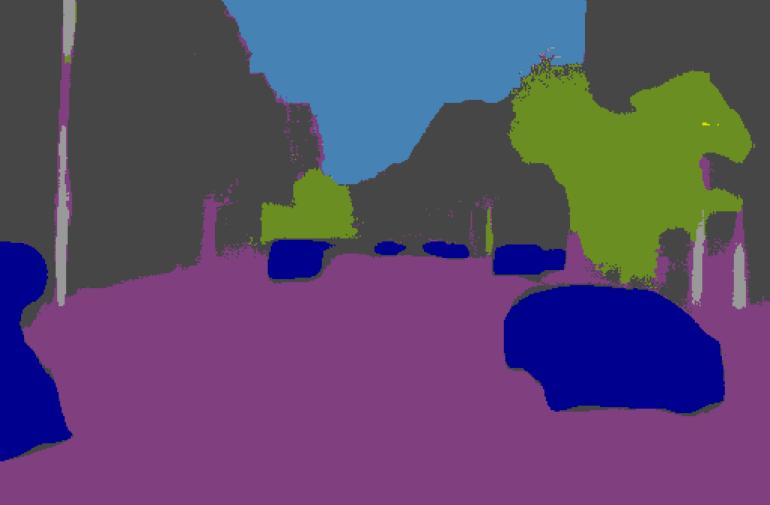}
            & \includegraphics[width=\imgsize]{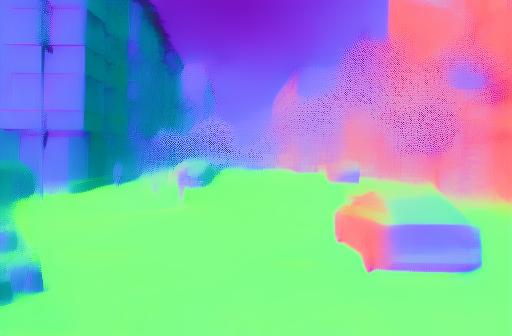}
            & \includegraphics[width=\imgsize]{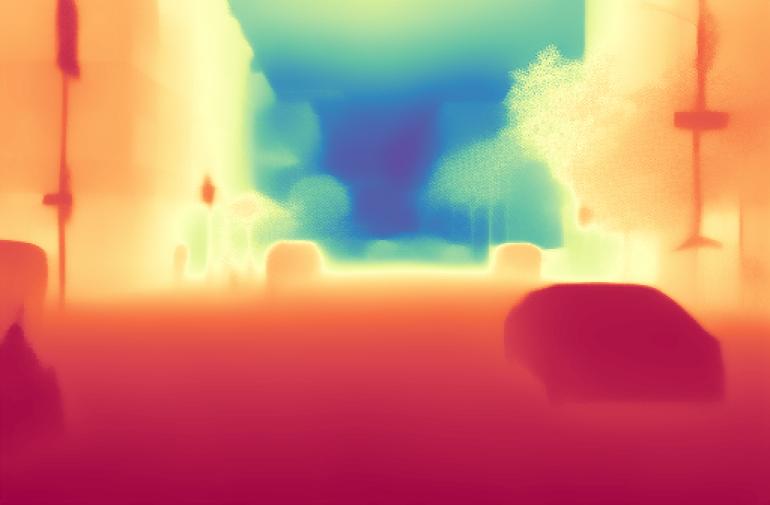}
            & \includegraphics[width=\imgsize]{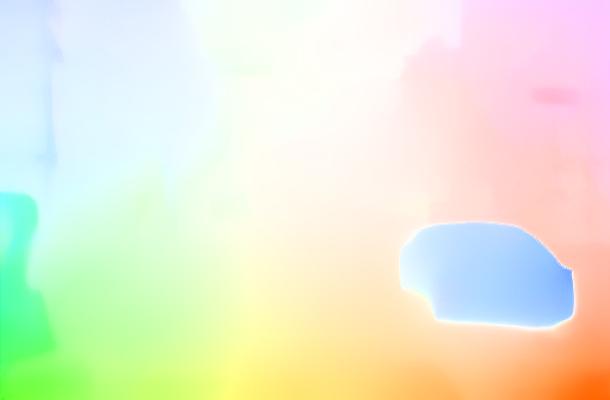}
            & \includegraphics[width=\imgsize]{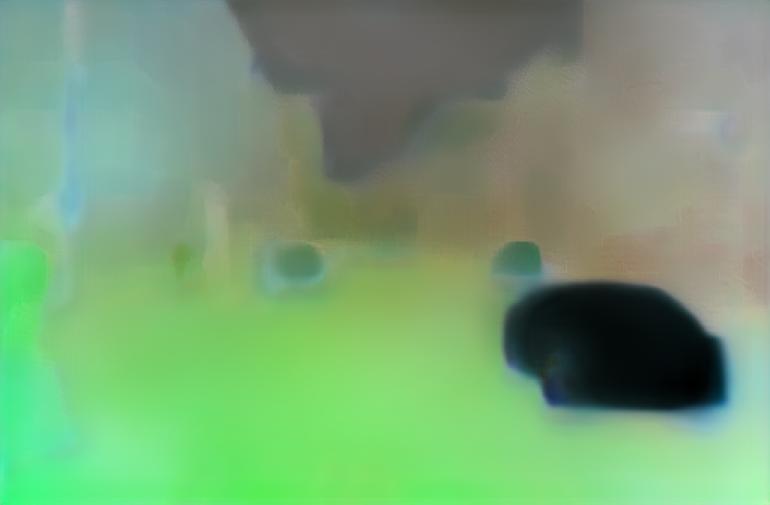}
            & \includegraphics[width=\imgsize]{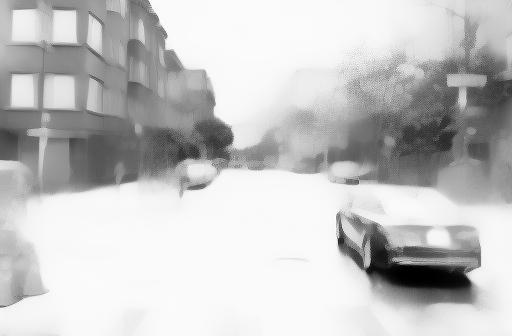}
            & \includegraphics[width=\imgsize]{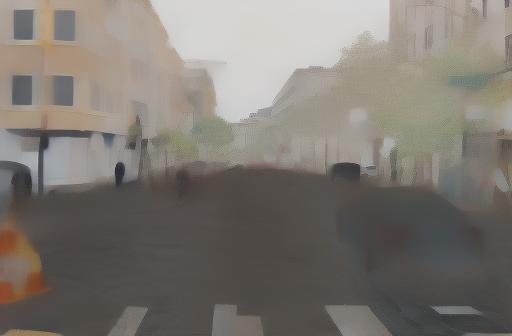}
            \\[0.2ex]

            & \includegraphics[width=\imgsize]{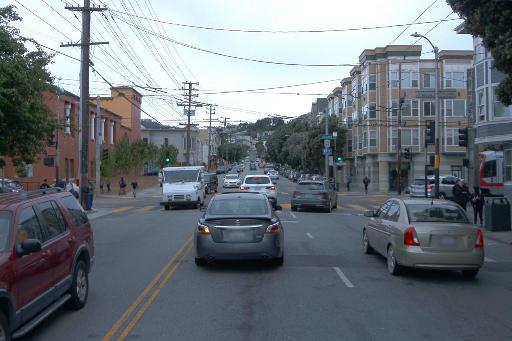}
            & \includegraphics[width=\imgsize]{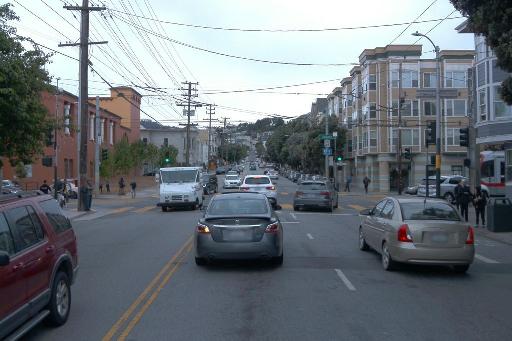}
            & \includegraphics[width=\imgsize]{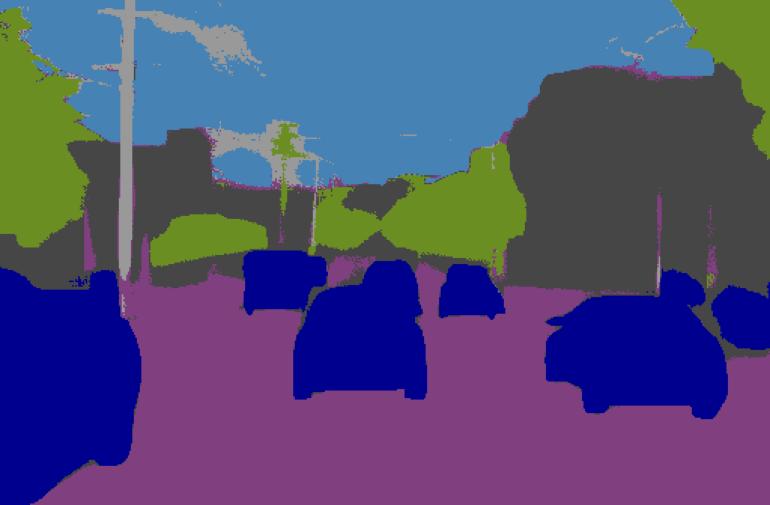}
            & \includegraphics[width=\imgsize]{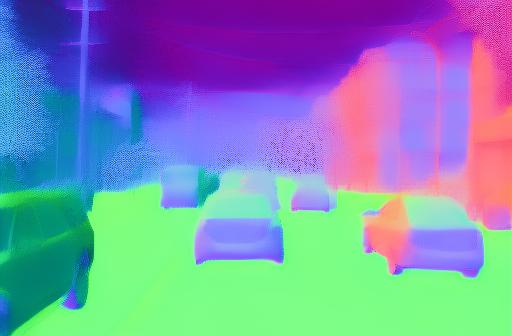}
            & \includegraphics[width=\imgsize]{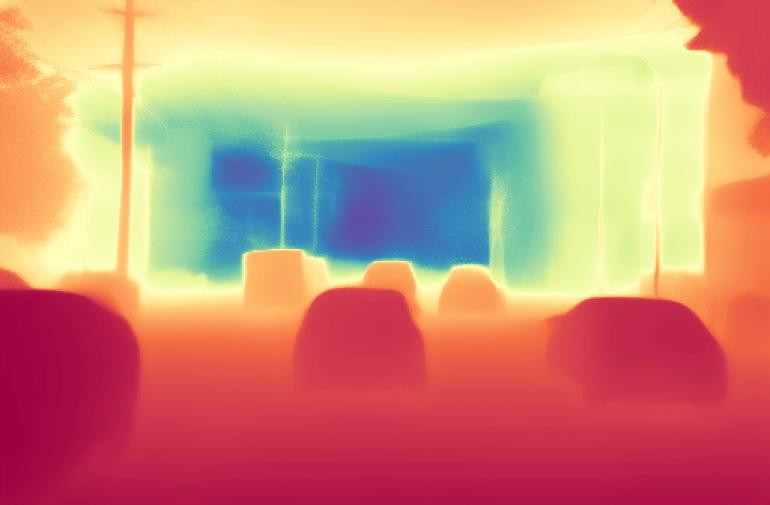}
            & \includegraphics[width=\imgsize]{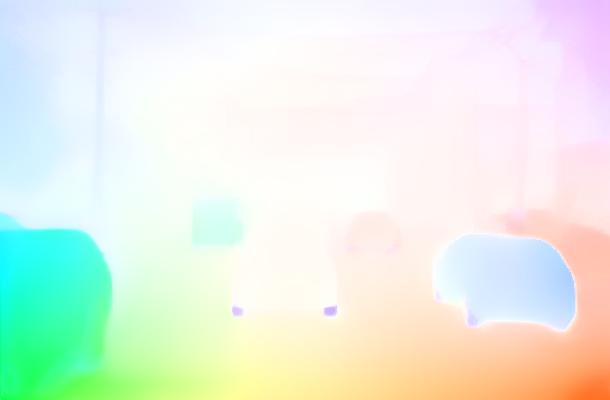}
            & \includegraphics[width=\imgsize]{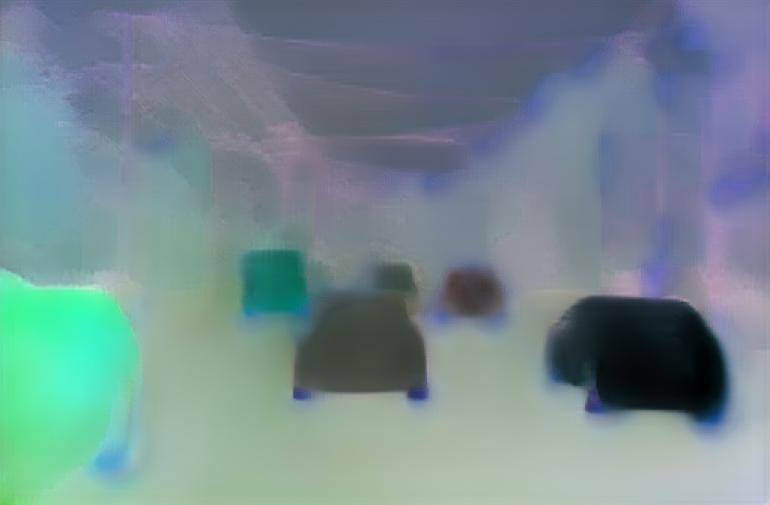}
            & \includegraphics[width=\imgsize]{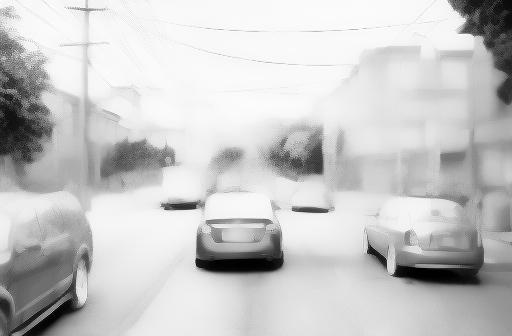}
            & \includegraphics[width=\imgsize]{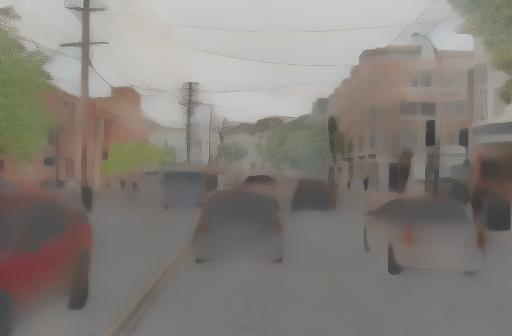}
            \\[0.6ex]

            \verti{5ex}{\textbf{Argoverse2}} & \includegraphics[width=\imgsize]{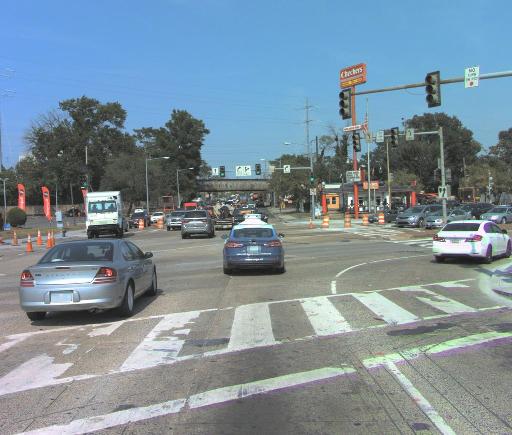}
            & \includegraphics[width=\imgsize]{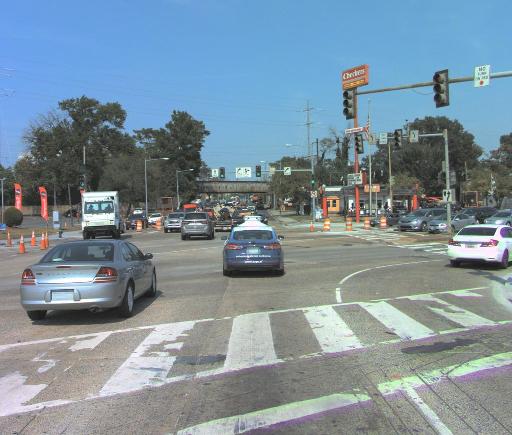}
            & \includegraphics[width=\imgsize]{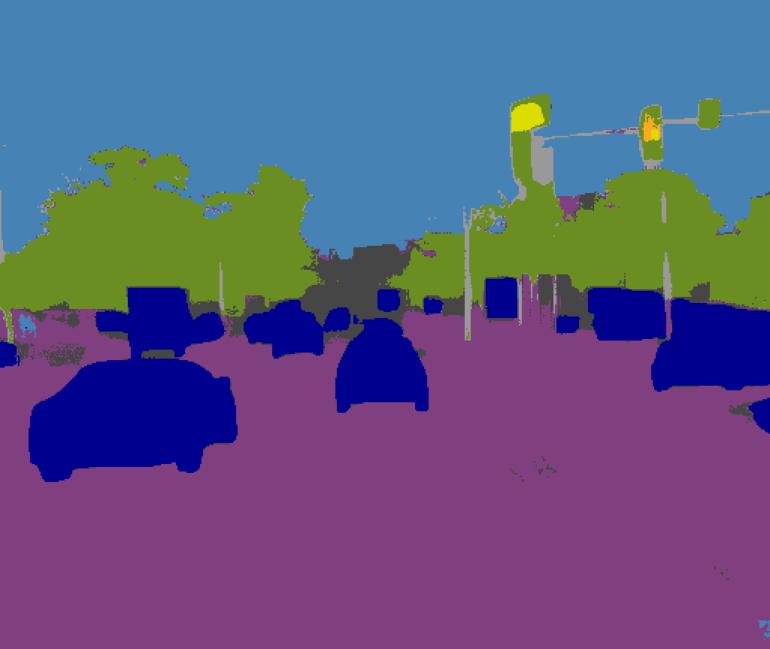}
            & \includegraphics[width=\imgsize]{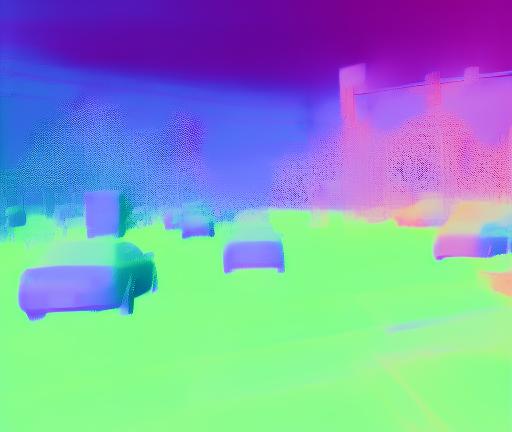}
            & \includegraphics[width=\imgsize]{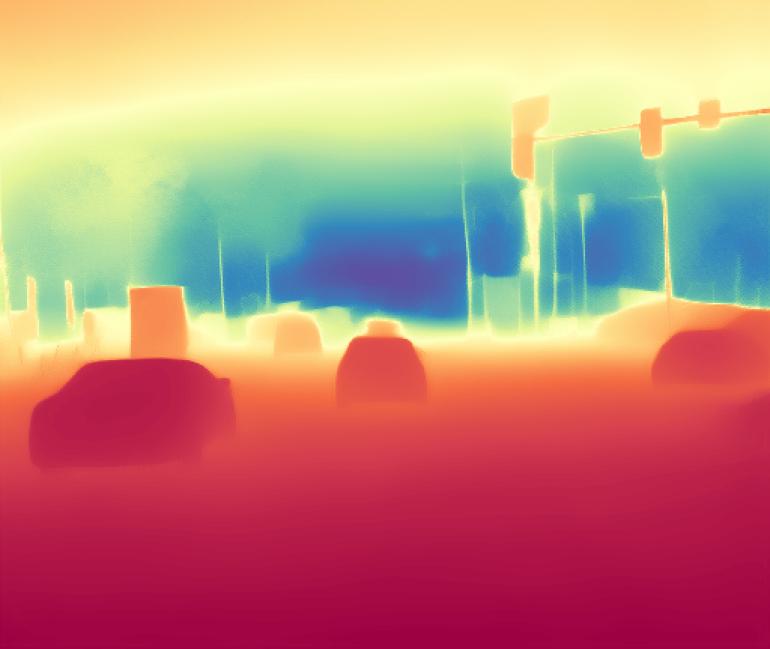}
            & \includegraphics[width=\imgsize]{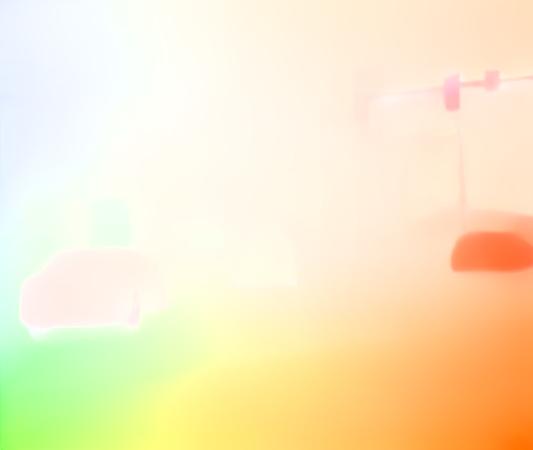}
            & \includegraphics[width=\imgsize]{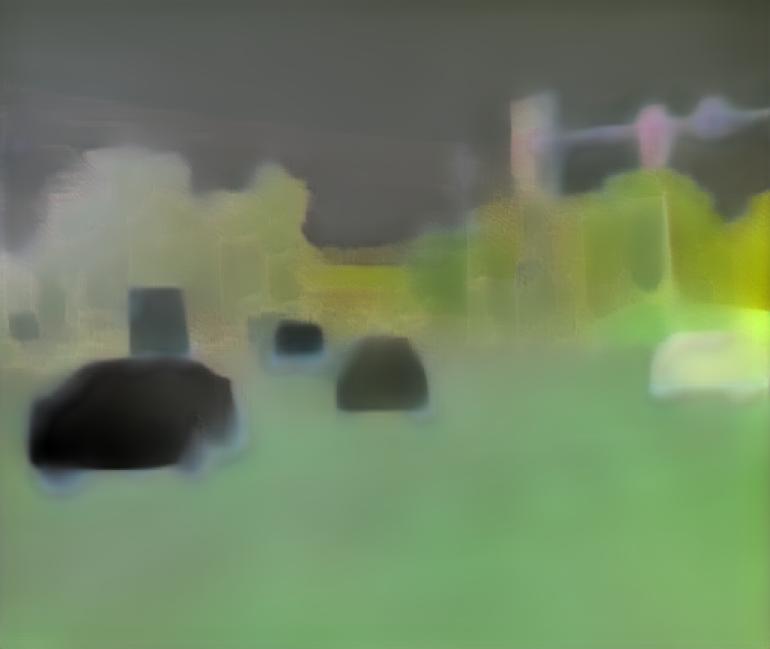}
            & \includegraphics[width=\imgsize]{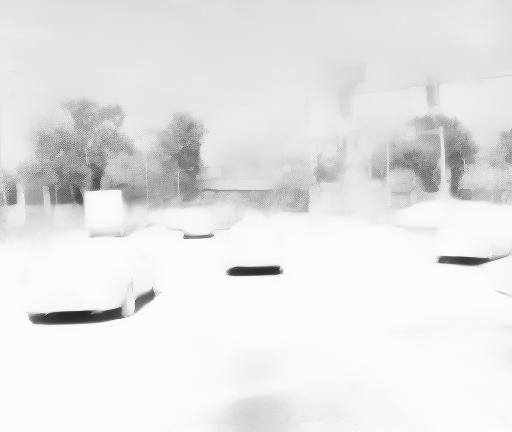}
            & \includegraphics[width=\imgsize]{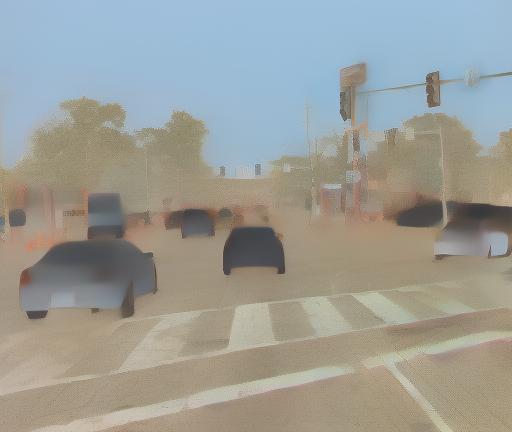}
            \\[0.2ex]

            & \includegraphics[width=\imgsize]{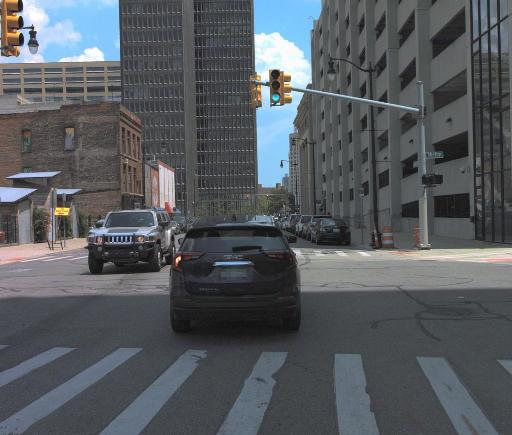}
            & \includegraphics[width=\imgsize]{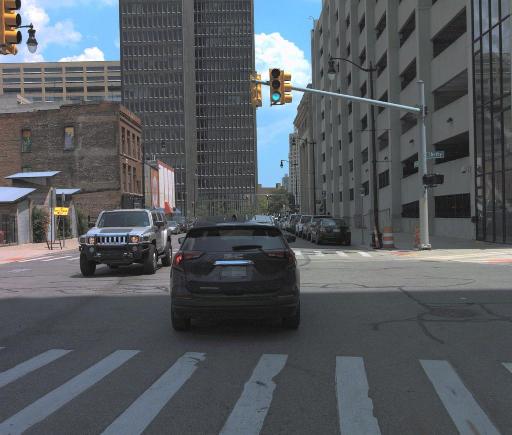}
            & \includegraphics[width=\imgsize]{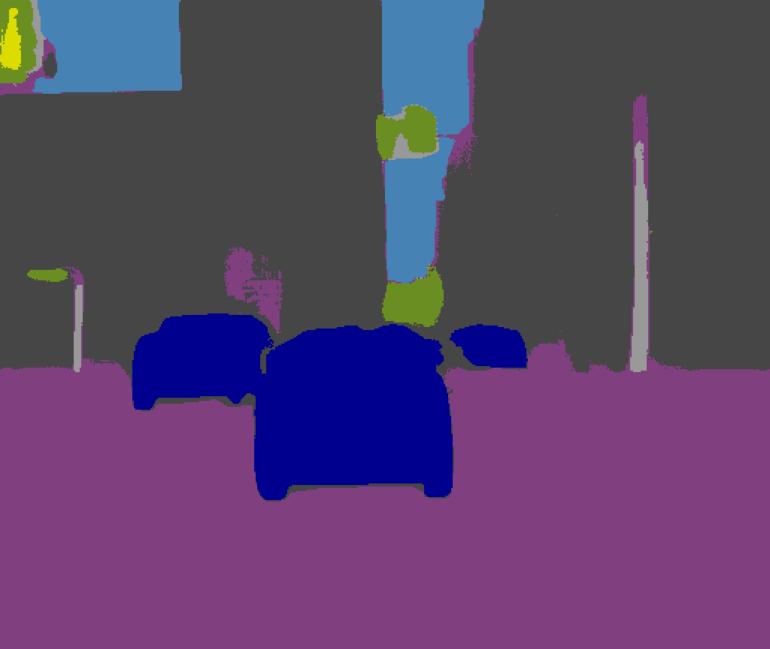}
            & \includegraphics[width=\imgsize]{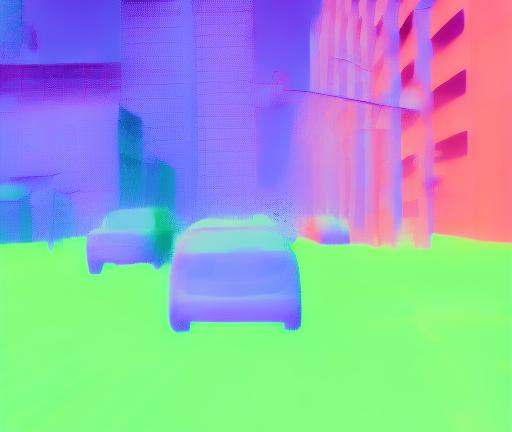}
            & \includegraphics[width=\imgsize]{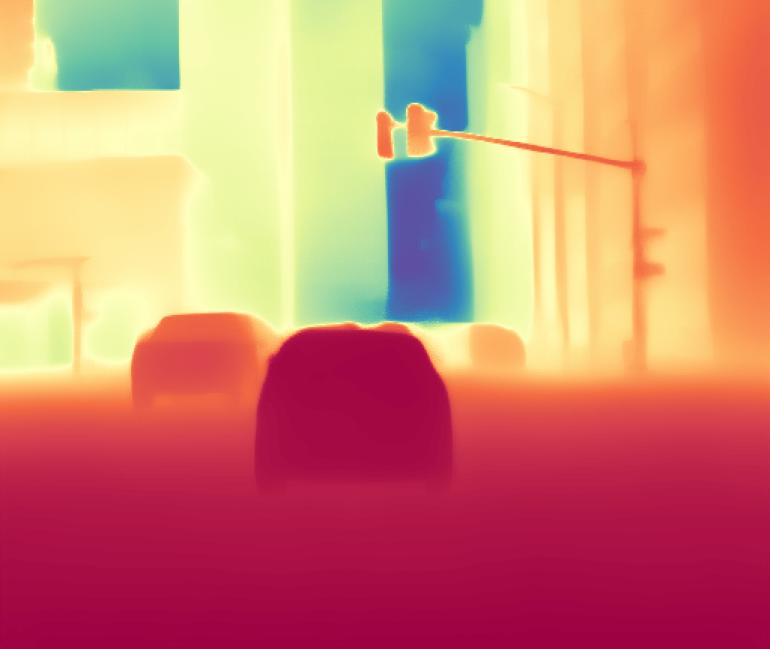}
            & \includegraphics[width=\imgsize]{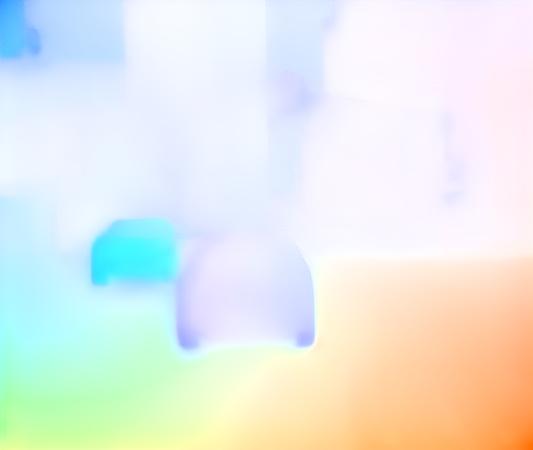}
            & \includegraphics[width=\imgsize]{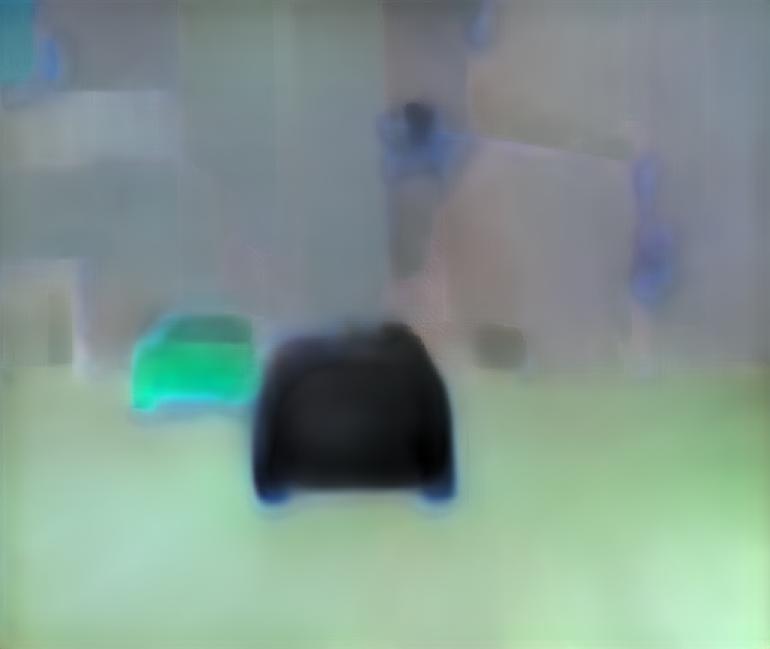}
            & \includegraphics[width=\imgsize]{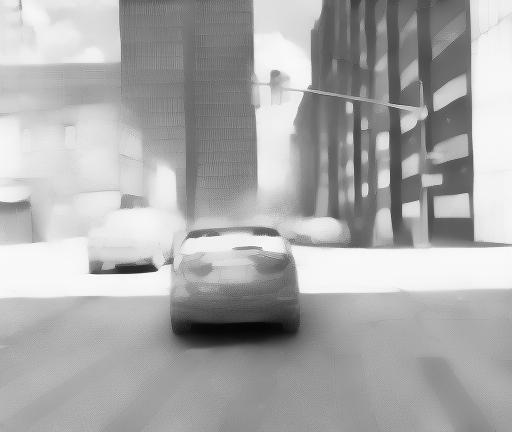}
            & \includegraphics[width=\imgsize]{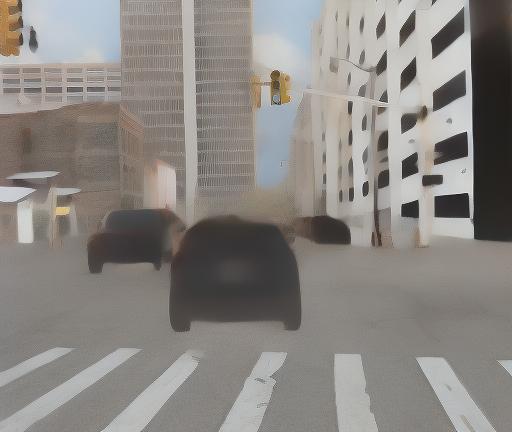}
            \\[0.6ex]

            \verti{3ex}{\textbf{DDAD}} & \includegraphics[width=\imgsize]{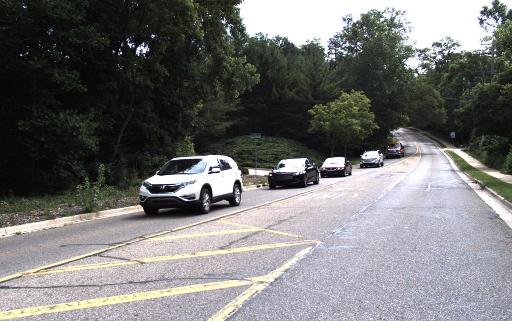}
            & \includegraphics[width=\imgsize]{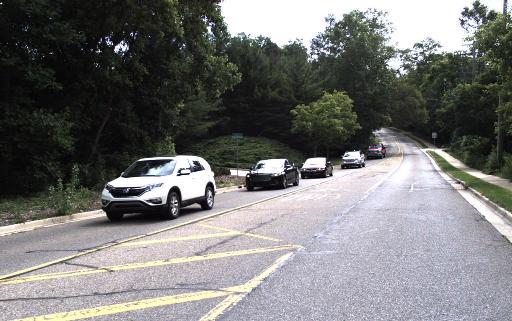}
            & \includegraphics[width=\imgsize]{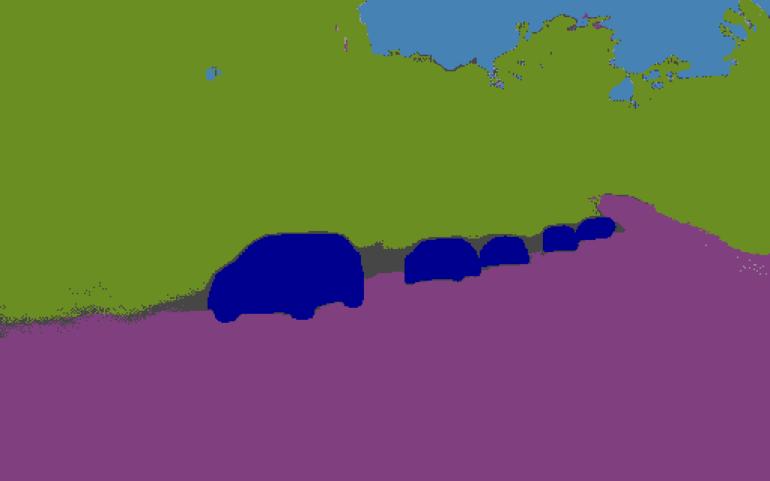}
            & \includegraphics[width=\imgsize]{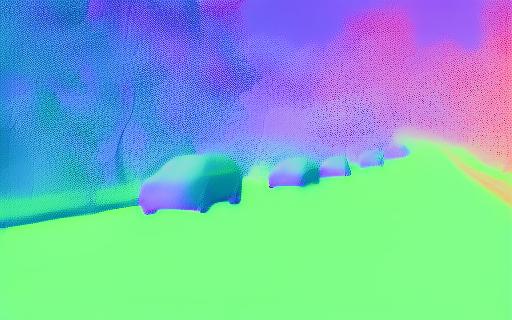}
            & \includegraphics[width=\imgsize]{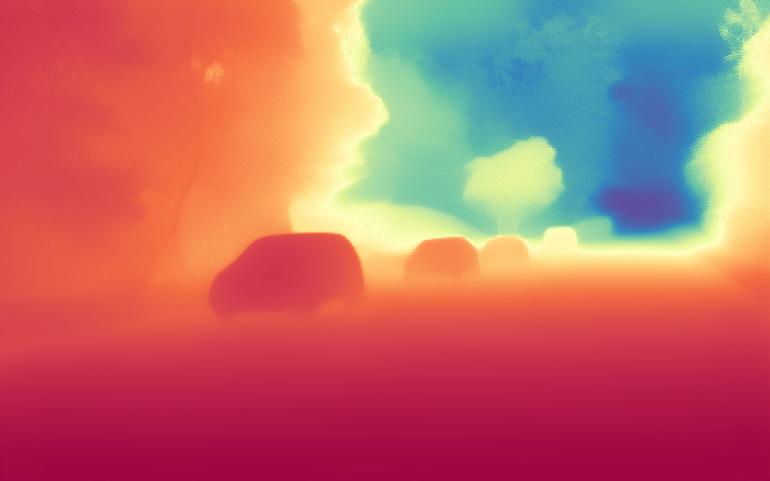}
            & \includegraphics[width=\imgsize]{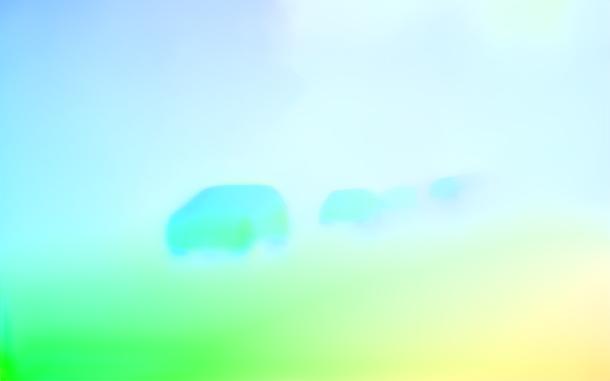}
            & \includegraphics[width=\imgsize]{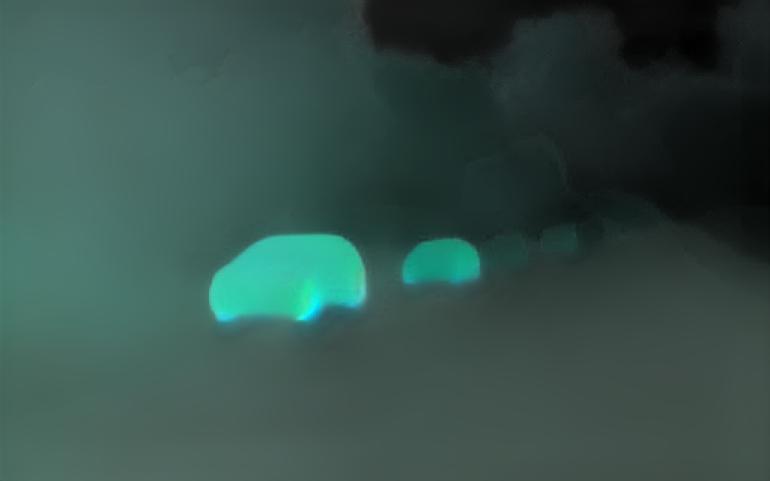}
            & \includegraphics[width=\imgsize]{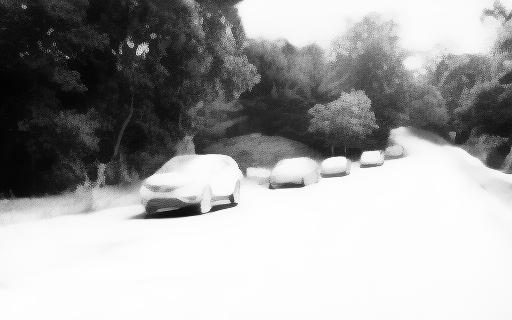}
            & \includegraphics[width=\imgsize]{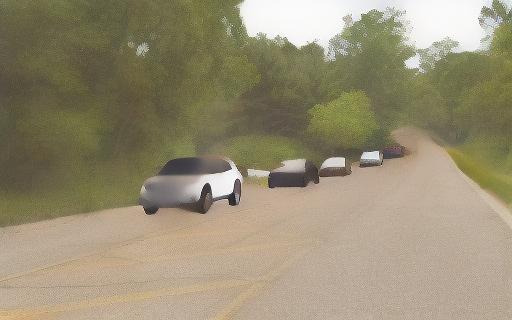}
            \\[0.2ex]

            & \includegraphics[width=\imgsize]{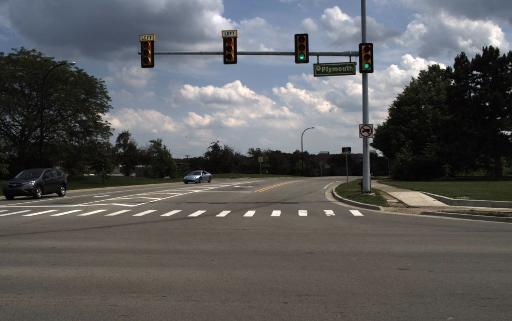}
            & \includegraphics[width=\imgsize]{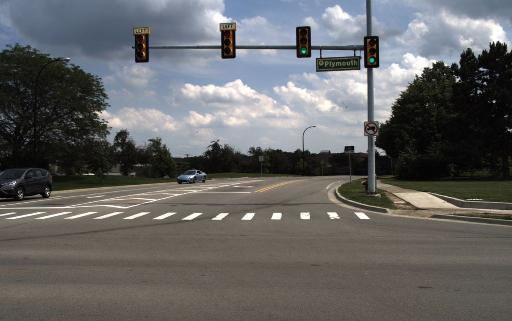}
            & \includegraphics[width=\imgsize]{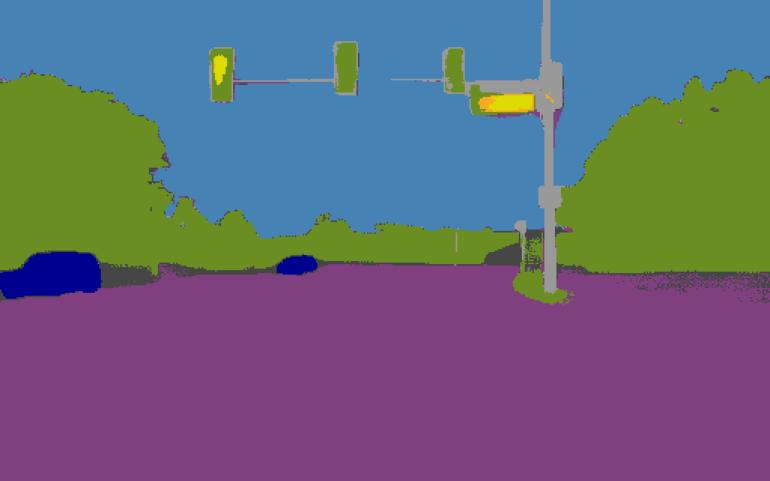}
            & \includegraphics[width=\imgsize]{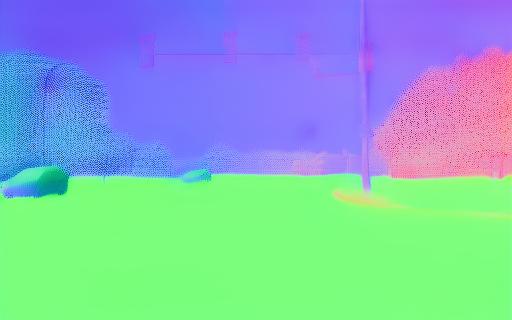}
            & \includegraphics[width=\imgsize]{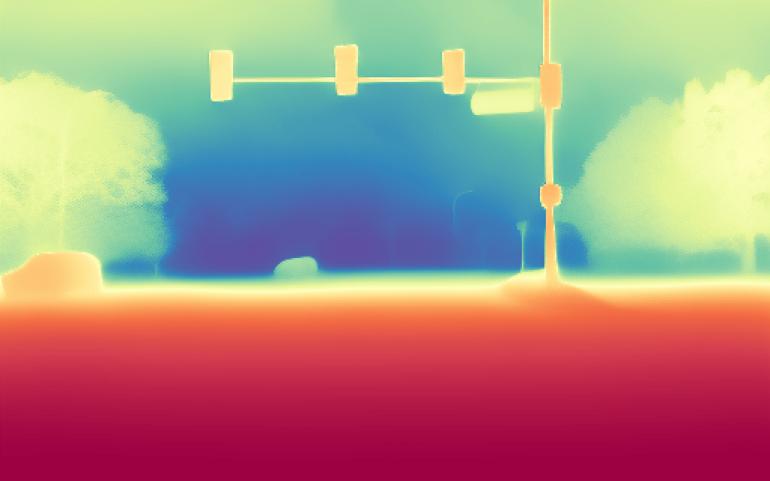}
            & \includegraphics[width=\imgsize]{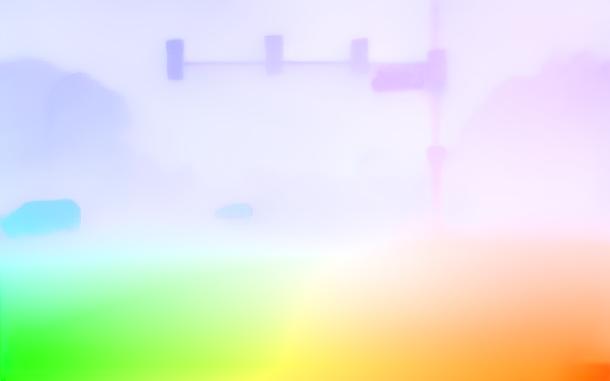}
            & \includegraphics[width=\imgsize]{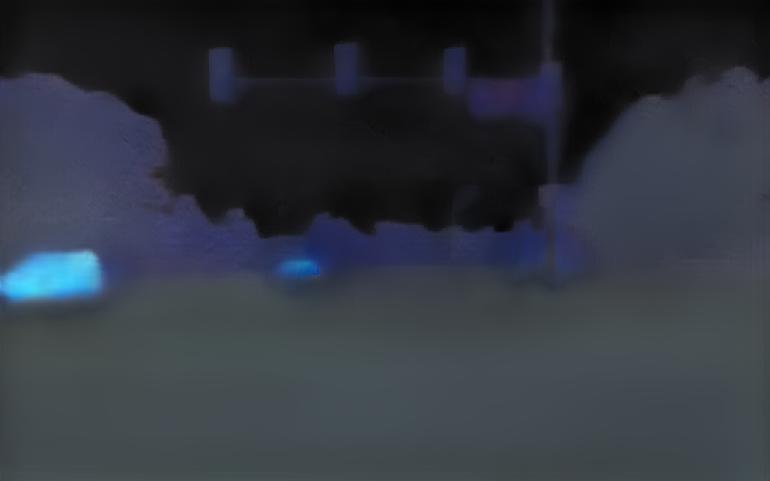}
            & \includegraphics[width=\imgsize]{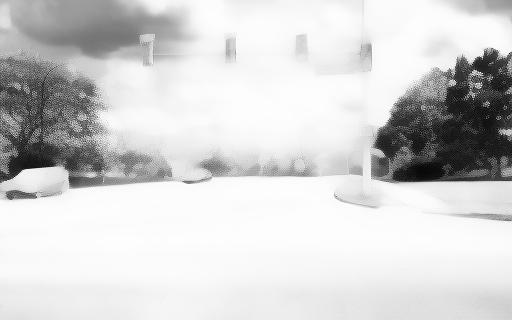}
            & \includegraphics[width=\imgsize]{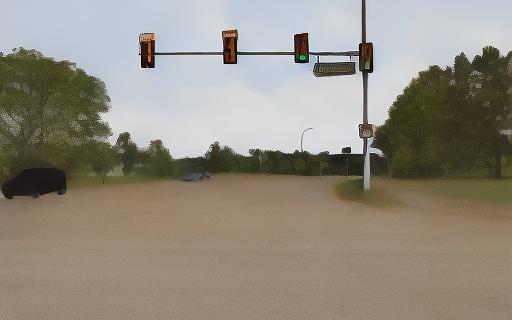}
            \\[0.6ex]

            \verti{5ex}{\textbf{PandaSet}} & \includegraphics[width=\imgsize]{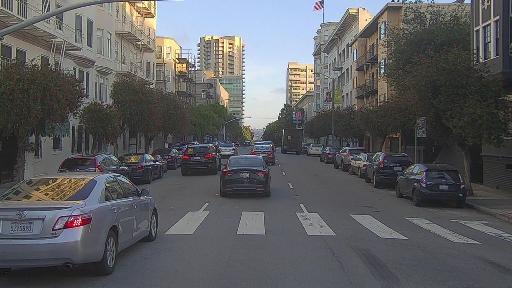}
            & \includegraphics[width=\imgsize]{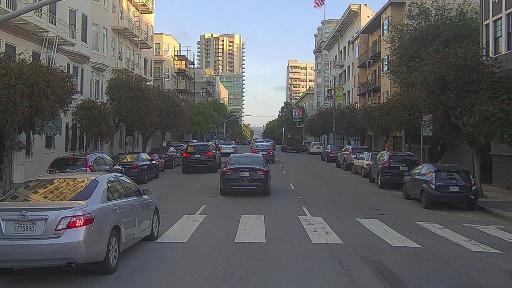}
            & \includegraphics[width=\imgsize]{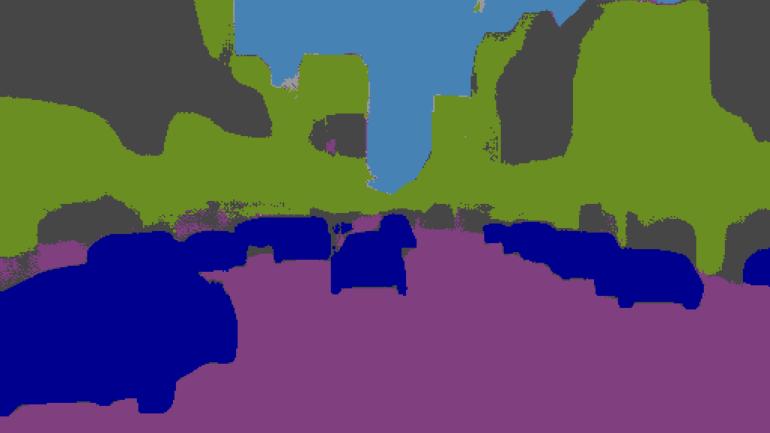}
            & \includegraphics[width=\imgsize]{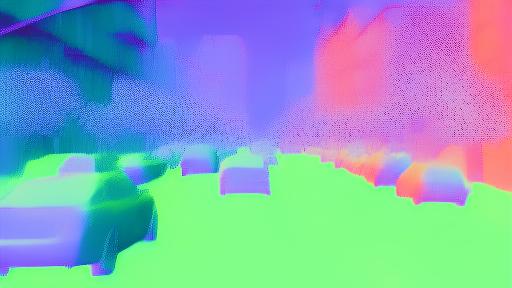}
            & \includegraphics[width=\imgsize]{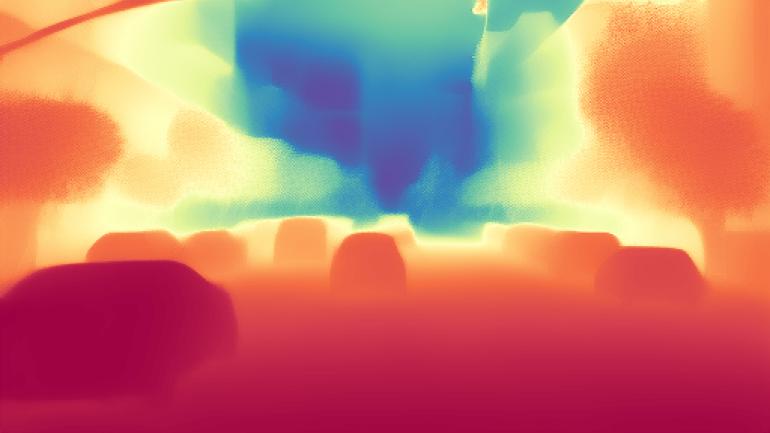}
            & \includegraphics[width=\imgsize]{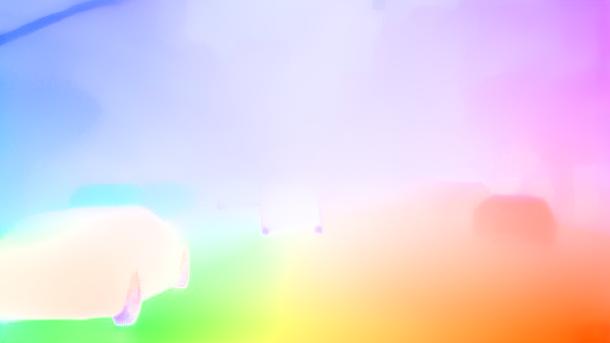}
            & \includegraphics[width=\imgsize]{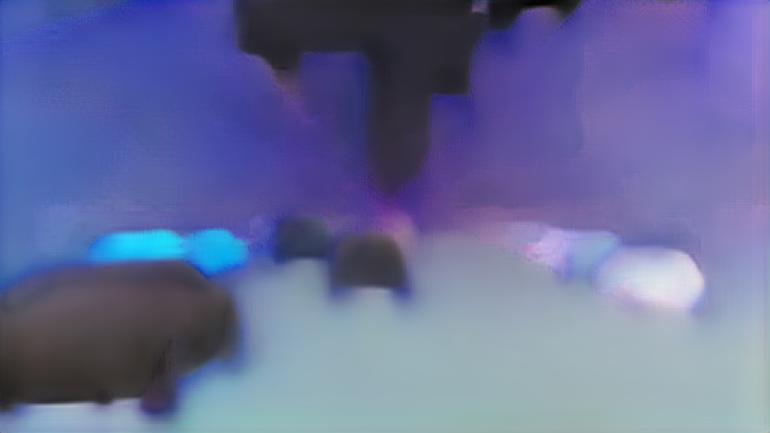}
            & \includegraphics[width=\imgsize]{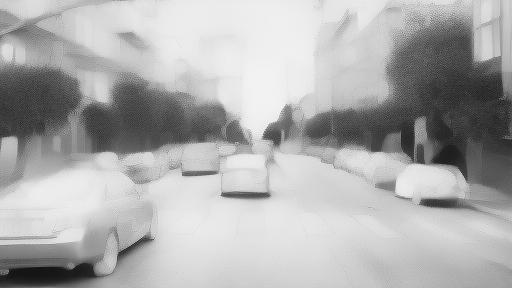}
            & \includegraphics[width=\imgsize]{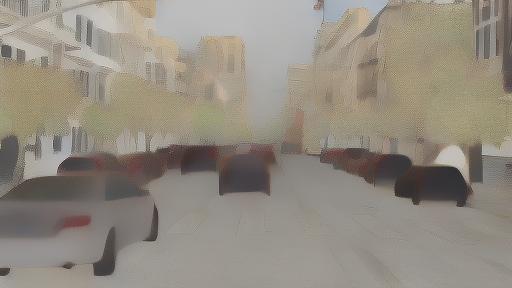}
            \\[0.2ex]

            & \includegraphics[width=\imgsize]{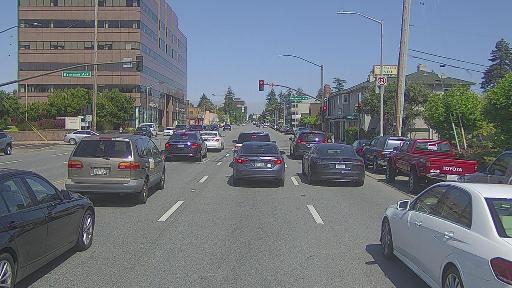}
            & \includegraphics[width=\imgsize]{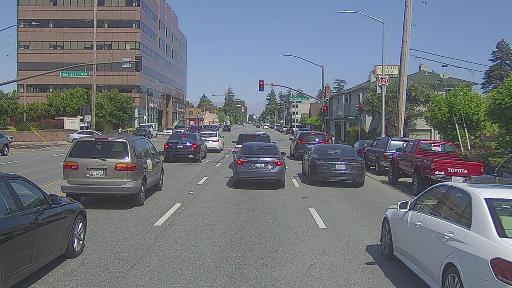}
            & \includegraphics[width=\imgsize]{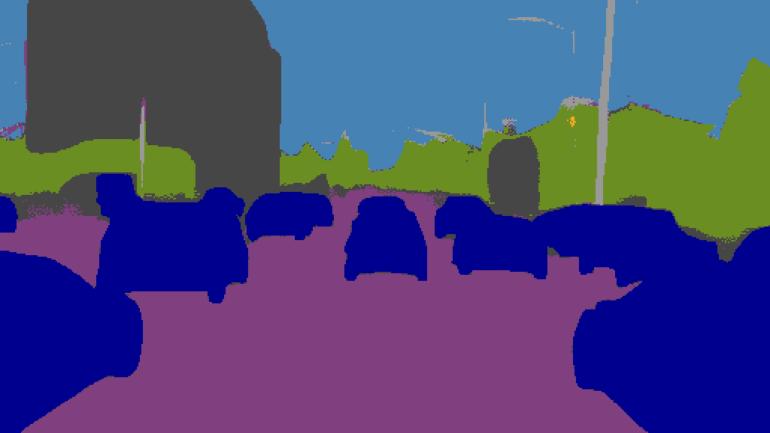}
            & \includegraphics[width=\imgsize]{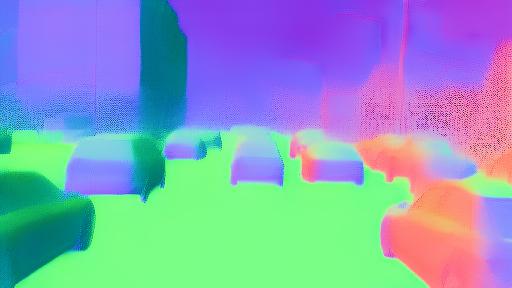}
            & \includegraphics[width=\imgsize]{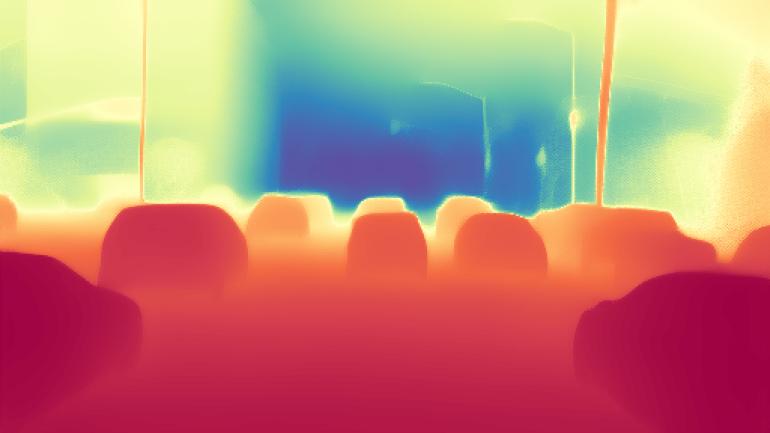}
            & \includegraphics[width=\imgsize]{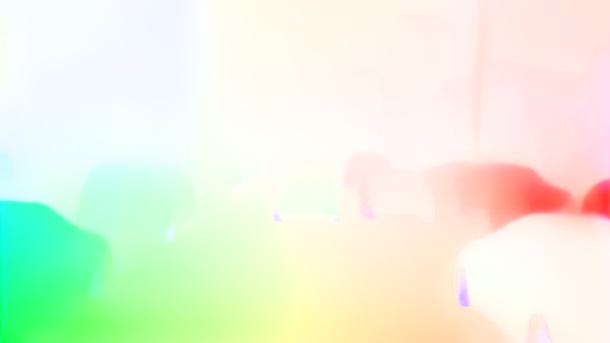}
            & \includegraphics[width=\imgsize]{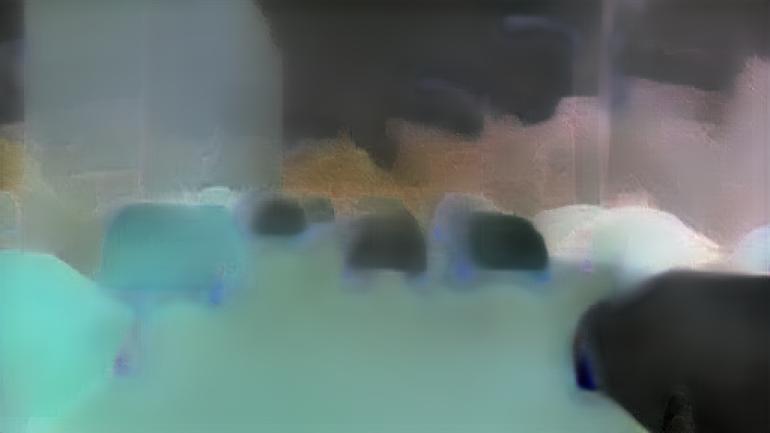}
            & \includegraphics[width=\imgsize]{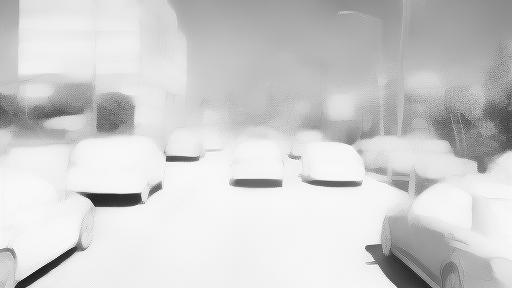}
            & \includegraphics[width=\imgsize]{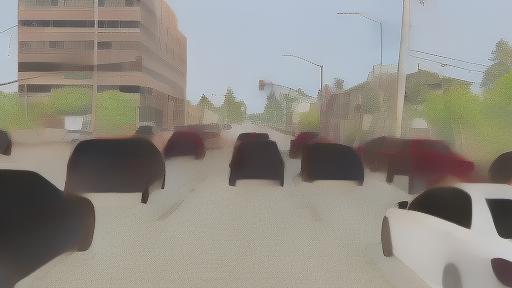}
            \\[0.6ex]

            \verti{5ex}{\textbf{Youtube-VOS}}  & \teaserrowasterix{youtubevos}{bfa904a786_00200}\\[0.2ex]
            & \teaserrowasterix{youtubevos}{0ad4b1255c_00070}\\[0.2ex]
            & \teaserrowasterix{youtubevos}{9c81364fc6_00040}\\[0.6ex]

            \verti{-0.6ex}{\scalebox{0.9}{\textbf{DAVIS}}}
            & \teaserrowasterix{davis}{motocross-bumps_00031}\\[0.2ex]
            & \teaserrowasterix{davis}{train_00038}\\[0.2ex]
            & \teaserrowasterix{davis}{hockey_00046}\\[0.2ex]

    \end{tabular}}
	\caption{\textbf{Additional qualitative results on real-world data.} Despite being trained on partially annotated synthetic datasets, \OURS{} demonstrates generalization to multi-task real-world scenarios. *Note that \semantic{} is trained on closed-set driving classes although it generalizes to some extent to other domains.}
	\label{fig:supp:qualitative}
\end{figure*}

\newpage
\clearpage
\clearpage

{
    \small
    \bibliographystyle{ieeenat_fullname}
    \bibliography{main}
}

\end{document}